\documentclass{article}

\PassOptionsToPackage{numbers}{natbib}
\usepackage[pdftex]{graphicx}

\usepackage[preprint]{neurips_2024}

\usepackage{amsmath}  
\usepackage{algorithm}  
\usepackage{algorithmic}  
\usepackage{amsthm}
\newtheorem{theorem}{Theorem}
\newtheorem{definition}{Definition}
\newtheorem{lemma}{Lemma}

\newtheorem*{proposition*}{Proposition}

\newtheorem*{corollary*}{Corollary}
\newtheorem{assumption}[theorem]{Assumption}
\usepackage{multirow}
\usepackage{subcaption}
\usepackage{amssymb}
\usepackage{caption} 
\usepackage{booktabs} 
\usepackage{multirow} 

\usepackage[utf8]{inputenc} 
\usepackage[T1]{fontenc}    
\usepackage{hyperref}       
\usepackage{url}            
\usepackage{booktabs}       
\usepackage{amsfonts}       
\usepackage{nicefrac}       
\usepackage{microtype}      
\usepackage{xcolor}   
\usepackage{natbib}
\usepackage{amsmath}  
\usepackage{amssymb}  
\usepackage{rotating}
\usepackage{graphicx}
\usepackage{pgffor}   
\usepackage{subcaption} 
\usepackage{array}
\usepackage{afterpage}

\usepackage{cleveref}  

\title{Augmenting Offline RL with Unlabeled Data}

\author{%
  Zhao Wang\thanks{Please contact Zhao.Wang@sony.com for more information} \\
  Sony Group Corporation\\
  Waseda University\\
  \And
  Briti Gangopadhyay \\
  Sony Group Corporation\\
  \AND
  Jia-Fong Yeh\\
  National Taiwan University \\
  \And
  Shingo Takamatsu \\
  Sony Group Corporation\\
}

\begin{document}

\maketitle

\begin{abstract}


Recent advancements in offline Reinforcement Learning (Offline RL) have led to an increased focus on methods based on conservative policy updates to address the Out-of-Distribution (OOD) issue. These methods typically involve adding behavior regularization or modifying the critic learning objective, focusing primarily on states or actions with substantial dataset support. However, we challenge this prevailing notion by asserting that the absence of an action or state from a dataset does not necessarily imply its suboptimality. In this paper, we propose a novel approach to tackle the OOD problem. We introduce an offline RL teacher-student framework, complemented by a policy similarity measure. This framework enables the student policy to gain insights not only from the offline RL dataset but also from the knowledge transferred by a teacher policy. The teacher policy is trained using another dataset consisting of state-action pairs, which can be viewed as practical domain knowledge acquired without direct interaction with the environment. We believe this additional knowledge is key to effectively solving the OOD issue. This research represents a significant advancement in integrating a teacher-student network into the actor-critic framework, opening new avenues for studies on knowledge transfer in offline RL and effectively addressing the OOD challenge.

\end{abstract}

\section{Introduction}

The burgeoning field of offline Reinforcement Learning (offline RL) \citep{fujimoto2019off,fujimoto2021minimalist,cql,kostrikov2021iql,dt} offers a promising avenue for leveraging pre-collected datasets to train RL agents without further interactions with the environment. This approach reduces the cost and complexity of data collection~\citep{yang2022rorl} and enables safer and more efficient learning compared with online RL\citep{offlineRLSurvey}.

Unlike in online RL, the Out-of-Distribution (OOD) problem arises when the trained policy encounters states or actions not represented in the offline training dataset ~\citep{fujimoto2019off, bear}. Classical methods in offline RL have focused on addressing OOD action issues. These methods include adding a behavior regularization term ~\citep{fujimoto2021minimalist,kumar2019stabilizing,wu2019behavior}, modifying the critic learning objective ~\citep{cql,kostrikov2021iql,eql}, or incorporating behavior regularization through uncertainty ~\citep{edac, UncertaintyWA} to prevent the risk of selecting OOD actions, they are based on the assumption that OOD actions not present in the offline dataset are potentially harmful since the offline RL dataset is comprehensive enough to cover the distribution of almost all possible actions and states ~\citep{singh2022offline,diverse}.

However, being conservative and sticking strictly to offline data can be detrimental in most practical scenarios. Our preliminary experiments show that across all domains in the MuJoCo tests, when a specific range of states and their corresponding actions are removed from the offline data, the trained offline RL policy suffers significantly. For example, if 60\% of the data in a specific range is removed from the Walker2d task, the score of the trained offline TD3BC policy drops from 93.21 to 2.68, and the IQL policy score also falls to 14.04. See Table~\ref{tab1:general_mujoco} for all the performance declines.  

Another simple yet direct examples are presented in ~\citep{ghosh2022offline, jiang2023offline, singh2022offline, diverse}. \citep{ghosh2022offline} uses a navigation task to illustrate that if offline data is focused exclusively on major city roads due to sampling cost constraints, the trained agent will have little opportunity to choose smaller but shorter side streets when inference. \citep{jiang2023offline} also claims offline RL methods ~\citep{fujimoto2021minimalist,kumar2019stabilizing,wu2019behavior} neglect of the imbalance of real-world dataset distributions (OOD data) in the development of models, which also leads significant performance drop in their test as same reason we claim. \citep{diverse} investigates the performance drop of current offline RL algorithms under comprehensive data corruption problem settings.  

Our motivation is to train a generalized offline RL policy without the impractical requirement of collecting an enormous amount of data to cover most possible state-action pairs and labeling rewards for each data transition. Instead, can we train the policy using partially covered offline RL data and leverage another relevant dataset that can be easily obtained without reward labeling? For example, in a navigation task, can we use the carefully labeled driving recorder data from our own car and supplement it with other drivers' recorder data without reward labeling to train a general offline navigation policy for our city? 

Recently, researchers have shown enthusiasm for using additional unlabeled data ~\citep{zolna2020offline, pmlr-v202-li23b, yu2022leverage} to augment traditional offline RL, as they argue that labeling rewards for classical RL tasks might be costly or unrealistic. A more practical scenario claimed in ~\citep{zolna2020offline, yu2022leverage} is to have a portion of task-specific (e.g., cutting an onion) offline RL data, combined with another portion of data (e.g., picking up an onion or cutting a carrot) that is not labeled with task-agnostic rewards. One natural solution is to learn a reward function from the labeled RL data and use it to label the unlabeled data, as done in ~\citep{zolna2020offline}. Another solution is to simply assign a zero reward ~\citep{yu2022leverage}. However, as suggested in our baseline comparision in Section~\ref{experiments}, both two possible solutions fails into sub-optimality due to their incapability to solve the OOD data issue as they also need to guarantee the almost full data coverage on task. Furthermore, one could employ behavior cloning (BC) or x\% BC on either the combined data or the unlabeled data. However, due to the differing sampling policies, BC or x\% BC from neither unlabeled data nor two datasets yield augmented results.

 In this paper, we propose \textbf{Ludor}, which is a novel offline RL teacher-student framework. This approach enables the student policy to learn from offline OOD RL data, augmented by knowledge transferred from a teacher policy, which is trained on unlabeled medium or expert data. Besides, we introduce policy discrepancy measures to refine loss computation and reduce bootstrapping errors, which may lead to the overestimation of critic value for OOD data. This measure is a non-probabilistic, one-step-free policy metric, designed specifically to enhance the computation of the loss function in our offline RL teacher-student method. Experimental results indicate that our algorithm can be readily integrated into any actor-critic algorithms, demonstrating the superiority of our method compared to existing baselines in various standardized environments. These results also validate the effectiveness of a student policy learning concurrently from offline data and teacher-derived insights and highlight the significance of each component in our methodology.

\section{Related Work}
\label{rw}

\textbf{Offline RL with Unlabeled Data.} 
Lu ~\citep{yu2022leverage, singh2020cog} considers a setting that reward can be missing from a subset of data, and uses a strategy of giving zero reward to these transitions. ORIL ~\citep{zolna2020offline} trains a reward function based on a subset of offline RL data, and labels the other subset of unlabeled data with the trained reward function. We consider these approaches as a part of our comparison baselines. 
Another trend including PDS ~\citep{hu2023provable} and MAHALO ~\citep{pmlr-v202-li23b} construct a pessimistic reward function and a critic function by using a ensemble of neural networks are similar but not applicable for our work. They uses additional penalties on the reward function or critic network learned from labeled data to prevent overestimation, ensuring a conservative algorithm to stick with offline labeled data. But our problem setting is not to be conservative, the instability in either reward function or critic network does not necessarily imply sub-optimal policy. 

\textbf{Offline RL without Conservatism.} 
\citep{ghosh2022offline} provides a clear example of a navigation task to demonstrate why conservatism is often unnecessary in practical scenarios where offline RL data is biased, similar to previously discussed out-of-distribution (OOD) issues. Similarly, \citep{jiang2023offline} argues that in cases of "imbalanced" data, adhering conservatively to the dataset can be detrimental. \citep{singh2022offline} introduces the concept of "Heteroskedasticity" to illustrate that typical offline RL problems, which rely on distributional constraints, fail to learn from data characterized by non-uniform variability due to the requirement to stay close to the behavior policy. \citep{diverse} addresses the issue of "data corruption" by both mathematically and empirically analyzing Implicit Q-Learning (IQL) \citep{kostrikov2021iql}. They use different terms to articulate that conservatism is not necessary in many practical scenarios where offline data cannot encompass all possible states and actions, providing substantial inspiration. However, since they did not utilize additional unlabeled data in their analyses, it is neither fair nor appropriate to consider them as baselines for comparison. For simplicity, we adopt the term \textbf{OOD data} to represent a relatively general problem statement in our later literature.

\textbf{Teacher-Student Network.} Teacher-student network architectures have been widely applied across various research fields, including knowledge distillation~\citep{HintonVD15,8100237}, computer vision\citep{xie2020self,sohn2020simple,wang2021data,wang2018kdgan,u2pl,gtaseg}, natural language processing~\citep{bert,tang2019distilling}, recommendation systems~\citep{tang2018ranking} and RL applications~\citep{offlineRLKnowledge,marlteacher}. Diverging from traditional knowledge distillation~\citep{HintonVD15}, recent teacher-student architecture has been successfully adopted for other knowledge learning objectives, such as knowledge expansion~\citep{xie2020self,sohn2020simple}, adaptation~\citep{matiisen2019teacher,li2019bidirectional}, and multi-task learning~\citep{wang2021data,hu2022teacher}. In the realm of semi-supervised image segmentation, the teacher-student network has been utilized to process a mix of a small amount of labeled data and a larger volume of unlabeled data, with the student and teacher networks learning respectively~\citep{sohn2020simple}. The merging of this acquired knowledge is facilitated through Knowledge Transfer via Exponential Moving Average (EMA) of all parameters~\citep{gtaseg, u2pl}. Drawing inspiration from these methodologies, we propose that the teacher and student networks learn from knowledge-based and OOD data, respectively. Subsequently, they integrate their acquired knowledge, with the transfer to the student being facilitated through EMA.

\section{Preliminaries and Proposed Method}
\subsection{Preliminaries}
Offline RL learns from an offline dataset \(\mathcal{D}\) consisting of $(s, a, r, s')$ tuples that follows Markov Decision Process (MDP) dynamics $\mathcal{M}:=(\mathcal{S}, \mathcal{A}, P, R, \gamma)$. Here, \( s \in \mathcal{S} \), and \( \mathcal{S} \) is the set of states. Similarly, \( \mathcal{A} \) denotes the set of all possible actions, with \( a \in \mathcal{A} \). The state transition probability matrix \( P \) defines the probability \( P(s'|s, a) \) of transitioning to state \( s' \) from state \( s \) after taking action \( a \). The reward \( r \in R \) is received after making a transition from state \( s \) due to action \( a \). The discount factor \( \gamma \), a value between 0 and 1, reflects the relative importance of future rewards compared to immediate rewards.

We use $\pi_\phi: \mathcal{S} \rightarrow  \Delta(\mathcal{A})$ to denote the policy network ($\Delta(\mathcal{A})$ indicates choosing action from state $s$) with parameter $\phi$ (or actor policy in any actor-critic based offline RL algorithm), and $\mathcal{J}(\phi)=\mathbb{E}\left[ \sum_{t=0}^\infty \gamma  r \sim \phi (\cdot |s_t)\right] $ to denote the expected discounted return of $\phi$. The goal of offline RL is to find the optimal $\phi^*$ that can maximizes $\mathcal{J}(\phi)$.

In continuous action spaces of offline RL problem, finding $\pi_\phi^*$ is always achieved by finding optimal action that maximum the value function $Q^\phi(s,a)$ as: $\max_{\phi} \mathbb{E}_{s \sim \mathcal{D}, a \sim \pi_{\phi}(s)} \left[ Q^\phi(s,a) \right]$, here $Q^\phi(s,a)$ is approximately predicted by a critic network $Q_{\theta}^{\phi}(s, a)$ with the parameter $\theta$. In off-policy or offline policy learning, $Q_{\theta}^{\phi}(s, a)$ is usually updated by minimizing the loss function $\left[ \delta^2 \right]$ using temporal difference (TD) error as: 
\begin{equation}\label{q_loss}
\mathcal{L}_{\theta} = \mathbb{E} \left[ \delta^2 \right] = \mathbb{E} \left[ \left( Q_{\theta}^{\phi}(s, a) - \left( r +  \gamma Q_{\theta^-}^{\phi}(s', a') \right) \right)^2 \right].
\end{equation}
here, \( \delta\) is the TD error, \( Q_{\theta^-}^{\phi}(s', a') \) is the maximum estimated Q-value predicted by using target network with parameter $\theta^-$.

The actor's goal is to learn a policy that maximizes the Q-values estimated by the critic as:
\begin{equation}\label{actor_loss}
\mathcal{L}_{\phi} = -\mathbb{E} \left[ \mu_\pi(s) \right] = -\mathbb{E} \left[  \max_{a} Q_{\theta}^{\phi}(s, a) \right] = -\mathbb{E} \left[ Q_{\theta}^{\phi}(s, \pi_\phi(s)) \right],
\end{equation}
here, $\mu_\pi(s)$ represents $\max_{a} Q_{\theta}^{\phi}(s, a)$, which is the maximum Q-value achievable under the current policy $\pi$ for a given state $s$. 

\begin{figure*}[t]
\centering
\includegraphics[width=0.99\textwidth]{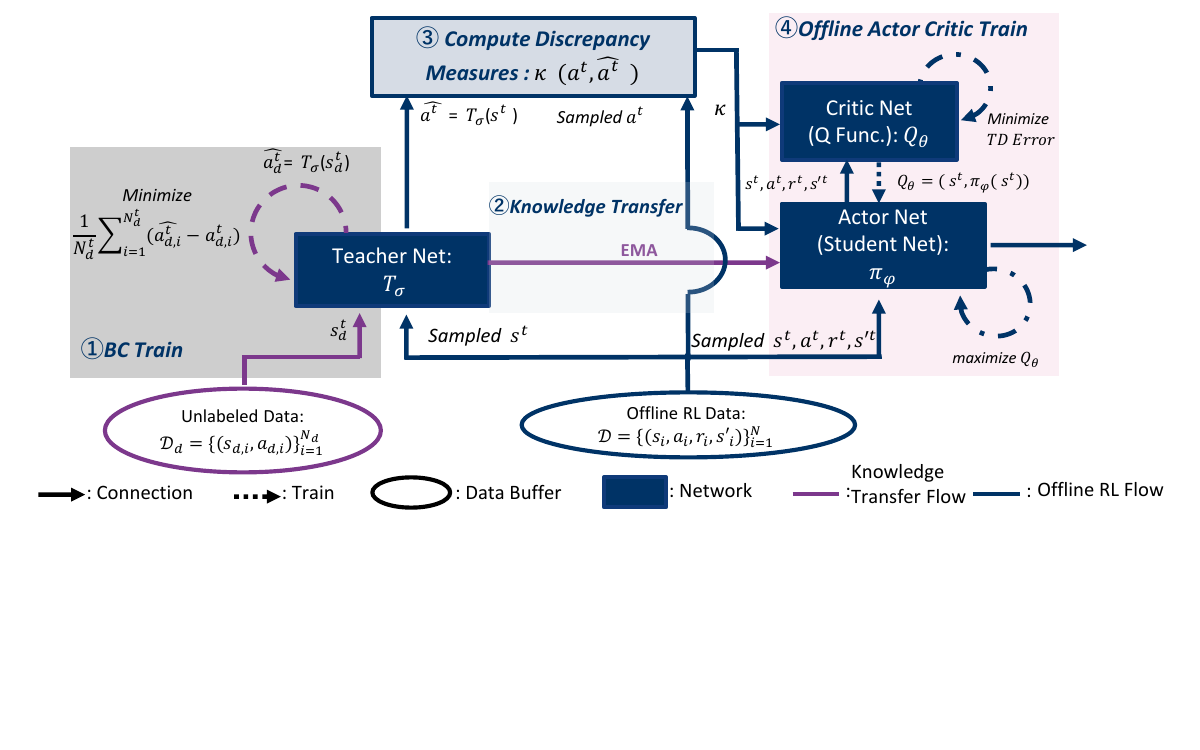}
\vskip -1.1in
\caption{Ludor consists of two networks of the same architecture but difference weight, named the teacher network and the student network respectively. The teacher $T_{\sigma}$ learns the knowledge via $\mathcal{D}_d$ by using BC, while the student learns from OOD data $\mathcal{D}$, and augmented by the knowledge transfered from the teacher via EMA. At the same time, the discrepancy measures are computed, and assign a vector weight to the loss function of actor-critic training.}
\label{fig:architecture}

\end{figure*}

\subsection{Overview of Ludor}
Given a dataset from an offline RL OOD $\mathcal{D} =(\mathcal{S}, \mathcal{A}, R, \mathcal{S}') = {(s_i, a_i, r_i, s'_i)}_{i=1}^N$ and a teacher unlabeled dataset $\mathcal{D}_d=(\mathcal{S}_d, \mathcal{A}_d)={(s_{d,i}, a_{d,i})}_{i=1}^{N_d}$, \textbf{our goal is to train a student model $\pi_\phi^*: \mathcal{S} \rightarrow \Delta(\mathcal{A})$} to overcome the OOD problem by utilizing a teacher-student network approach.

Figure \ref{fig:architecture} gives an overview of Ludor, which follows the typical teacher-student framework with two network models of the same architecture but different weights, named the teacher network $T_{\sigma}$ and the student network $\pi_{\phi}$, respectively. The teacher network $T_{\sigma}$ learns knowledge via $\mathcal{D}_d=(\mathcal{S}_d, \mathcal{A}_d)$ by using BC training. On the other hand, the student network $\pi_\phi$ learns from two sources: 1) an OOD offline RL dataset $\mathcal{D} =(\mathcal{S}, \mathcal{A}, R, \mathcal{S}')$ via a typical actor-critic based offline RL algorithm; and 2) knowledge transferred from the teacher via EMA. Regarding the first source, to handle continuous action control, we adopt two actor-critic baselines: TD3BC\citep{fujimoto2021minimalist} and IQL\citep{kostrikov2021iql}. We treat our student network identically to the actor network in each baseline.

\subsection{Methodology}
Ludor consists of the following steps: 

\begin{algorithm}[tb]
   \caption{The General Framework of Our Method}
   \label{alg:life}
\begin{algorithmic}
   \STATE {\bfseries Require:} Offline OOD data buffer \( \mathcal{D} \), unlabeled data buffer \(\mathcal{D}_d \),  teacher network  $T_{\sigma}$ , student (policy) network $\pi_{\phi}$, critic network $Q_{\theta}$ with parameters \( \sigma \), \( \phi \), \( \theta \) respectively
   \STATE Initialize target and other network if required
   \STATE {\bfseries at epoch 0}  train $T_{\sigma}$ using data sampled from \(\mathcal{D}_d \), and copy it weight to student (policy) network $\pi_{\phi}$
   \REPEAT
   \STATE   Sample a mini-batch $B= {(s, a, r, s')}$ from \( \mathcal{D} \)
   \STATE   Sample a mini-batch $B_d= {(s_d, a_d)}$ from \( \mathcal{D}_d \)
   \STATE   BC train $B_d$ to update teacher network defined in {\bfseries \cref{teacherbc,teacherleaning}}
   \STATE   Update actor (student) network via $\mathbf{EMA}$ defined in {\bfseries \cref{eq::ema}}
   \STATE   Compute discrepancy measures using {\bfseries \cref{kappa}} 
   \STATE   Update $Q_{\theta}$ using loss func. $\mathcal{L}_{\theta}^t$ in  {\bfseries \cref{q_loss_final}}
   \STATE   Update $\pi_{\phi}$ using loss func. $\mathcal{L}_{\phi}^t$ in {\bfseries \cref{actor_loss_final}}
   \STATE   Soft Update target network if required
   \UNTIL {Convergence criteria is met}

\end{algorithmic}
\end{algorithm}

\textbf{Step 1: pretrain the teacher network only at training step $t=0$.}
As reported in other knowledge learning works\citep{gtaseg,u2pl}, initializing the weight of teacher network and student network with same weight is a key factor to make the knowledge transfer succeed. For simplicity, we sample a batch from $\mathcal{D}_d$ to pretrain the teacher network at Epoch 0 during our training, and copy its weight to the student network.  

\textbf{Step 2: BC train the teacher.}
At each training step $t$ , we first train the teacher network $T_{\sigma}$ by minimizing the BC loss: 

\begin{equation} \label{teacherbc}
    \mathcal{L}_T{}_\sigma^t = \frac{1}{N_d^t} \sum_{i=1}^{N_d^t} (\hat{a_{d,i}^t} - a_{d,i}^t)^2, 
\end{equation}
here $\mathcal{L}_T{}_\sigma^t$ is the training loss of the teacher at training step $t$ , $\hat{a_{d,i}^t}=T_{\sigma}(s_{d,i}^t)$ is the $i$th output action of the teacher, $(s_{d,i}^t,a_{d,i}^t)$ is the $i$th sampled state and action from $\mathcal{D}_d$, ${N_d^t}$ is the sampling size at $t$.

\textbf{Step 3: transfer the knowledge of the teacher to the student network.}
After the optimization process of the teacher, which is: 
\begin{equation}\label{teacherleaning}
    \sigma^t := \sigma^t - \eta \frac{\partial}{\partial \sigma^t} (\mathcal{L}_T{}_\sigma^t),
\end{equation}
here \( \eta \) is the learning rate, \( \sigma^t \) is the parameters of the teacher network at training step \( t\).
we transfer the knowledge learned from the teacher to the student via EMA by Equation~\ref{eq::ema}. 

\begin{equation} 
    \phi^t := \alpha\phi^{(t-1)}+ (1-\alpha)\sigma^t,
    \label{eq::ema}
\end{equation}
here $\phi^t$ is the parameters of the student, $\sigma^t$ is the parameters of the teacher at training step $t$, $\alpha$ is the hyper-parameter of EMA. Through our knowledge transmission, the knowledge learned by teacher model is leveraged to be transferred to the student.

\textbf{Step 4: compute the policy discrepancy measures}. The calculation of our proposed policy discrepancy measures, $\kappa$, is detailed in \cref{kappa}. In general, $\kappa$ represents the cosine similarity between two actions: $a^t$, the action sampled from $\mathcal{D}$, and $\hat{a^t}=T_{\sigma}(s^t)$, the output of the teacher network $T_{\sigma}$, given the sampled state $s^t$ at training step $t$. The detailed calculation and the rationale behind this approach are provided in the following subsection.

\textbf{Step 5: train the student via actor-critic based offline RL method}. After computing the discrepancy measures $\kappa(a^t, \hat{a^t})$, we dot the product $\kappa(a^t, \hat{a^t})$, which is a vector indicating the weight of each data over the training batch $B$, we calculate a weighted loss for critic network as:

\begin{equation} \label{q_loss_final}
 \mathcal{L}_{\theta}^t = \frac{  \kappa(a^t, \hat{a^t}) \cdot \left[ \delta^2 \right]}{\sum_{b=1}^{B} \kappa(a_b^t, \hat{a_b^t})},  
\end{equation}
here, $\cdot$ denotes the dot product, $B$ is the batch size, $a_b^t$ is each action taken at training step $t$ in the action batch $a^t$, $\hat{a_b^t}$ is the correspondent teacher action batch, and $\left[ \delta^2 \right]$ is the loss function of critic network in any value estimation based actor-critic offline RL method in \cref{q_loss}. 

Similar to the critic network update, the loss of actor is: 
\begin{equation} \label{actor_loss_final}
 \mathcal{L}_{\phi}^t = \frac{ \kappa(a^t, \hat{a^t}) \cdot \left[ \mu_\pi(s) \right]}{\sum_{b=1}^{B} \kappa(a_b^t, \hat{a_b^t})} ,
\end{equation}
here $\mu_\pi(s)$ is the maximum Q-value achievable under the current policy $\pi_\phi$ for a given state $s$. 

Note that, \cref{q_loss_final,actor_loss_final} can be manipulated with any value estimation based actor-critic offline RL methods, i.e. TD3BC\citep{fujimoto2021minimalist} and IQL\citep{kostrikov2021iql}. For instance, as TD3BC\citep{fujimoto2021minimalist} uses a smaller output between two Q networks aiming to reduce overestimation bias, we simply use \cref{q_loss_final} on the smaller one; TD3BC\citep{fujimoto2021minimalist} adds a BC term to \cref{actor_loss} in order to stick with offline data, we also add a same term on \cref{actor_loss_final}; IQL\citep{kostrikov2021iql} adds an weighted implicit policy improvement term to \cref{actor_loss}, we also add the term but dot product with our policy similar measure term $\kappa$.

\subsection{Policy Discrepancy Measures}

Our proposed policy discrepancy measures plays a crucial role in our offline RL teacher-student framework. We argue that treating OOD offline RL data and unlabeled data equally like \citep{gtaseg} did can cause extrapolation errors in OOD offline data, potentially leading to overestimated critic values and suboptimal offline policies. Therefore, at each training step $t$, given a state $s^t$ sampled from the offline RL dataset $\mathcal{D}$ , we compute the cosine similarity between two actions: $a^t$, the corresponding action in the dataset, and $\hat{a^t}=T_{\sigma}(s)$, the output of the teacher network $T_{\sigma}$. The cosine similarity $\kappa$ is defined as:

\begin{equation}\label{kappa}
    \kappa(a^t, \hat{a^t}) = 1 + \frac{a^t \cdot \hat{a^t}}
    {\|a^t\| \|\hat{a^t}\|},
\end{equation}

here $\cdot$ denotes the dot product and $\|\cdot\|$ denotes the Euclidean norm. 
A $\kappa$ value close to 2 indicates that $a^t$ and $\hat{a^t}$ are very similar. Conversely, a low $\kappa$ value (close to 0) indicates dissimilar actions. This weight for each data piece is then incorporated into the loss function in offline RL (refer to Equation~\ref{q_loss}, Equation~\ref{actor_loss}) to indicate the relative importance of each data piece in the learning process. Intuitively, when a distribution shift occurs in the offline RL dataset, learning from the shifted part may lead to an overestimation of the extrapolation error. We aim to minimize learning from the shifted part, as this can be effectively addressed through teacher knowledge transfer.

\section{Experiments}
\label{experiments}

\textbf{Experimental Domains.} We evaluated Ludor on two domains: Mujoco~\citep{todorov2012mujoco} domain with 3 tasks: HalfCheetah, Walker2D, Hopper and a navigation domain Antmaze~\citep{fu2020d4rl} by using the D4RL benchmark~\citep{fu2020d4rl}.

\textbf{OOD Data Creation.}  For the OOD data problem setting, we intentionally created a OOD offline RL dataset. Furthermore, we removed a certain percentage of the transition data in specific dimensions of states. For example, in the HalfCheetah-Medium task, we removed a certain percentage (see \ref{knowledge-ood} for ablation study) of the transition data in a certain range (see \ref{dataDim} for ablation study) of state spaces. Conversely, for the unlabeled dataset, we extracted only 1\% of medium- or expert-level data (see \ref{AblationStudy} for ablation study) corresponding to the task, retaining only the state-action pairs for teacher training.


\textbf{Training and Evaluation.}
For training the student, we used OOD offline RL dataset of each task created mentioned before. For training the teacher, we used the correspondent unlabeled medium or expert dataset. We trained our algorithm for 0.12 million time steps and evaluated every 2000 time steps. Each evaluation consists of 10 episodes. All run time experiments were run with a single Tesla v100 GPU and Intel Xeon E5-2686 v4 CPUs at 2.3GHz. The training hyper-parameters can be found in Appendix~\ref{app:hyper}.

\subsection{General Performance on Various Tasks}
\label{general-performance-section}
In comparison, we considered the two related actor critic based offline RL methods: TD3BC\citep{fujimoto2021minimalist} and IQL\citep{kostrikov2021iql} as we wanted to prove our method can be built upon on any actor critic based method. The baselines of our comparison are: 

\textbf{UDS\citep{yu2022leverage}}: the method that leverages an additional unlabeled dataset into offline RL dataset by asserting 0 into reward;

\textbf{ORIL\citep{zolna2020offline}}: the method that leverages an additional unlabeled dataset into offline RL dataset by supervised training a reward function to label unlabeled data;

\textbf{BC(OOD+K)}: BC on both offline RL data and unlabeled data;

\textbf{BC(K)}: BC on only 1\% unlabeled data, which is also the teacher performance shown in \ref{teacher=bc} ;

\textbf{TD3BC(OOD+BC(K))} and \textbf{IQL(OOD+BC(K))}: the method that adds an additional loss term as defined in \cref{baselinea_actor_loss}
\begin{equation} \label{baselinea_actor_loss}
 \mathcal{L}_{\phi_{Baseline}}^t =  \left[ \mu_\pi(s) \right]  + \frac{1}{N_d^t} \sum_{i=1}^{N_d^t} ( \pi_\phi(s_{d,i}^t) - a_{d,i}^t)^2,
\end{equation}
here, the first term in \cref{baselinea_actor_loss} is identical to offline RL baseline, for instance, if the baseline is TD3BC, then it is computed based on the policy's ability to maximize the Q-value output with an additional BC term, if the baseline is IQL, then it is computed to maximize the estimated Q-values as shown in \cref{actor_loss}. The second term represents the BC loss when the student network performs BC on the unlabeled dataset $\mathcal{D}_d$. The objective of this baseline is straightforward: we aim for the student network to learn from both the OOD dataset in offline RL and the unlabeled dataset through BC.

We report the final performance results in Table~\ref{tab1:general_mujoco} and Table~\ref{tab1:general_maze}.

\begin{table}[t]
\caption{Average normalized scores from the final 10 evaluations on Mujoco domain with 3 tasks (3 seeds). Key: U.D. - Unlabeled Data, RL.D. - Offline RL Dataset, Exp. - Expert, Med. - Medium, Ran. - Random, 1\% - random 1\% extraction from the original dataset.}
\label{tab1:general_mujoco}

\centering
\scriptsize
\begin{tabular}{lccccccc}
\toprule
\multicolumn{3}{c}{Dataset} & BC & \multicolumn{2}{c}{IQL} & \multicolumn{2}{c}{TD3BC} \\
\cmidrule(lr){1-3} \cmidrule(lr){4-4} \cmidrule(lr){5-6} \cmidrule(lr){7-8}
U.D. & RL.D. & Task & OOD+K & OOD & OOD+K(BC) & OOD & OOD+K(BC) \\
\midrule
Exp.1\% & Med.1\% & HalfCheetah & 42.46$\pm$3.12 & 18.73$\pm$20.47 & 46.93$\pm$19.99 & 31.90$\pm$4.17 & 39.46$\pm$5.43 \\
Exp.1\% & Med.1\% & Hopper & 56.56$\pm$11.24 & 44.51$\pm$21.88 & 49.94$\pm$7.78 & 38.28$\pm$8.43 & 46.76$\pm$27.72 \\
Exp.1\% & Med.1\% & Walker2d & 62.26$\pm$32.52 & 14.04$\pm$6.53 & 29.83$\pm$10.25 & 2.68$\pm$8.55 & 61.49$\pm$31.5 \\
Med.1\% & Ran.1\% & HalfCheetah & 29.72$\pm$1.23 & 1.65$\pm$1.16 & 1.39$\pm$0.91 & 8.04$\pm$0.74 & 38.15$\pm$2.71 \\
Med.1\% & Ran.1\% & Hopper & 4.87$\pm$3.08 & 1.43$\pm$0.11 & 1.98$\pm$1.06 & 3.08$\pm$2.46 & 8.99$\pm$10.84 \\
Med.1\% & Ran.1\% & Walker2d & 3.60$\pm$2.34 & 0.02$\pm$0.05 & 0.08$\pm$0.04 & 0.64$\pm$0.18 & 1.23$\pm$1.32 \\
\bottomrule
\end{tabular}

\ContinuedFloat

\begin{tabular}{lccccccrcc}

\multicolumn{3}{c}{Dataset} & USD\citep{yu2022leverage} & ORIL\citep{zolna2020offline} & Ludor+IQL & Ludor+TD3BC \\
\cmidrule(lr){1-3}
U.D. & RL.D. & Task & & & & \\
\midrule
Exp.1\% & Med.1\% & HalfCheetah & 39.11$\pm$2.14 & 13.70$\pm$5.17 & \textbf{61.05$\pm$4.32} & \textbf{69.73$\pm$5.87} \\
Exp.1\% & Med.1\% & Hopper & 17.07$\pm$6.26 & 32.33$\pm$29.86 & \textbf{63.43$\pm$18.67} & \textbf{87.92$\pm$18.15} \\
Exp.1\% & Med.1\% & Walker2d & 7.16$\pm$3.06 & 8.35$\pm$11.47 & \textbf{84.83$\pm$18.9} & \textbf{93.21$\pm$24.9} \\
Med.1\% & Ran.1\% & HalfCheetah & 5.06$\pm$2.85 & 30.22$\pm$3.76 & \textbf{34.56$\pm$1.43} & \textbf{42.02$\pm$2.12} \\
Med.1\% & Ran.1\% & Hopper & 11.73$\pm$7.32 & 27.17$\pm$8.44 & \textbf{44.92$\pm$6.00} & \textbf{43.97$\pm$7.11} \\
Med.1\% & Ran.1\% & Walker2d & 0.50$\pm$0.61 & 1.13$\pm$1.13 & \textbf{42.65$\pm$13.01} & \textbf{49.23$\pm$33.86} \\
\bottomrule
\end{tabular}

\end{table}

\begin{table}[ht]
\caption{Average normalized scores from the final 10 evaluations on the navigation task (3 seeds). Key: Uma. - Umaze in Antmaze, Uma-d. - Umaze Diverse. The other keys are as same as \cref{tab1:general_mujoco}.}
\label{tab1:general_maze}
\scriptsize 
\centering

\begin{tabular}{lccccccc}
\toprule
\multicolumn{3}{c}{Dataset} & BC & \multicolumn{2}{c}{IQL} & \multicolumn{2}{c}{TD3BC} \\
\cmidrule(lr){1-3} \cmidrule(lr){4-4} \cmidrule(lr){5-6} \cmidrule(lr){7-8}
U.D. & RL.D. & Task & OOD+K & OOD & OOD+K(BC) & OOD & OOD+K(BC) \\
\midrule
Uma-d.1\% & Uma.1\% & Antmaze & \textbf{49.02$\pm$18.90} & 2.04$\pm$1.87 & 48.54$\pm$17.22 & 10.06$\pm$6.20 & 29.41$\pm$31.94 \\
Med-d.1\% & Med-p.1\% & Antmaze & 0.00$\pm$0.00 & 0.00$\pm$0.00 & 0.00$\pm$0.00 & 0.00$\pm$0.00 & 0.00$\pm$0.00 \\
\bottomrule
\end{tabular}

\ContinuedFloat

\begin{tabular}{lcccccc}
\multicolumn{3}{c}{Dataset} & USD\citep{yu2022leverage} & ORIL\citep{zolna2020offline} & Ludor+IQL & Ludor+TD3BC \\
\cmidrule(lr){1-3}
U.D. & RL.D. & Task & & & \\
\midrule
Uma-d.1\% & Uma.1\% & Antmaze & 6.12$\pm$7.74 & 2.08$\pm$2.94 & \textbf{46.75$\pm$12.17} & \textbf{52.17$\pm$16.34} \\
Med-d.1\% & Med-p.1\% & Antmaze & 0.00$\pm$0.00 & 0.00$\pm$0.00 & 0.00$\pm$0.00 & 0.00$\pm$0.00 \\
\bottomrule
\end{tabular}

\end{table}

\begin{table}[ht]
\centering
\caption{Performance of Teacher and Student in various tasks with Our Method built on TD3BC}
\vskip 0.05in
\begin{scriptsize}
\begin{tabular}{lcccc}
\toprule
 & HalfCheetah & Hopper & Walker2D & Antmaze \\  
\midrule
Teacher & 52.23$\pm$3.33 & 79.10$\pm$36.80 & \textbf{98.99$\pm$3.20} & 51.09$\pm$20.22 \\ 

Student & \textbf{69.73$\pm$5.87} & \textbf{87.92$\pm$18.15} & 93.21$\pm$24.9 & \textbf{52.17$\pm$16.34} \\ \bottomrule
\end{tabular}
\label{tab:teacher-student}
\end{scriptsize}
\end{table}

\subsection{Teacher (1\% BC on Unlabeled Data) and Student Analysis}
\label{teacher=bc}
As the teacher acquires knowledge by training on expert-level dataset and transfers this learned knowledge to the student via EMA during training, similar to other studies\citep{gtaseg,u2pl} on knowledge transfer tasks, a major concern arises: whether the student is simply replicating the teacher's policy. To address this, we present a comparison between the teacher's and the student's performances in Table \ref{tab:teacher-student}, highlighting various tasks in D4RL\citep{fu2020d4rl}.

Table \ref{tab:teacher-student} reveals that, with the exception of the Walker2D task in Mujoco\citep{todorov2012mujoco}, the student's performance significantly exceeds that of the teacher in other tasks, including HalfCheetah and Hopper in Mujoco\citep{todorov2012mujoco}, as well as AntMaze. This outcome indicates that our student network is adept at learning from an offline, OOD dataset and effectively assimilating knowledge transferred from the teacher's policy, which is informed by a dataset rich in knowledge. Regarding the Walker2D task, its inherent simplicity may explain why BC from a 1\% expert dataset rapidly achieves optimal performance during training, as detailed in Appendix \ref{app:teacher-student}. This simplicity poses a challenge for our methods in producing comparable results. The performance curves for all these tasks are presented in Appendix \ref{app:teacher-student}.

\subsection{Unlabeled Data Coverage and Removal Ratios}
\label{knowledge-ood}
In Table~\ref{tab:robustness}, we present the performance differences under two experimental settings: the teacher unlabeled  data coverage and the removal ratio in the OOD offline RL dataset. For the teacher unlabeled  data coverage ratio, we consider three scenarios: unlabeled data covering 60\%, 80\%, and 100\% of the state space. Regarding the removal ratio, we intentionally create a OOD dataset by removing portions of the data — specifically 0.4, 0.6, and 0.8 (using fraction numbers to distinguish from unlabeled data coverage percentages) — from a specific range of the state space. Observations from Table~\ref{tab:robustness} reveal that: 1) unlabeled data coverage affects performance, with greater coverage correlating to higher performance, and in some tasks, such as Halfcheetah and Antmaze, the unlabeled data needs to cover at least 60\% of the state spaces; 2) the removal ratio has a marginal effect on performance, where increased data removal in the offline RL dataset leads to lower performance. Performance curves for all these tasks are presented in Appendix \ref{app::robustness}.
	
\begin{table}
\centering
\caption{Performance Differences under Teacher unlabeled data Coverage and Removal Ratio}
\label{tab:robustness}
\vskip 0.05in
\begin{scriptsize}
\begin{tabular}{lccc}
\toprule
Unlabeled Data Coverage & 60\%  & 80\%  & 100\%  \\
\midrule
Half-Cheetah & 3.26$\pm$0.88 & 13.71$\pm$2.91 & \textbf{69.73$\pm$5.87} \\
Hopper       & 38.26$\pm$20.02 & 48.23$\pm$22.75 & \textbf{87.92$\pm$18.15} \\
Walker2D     & 42.33$\pm$22.00 &69.12$\pm$22.78 & \textbf{93.21$\pm$24.90}\\
Antmaze      & 2.03$\pm$4.70 &39.23$\pm$31.29 & \textbf{52.16$\pm$16.34}\\
\midrule
Removal Ratio & 0.4  & 0.6  & 0.8 \\
\midrule
Half-Cheetah & \textbf{81.16$\pm$5.10} & 69.73$\pm$5.87 & 51.75$\pm$9.55 \\
Hopper       & 83.87$\pm$16.33 & \textbf{87.92$\pm$18.15} & 48.60$\pm$18.69 \\
Walker2D     & \textbf{93.73$\pm$14.26} & 93.21$\pm$24.90 & 59.50$\pm$13.49\\
Antmaze      & \textbf{52.36$\pm$8.46} & 52.16$\pm$16.34 & 49.65$\pm$10.91\\
\bottomrule
\end{tabular}
\end{scriptsize}
\end{table}

\subsection{Ablation Studies}
\label{AblationStudy}

\textbf{Components Ablation}: we strategically combine three distinct components of our proposed method: the teacher-student network, the discrepancy measures, and knowledge transfer via EMA. The result is shown in Table~\ref{tab:Ablation-com}, where 'cross' refers to the component missing.

\begin{table*}[ht]
\caption{Components ablation study for Ludor built on TD3BC.}
\label{tab:Ablation-com}
\centering 
{\scriptsize 
\begin{tabular}{lllcccc}
\toprule
Teacher-Student & Discrepancy Measures & EMA & HalfCheetah & Hopper & Walker2D & Antmaze \\
\midrule
$\times$   & $\times$  & $\times$  & 31.90$\pm$4.17 & 38.28$\pm$8.43   & 2.68 $\pm$ 8.55 & 10.06 $\pm$ 8.55 \\
$\surd$  & $\surd$ & $\times$  & 35.55 $\pm$ 7.20 & 42.49$\pm$4.28    & 5.43 $\pm$ 2.20 & 4.90$\pm$9.27 \\
$\surd$  & $\times$  & $\surd$ & 41.87$\pm$32.48   & 65.00$\pm$29.70  & 66.41$\pm$45.63 & 50.83$\pm$24.22 \\
$\surd$  & $\surd$ & $\surd$ & \textbf{69.73$\pm$5.87}   & \textbf{87.92$\pm$18.15 } & \textbf{93.21$\pm$19.98} &\textbf{52.17$\pm$16.34} \\
\bottomrule
\end{tabular}
} 
\end{table*}

\textbf{EMA Ablation}: as shown at Table~\ref{tab:ema_performance}, Ludor performs steadily under different EMA weight settings.

\begin{table}[ht]
\centering
\caption{EMA ablation study for Ludor built on TD3BC.}
\label{tab:ema_performance}

{\scriptsize 
\begin{tabular}{lllccc}
\toprule
K.D & RL.D & Task & ema 0.999 & ema 0.99 & ema 0.9 \\
\midrule
Exp.1\% & Med.1\% & HalfCheetah & 61.32$\pm$10.82 & 65.21$\pm$6.11 & 69.73$\pm$5.87 \\
Exp.1\% & Med.1\% & Hopper & 82.13$\pm$7.57 & 85.86$\pm$8.28 & 87.92$\pm$18.15 \\
Exp.1\% & Med.1\% & Walker2d & 90.48$\pm$27.65 & 92.76$\pm$26.77 & 93.21$\pm$24.90 \\
Uma-d.1\% & Uma.1\% & Antmaze & 48.38$\pm$21.24 & 50.23$\pm$19.66 & 52.17$\pm$16.34 \\
\bottomrule
\end{tabular}
} 
\end{table}

\textbf{Data Utilization Percentage Ablation}: To identify the data usage boundaries of original dataset beyond which Ludor yields no significant benefits, we present the following results at  Table~\ref{tab:data_utilization}. Generally, (1)if 1\% of the original data (10k transitions not trajectories) can be obtained, we can ensure the acceptable performance for D4RL tasks, (2) when the amount of data reaches 30\% of the original dataset (both for unlabeled data and RL Data), our algorithm can achieve performance comparable to training solely on the 100\% expert dataset.

\begin{table}[ht]
\centering
\caption{Training data utilization ablation study for Ludor built on TD3BC.}
\label{tab:data_utilization}

{\scriptsize
\begin{tabular}{lllccccc}
\toprule
K.D & RL.D & Task & 0.1\% & 0.5\% & 1\% & 30\% & 50\% \\
\midrule
Exp. & Med. & HalfCheetah & 2.49$\pm$0.58 & 6.98$\pm$0.79 & 69.73$\pm$5.87 & 85.08$\pm$1.46 & 86.49$\pm$1.48 \\
Exp. & Med. & Hopper & 12.65$\pm$8.27 & 50.37$\pm$29.01 & 87.92$\pm$18.15 & 103.85$\pm$9.93 & 106.89$\pm$4.82 \\
Exp. & Med. & Walker2d & 5.05$\pm$2.91 & 57.69$\pm$18.98 & 93.21$\pm$24.90 & 107.81$\pm$0.82 & 107.82$\pm$0.32 \\
Uma-d. & Uma. & Antmaze & 0.00$\pm$0.00 & 3.18$\pm$1.22 & 52.17$\pm$16.34 & 48.54$\pm$15.77 & 49.12$\pm$12.12 \\
\bottomrule
\end{tabular}
}
\end{table}

\textbf{Discrepancy Measure Ablation}: We conducted a brief experiment to demonstrate the performance of three distinct kinds of action constraints for all three D4RL tasks. For KL divergence, we implemented two types as mentioned in the previous rebuttal box: distribution formulated by a. (KL-1)automatically formulation across batch and b. (KL-2) manually specifying std. The detailed results of this analysis are presented in Appendix \ref{app::cos}.

\subsection{Robustness under Different Data Removal Dimension in Observation Space}
\label{dataDim}
Another concern regarding our policy for creating offline distribution RL datasets is that removing data from different dimensions of the observation space might lead to varied consequences, given the varying importance of each dimension. We utilized the Hopper-Medium-v2 dataset from D4RL\citep{fu2020d4rl} to test how the performance of our proposed method changes when data from different dimensions are removed from the observation spaces. In our experiments, we selectively removed the most densely populated segments within each observational dimension for each sub-experiment. Our findings indicate that: 1) removing the same percentage of data from different dimensions of observation spaces results in varying performance scores; 2) overall, no features are significantly crucial to the extent of causing a substantial performance gap when 60\% of the data is removed from the most densely populated parts of those dimensions. The detailed results of this analysis are presented in Appendix \ref{app:features} due to space limitations.

\section*{Conclusions and Limitations}
Most recent offline RL algorithms adopt a conservative approach to address OOD problem. These methods often involve adding behavior regularization, modifying the critic's learning objective, or incorporating uncertainty, with a focus on states or actions well-supported in the datasets. In this paper, we introduce Ludor, an innovative offline RL teacher-student framework that enables the student policy to glean insights not only from the offline RL dataset but also from the knowledge imparted by a teacher policy, informed by an unlabeled dataset. Our experimental results demonstrate that our algorithm can seamlessly integrate with existing actor-critic algorithms, outperforming current baselines in various standardized environments and robust enougn under different experimental settings.

We identify limitations of our method as follows: (1) it performs poorly when two conditions are met: sparse rewards and a relatively complex environment, as observed in our experiment with the Antmaze medium environment; (2) for some tasks, such as Halfcheetah and Antmaze, the unlabeled data data needs to cover at least 60\% of the state spaces. These findings provide direction for our future research.

\bibliography{Neurips2024}

\begin{thebibliography}{47}
\providecommand{\natexlab}[1]{#1}
\providecommand{\url}[1]{\texttt{#1}}
\expandafter\ifx\csname urlstyle\endcsname\relax
  \providecommand{\doi}[1]{doi: #1}\else
  \providecommand{\doi}{doi: \begingroup \urlstyle{rm}\Url}\fi

\bibitem[Fujimoto et~al.(2019)Fujimoto, Meger, and Precup]{fujimoto2019off}
Scott Fujimoto, David Meger, and Doina Precup.
\newblock Off-policy deep reinforcement learning without exploration.
\newblock In \emph{International Conference on Machine Learning}, pages 2052--2062, 2019.

\bibitem[Fujimoto and Gu(2021)]{fujimoto2021minimalist}
Scott Fujimoto and Shixiang~Shane Gu.
\newblock A minimalist approach to offline reinforcement learning.
\newblock In \emph{Thirty-Fifth Conference on Neural Information Processing Systems}, 2021.

\bibitem[Kumar et~al.(2020)Kumar, Zhou, Tucker, and Levine]{cql}
Aviral Kumar, Aurick Zhou, George Tucker, and Sergey Levine.
\newblock Conservative q-learning for offline reinforcement learning.
\newblock \emph{Advances in Neural Information Processing Systems}, 33:\penalty0 1179--1191, 2020.

\bibitem[Kostrikov et~al.(2021)Kostrikov, Nair, and Levine]{kostrikov2021iql}
Ilya Kostrikov, Ashvin Nair, and Sergey Levine.
\newblock Offline reinforcement learning with implicit q-learning.
\newblock \emph{arXiv}, 2021.

\bibitem[Chen et~al.(2021)Chen, Lu, Rajeswaran, Lee, Grover, Laskin, Abbeel, Srinivas, and Mordatch]{dt}
Lili Chen, Kevin Lu, Aravind Rajeswaran, Kimin Lee, Aditya Grover, Michael Laskin, Pieter Abbeel, Aravind Srinivas, and Igor Mordatch.
\newblock Decision transformer: Reinforcement learning via sequence modeling.
\newblock \emph{arXiv preprint arXiv:2106.01345}, 2021.

\bibitem[Yang et~al.(2022)Yang, Bai, Ma, Wang, Zhang, and Han]{yang2022rorl}
Rui Yang, Chenjia Bai, Xiaoteng Ma, Zhaoran Wang, Chongjie Zhang, and Lei Han.
\newblock Rorl: Robust offline reinforcement learning via conservative smoothing.
\newblock In \emph{Advances in Neural Information Processing Systems}, 2022.

\bibitem[Singh et~al.(2022{\natexlab{a}})Singh, Kumar, and Singh]{offlineRLSurvey}
Bharat Singh, Rajesh Kumar, and Vinay~Pratap Singh.
\newblock Reinforcement learning in robotic applications: A comprehensive survey.
\newblock \emph{Artif. Intell. Rev.}, 55\penalty0 (2):\penalty0 945–990, feb 2022{\natexlab{a}}.
\newblock ISSN 0269-2821.
\newblock URL \url{https://doi.org/10.1007/s10462-021-09997-9}.

\bibitem[Kumar et~al.(2019{\natexlab{a}})Kumar, Fu, Tucker, and Levine]{bear}
Aviral Kumar, Justin Fu, George Tucker, and Sergey Levine.
\newblock Stabilizing off-policy q-learning via bootstrapping error reduction.
\newblock In \emph{Neural Information Processing Systems}, 2019{\natexlab{a}}.

\bibitem[Kumar et~al.(2019{\natexlab{b}})Kumar, Fu, Soh, Tucker, and Levine]{kumar2019stabilizing}
Aviral Kumar, Justin Fu, Matthew Soh, George Tucker, and Sergey Levine.
\newblock Stabilizing off-policy q-learning via bootstrapping error reduction.
\newblock \emph{Advances in Neural Information Processing Systems}, 32, 2019{\natexlab{b}}.

\bibitem[Wu et~al.(2019)Wu, Tucker, and Nachum]{wu2019behavior}
Yifan Wu, George Tucker, and Ofir Nachum.
\newblock Behavior regularized offline reinforcement learning.
\newblock \emph{arXiv preprint arXiv:1911.11361}, 2019.

\bibitem[Xu et~al.(2023)Xu, Jiang, Li, Yang, Wang, Chan, and Zhan]{eql}
Haoran Xu, Li~Jiang, Jianxiong Li, Zhuoran Yang, Zhaoran Wang, Victor Wai~Kin Chan, and Xianyuan Zhan.
\newblock Offline rl with no ood actions: In-sample learning via implicit value regularization.
\newblock In \emph{International Conference on Learning Representations}, 2023.

\bibitem[An et~al.(2021)An, Moon, Kim, and Song]{edac}
Gaon An, Seungyong Moon, Jang-Hyun Kim, and Hyun~Oh Song.
\newblock Uncertainty-based offline reinforcement learning with diversified q-ensemble.
\newblock In \emph{Neural Information Processing Systems}, 2021.

\bibitem[Wu et~al.(2021)Wu, Zhai, Srivastava, Susskind, Zhang, Salakhutdinov, and Goh]{UncertaintyWA}
Yue Wu, Shuangfei Zhai, Nitish Srivastava, Joshua~M. Susskind, Jian Zhang, Ruslan Salakhutdinov, and Hanlin Goh.
\newblock Uncertainty weighted actor-critic for offline reinforcement learning.
\newblock In \emph{International Conference on Machine Learning}, 2021.
\newblock URL \url{https://api.semanticscholar.org/CorpusID:234763307}.

\bibitem[Singh et~al.(2022{\natexlab{b}})Singh, Kumar, Vuong, Chebotar, and Levine]{singh2022offline}
Anikait Singh, Aviral Kumar, Quan Vuong, Yevgen Chebotar, and Sergey Levine.
\newblock Offline rl with realistic datasets: Heteroskedasticity and support constraints.
\newblock \emph{arXiv preprint arXiv:2211.01052}, 2022{\natexlab{b}}.

\bibitem[Yang et~al.(2023)Yang, Zhong, Xu, Zhang, Zhang, Han, and Zhang]{diverse}
Rui Yang, Han Zhong, Jiawei Xu, Amy Zhang, Chongjie Zhang, Lei Han, and Tong Zhang.
\newblock Towards robust offline reinforcement learning under diverse data corruption.
\newblock \emph{arXiv preprint arXiv:2310.12955}, 2023.

\bibitem[Ghosh et~al.(2022)Ghosh, Ajay, Agrawal, and Levine]{ghosh2022offline}
Dibya Ghosh, Anurag Ajay, Pulkit Agrawal, and Sergey Levine.
\newblock Offline rl policies should be trained to be adaptive.
\newblock In \emph{International Conference on Machine Learning}, pages 7513--7530. PMLR, 2022.

\bibitem[Jiang et~al.(2023)Jiang, Chen, Qiu, Xu, Chan, and Ding]{jiang2023offline}
Li~Jiang, Sijie Chen, Jielin Qiu, Haoran Xu, Wai~Kin Chan, and Zhao Ding.
\newblock Offline reinforcement learning with imbalanced datasets.
\newblock \emph{arXiv preprint arXiv:2307.02752}, 2023.

\bibitem[Zolna et~al.(2020)Zolna, Novikov, Konyushkova, Gulcehre, Wang, Aytar, Denil, de~Freitas, and Reed]{zolna2020offline}
Konrad Zolna, Alexander Novikov, Ksenia Konyushkova, Caglar Gulcehre, Ziyu Wang, Yusuf Aytar, Misha Denil, Nando de~Freitas, and Scott Reed.
\newblock Offline learning from demonstrations and unlabeled experience.
\newblock \emph{arXiv preprint arXiv:2011.13885}, 2020.

\bibitem[Li et~al.(2023)Li, Boots, and Cheng]{pmlr-v202-li23b}
Anqi Li, Byron Boots, and Ching-An Cheng.
\newblock {MAHALO}: Unifying offline reinforcement learning and imitation learning from observations.
\newblock In Andreas Krause, Emma Brunskill, Kyunghyun Cho, Barbara Engelhardt, Sivan Sabato, and Jonathan Scarlett, editors, \emph{Proceedings of the 40th International Conference on Machine Learning}, volume 202 of \emph{Proceedings of Machine Learning Research}, pages 19360--19384, 23--29 Jul 2023.

\bibitem[Yu et~al.(2022)Yu, Kumar, Chebotar, Hausman, Finn, and Levine]{yu2022leverage}
Tianhe Yu, Aviral Kumar, Yevgen Chebotar, Karol Hausman, Chelsea Finn, and Sergey Levine.
\newblock How to leverage unlabeled data in offline reinforcement learning.
\newblock In \emph{International Conference on Machine Learning}, pages 25611--25635. PMLR, 2022.

\bibitem[Singh et~al.(2020)Singh, Yu, Yang, Zhang, Kumar, and Levine]{singh2020cog}
Avi Singh, Albert Yu, Jonathan Yang, Jesse Zhang, Aviral Kumar, and Sergey Levine.
\newblock Cog: Connecting new skills to past experience with offline reinforcement learning.
\newblock \emph{arXiv preprint arXiv:2010.14500}, 2020.

\bibitem[Hu et~al.(2023)Hu, Yang, Zhao, and Zhang]{hu2023provable}
Hao Hu, Yiqin Yang, Qianchuan Zhao, and Chongjie Zhang.
\newblock The provable benefits of unsupervised data sharing for offline reinforcement learning.
\newblock \emph{arXiv preprint arXiv:2302.13493}, 2023.

\bibitem[Hinton et~al.(2015)Hinton, Vinyals, and Dean]{HintonVD15}
Geoffrey~E. Hinton, Oriol Vinyals, and Jeffrey Dean.
\newblock Distilling the knowledge in a neural network.
\newblock \emph{CoRR}, abs/1503.02531, 2015.
\newblock URL \url{http://dblp.uni-trier.de/db/journals/corr/corr1503.html#HintonVD15}.

\bibitem[Yim et~al.(2017)Yim, Joo, Bae, and Kim]{8100237}
Junho Yim, Donggyu Joo, Jihoon Bae, and Junmo Kim.
\newblock A gift from knowledge distillation: Fast optimization, network minimization and transfer learning.
\newblock In \emph{2017 IEEE Conference on Computer Vision and Pattern Recognition (CVPR)}, pages 7130--7138, 2017.
\newblock \doi{10.1109/CVPR.2017.754}.

\bibitem[Xie et~al.(2020)Xie, Luong, Hovy, and Le]{xie2020self}
Qizhe Xie, Minh-Thang Luong, Eduard Hovy, and Quoc~V Le.
\newblock Self-training with noisy student improves imagenet classification.
\newblock In \emph{Proceedings of the IEEE/CVF conference on computer vision and pattern recognition}, pages 10687--10698, 2020.

\bibitem[Sohn et~al.(2020)Sohn, Zhang, Li, Zhang, Lee, and Pfister]{sohn2020simple}
Kihyuk Sohn, Zizhao Zhang, Chun-Liang Li, Han Zhang, Chen-Yu Lee, and Tomas Pfister.
\newblock A simple semi-supervised learning framework for object detection.
\newblock \emph{arXiv preprint arXiv:2005.04757}, 2020.

\bibitem[Wang et~al.(2021)Wang, Li, Guo, Fang, and Wang]{wang2021data}
Zhenyu Wang, Yali Li, Ye~Guo, Lu~Fang, and Shengjin Wang.
\newblock Data-uncertainty guided multi-phase learning for semi-supervised object detection.
\newblock In \emph{Proceedings of the IEEE/CVF Conference on Computer Vision and Pattern Recognition}, pages 4568--4577, 2021.

\bibitem[Wang et~al.(2018)Wang, Zhang, Sun, and Qi]{wang2018kdgan}
Xiaojie Wang, Rui Zhang, Yu~Sun, and Jianzhong Qi.
\newblock Kdgan: Knowledge distillation with generative adversarial networks.
\newblock In \emph{Advances in neural information processing systems}, 2018.

\bibitem[Wang et~al.(2022)Wang, Wang, Shen, Fei, Li, Jin, Wu, Zhao, and Le]{u2pl}
Yuchao Wang, Haochen Wang, Yujun Shen, Jingjing Fei, Wei Li, Guoqiang Jin, Liwei Wu, Rui Zhao, and Xinyi Le.
\newblock Semi-supervised semantic segmentation using unreliable pseudo-labels.
\newblock \emph{the IEEE/CVF Conference on Computer Vision and Pattern Recognition}, pages 4248--4257, 2022.

\bibitem[Jin et~al.(2022)Jin, Wang, and Lin]{gtaseg}
Ying Jin, Jiaqi Wang, and Dahua Lin.
\newblock Semi-supervised semantic segmentation via gentle teaching assistant.
\newblock \emph{Advances in Neural Information Processing Systems}, 2022.

\bibitem[Devlin et~al.(2018)Devlin, Chang, Lee, and Toutanova]{bert}
Jacob Devlin, Ming-Wei Chang, Kenton Lee, and Kristina Toutanova.
\newblock Bert: Pre-training of deep bidirectional transformers for language understanding.
\newblock \emph{arXiv preprint arXiv:1810.04805}, 2018.

\bibitem[Tang et~al.(2019)Tang, Lu, Liu, Mou, Vechtomova, and Lin]{tang2019distilling}
Raphael Tang, Yao Lu, Linqing Liu, Lili Mou, Olga Vechtomova, and Jimmy Lin.
\newblock Distilling task-specific knowledge from bert into simple neural networks.
\newblock \emph{arXiv preprint arXiv:1903.12136}, 2019.

\bibitem[Tang and Wang(2018)]{tang2018ranking}
Jiaxi Tang and Ke~Wang.
\newblock Ranking distillation: Learning compact ranking models with high performance for recommender system.
\newblock In \emph{Proceedings of the 24th ACM SIGKDD international conference on knowledge discovery \& data mining}, pages 2289--2298, 2018.

\bibitem[Tseng et~al.(2022)Tseng, Wang, Lin, and Isola]{offlineRLKnowledge}
Wei-Cheng Tseng, Tsun-Hsuan~Johnson Wang, Yen-Chen Lin, and Phillip Isola.
\newblock Offline multi-agent reinforcement learning with knowledge distillation.
\newblock In S.~Koyejo, S.~Mohamed, A.~Agarwal, D.~Belgrave, K.~Cho, and A.~Oh, editors, \emph{Advances in Neural Information Processing Systems}, volume~35, pages 226--237. Curran Associates, Inc., 2022.

\bibitem[Zheng et~al.(2021)Zheng, Chen, Duan, Lin, Shao, Wang, Wang, and Xu]{marlteacher}
Ying Zheng, Haoyu Chen, Qingyang Duan, Lixiang Lin, Yiyang Shao, Wei Wang, Xin Wang, and Yuedong Xu.
\newblock Leveraging domain knowledge for robust deep reinforcement learning in networking.
\newblock In \emph{IEEE INFOCOM 2021 - IEEE Conference on Computer Communications}, pages 1--10, 2021.
\newblock \doi{10.1109/INFOCOM42981.2021.9488863}.

\bibitem[Matiisen et~al.(2019)Matiisen, Oliver, Cohen, and Schulman]{matiisen2019teacher}
Tambet Matiisen, Avital Oliver, Taco Cohen, and John Schulman.
\newblock Teacher--student curriculum learning.
\newblock \emph{IEEE transactions on neural networks and learning systems}, 31\penalty0 (9):\penalty0 3732--3740, 2019.

\bibitem[Li et~al.(2019)Li, Yuan, and Vasconcelos]{li2019bidirectional}
Yunsheng Li, Lu~Yuan, and Nuno Vasconcelos.
\newblock Bidirectional learning for domain adaptation of semantic segmentation.
\newblock In \emph{Proceedings of the IEEE/CVF Conference on Computer Vision and Pattern Recognition}, pages 6936--6945, 2019.

\bibitem[Hu et~al.(2022)Hu, Li, Liu, Chen, Wang, and Liu]{hu2022teacher}
Chengming Hu, Xuan Li, Dan Liu, Xi~Chen, Ju~Wang, and Xue Liu.
\newblock Teacher-student architecture for knowledge learning: A survey.
\newblock \emph{arXiv preprint arXiv:2210.17332}, 2022.

\bibitem[Todorov et~al.(2012)Todorov, Erez, and Tassa]{todorov2012mujoco}
Emanuel Todorov, Tom Erez, and Yuval Tassa.
\newblock Mujoco: A physics engine for model-based control.
\newblock In \emph{2012 IEEE/RSJ International Conference on Intelligent Robots and Systems}, pages 5026--5033. IEEE, 2012.
\newblock \doi{10.1109/IROS.2012.6386109}.

\bibitem[Fu et~al.(2020)Fu, Kumar, Nachum, Tucker, and Levine]{fu2020d4rl}
Justin Fu, Aviral Kumar, Ofir Nachum, George Tucker, and Sergey Levine.
\newblock D4rl: Datasets for deep data-driven reinforcement learning, 2020.

\bibitem[Tarasov et~al.(2022)Tarasov, Nikulin, Akimov, Kurenkov, and Kolesnikov]{tarasov2022corl}
Denis Tarasov, Alexander Nikulin, Dmitry Akimov, Vladislav Kurenkov, and Sergey Kolesnikov.
\newblock {CORL}: Research-oriented deep offline reinforcement learning library.
\newblock In \emph{3rd Offline RL Workshop: Offline RL as a ''Launchpad''}, 2022.
\newblock URL \url{https://openreview.net/forum?id=SyAS49bBcv}.

\bibitem[Saglam et~al.(2022{\natexlab{a}})Saglam, Cicek, Mutlu, and Kozat]{offpolicy}
Baturay Saglam, Dogan~C Cicek, Furkan~B Mutlu, and Suleyman~S Kozat.
\newblock Off-policy correction for actor-critic methods without importance sampling.
\newblock \emph{arXiv preprint arXiv:2208.00755}, 2022{\natexlab{a}}.

\bibitem[Mahmood et~al.(2014)Mahmood, van Hasselt, and Sutton]{NIPS2014_be53ee61}
A.~Rupam Mahmood, Hado~P van Hasselt, and Richard~S Sutton.
\newblock Weighted importance sampling for off-policy learning with linear function approximation.
\newblock In Z.~Ghahramani, M.~Welling, C.~Cortes, N.~Lawrence, and K.Q. Weinberger, editors, \emph{Advances in Neural Information Processing Systems}, volume~27. Curran Associates, Inc., 2014.
\newblock URL \url{https://proceedings.neurips.cc/paper_files/paper/2014/file/be53ee61104935234b174e62a07e53cf-Paper.pdf}.

\bibitem[Saglam et~al.(2022{\natexlab{b}})Saglam, Cicek, Mutlu, and Kozat]{saglam2022safe}
Baturay Saglam, Dogan~C Cicek, Furkan~B Mutlu, and Suleyman~S Kozat.
\newblock Safe and robust experience sharing for deterministic policy gradient algorithms.
\newblock \emph{arXiv preprint arXiv:2207.13453}, 2022{\natexlab{b}}.

\bibitem[Han and Sung(2019)]{han2019dimension}
Seungyul Han and Youngchul Sung.
\newblock Dimension-wise importance sampling weight clipping for sample-efficient reinforcement learning.
\newblock In \emph{International Conference on Machine Learning}, pages 2586--2595. PMLR, 2019.

\bibitem[Cicek et~al.(2021)Cicek, Duran, Saglam, Mutlu, and Kozat]{cicek2021off}
Dogan~C Cicek, Enes Duran, Baturay Saglam, Furkan~B Mutlu, and Suleyman~S Kozat.
\newblock Off-policy correction for deep deterministic policy gradient algorithms via batch prioritized experience replay.
\newblock In \emph{2021 IEEE 33rd International Conference on Tools with Artificial Intelligence (ICTAI)}, pages 1255--1262. IEEE, 2021.

\bibitem[Ying et~al.(2022)Ying, Hao, Zhou, Su, Yan, and Zhu]{ying2022reuse}
Chengyang Ying, Zhongkai Hao, Xinning Zhou, Hang Su, Dong Yan, and Jun Zhu.
\newblock On the reuse bias in off-policy reinforcement learning.
\newblock \emph{arXiv preprint arXiv:2209.07074}, 2022.

\end{thebibliography}
\bibliographystyle{unsrtnat}

\newpage
\appendix
\onecolumn

\section{Hyper-parameters of Our Proposed Method}
\label{app:hyper}
We code our algorithm based on CORL library\citep{tarasov2022corl}, which is a research-friendly codebase for customize offline RL algorithm. All hyper-parameters can be found in the Table below. 

\begin{table}[ht]
\centering
\caption{Hyper-parameters of Our Proposed Method}
\label{tab:hyperparameters}
\begin{tabular}{lllp{5cm}}
\toprule
\multicolumn{1}{l}{Subcategory} & \multicolumn{1}{l}{Hyper-parameter} & \multicolumn{1}{l}{Value} & \multicolumn{1}{l}{Explanation} \\
\hline
\multirow{7}{*}{Training \& Evaluation } & max\_timesteps & 120000 & Max time steps in training. \\
& eval\_freq & 2000 & Evaluation frequency during training. \\
&n\_episodes & 10 & How many episodes run during evaluation. \\
&batch\_size & 256 & Batch size. \\
&buffer\_size & 20000 & Replay buffer size. \\
&qf\_lr & 3e-4 & Critic learning rate. \\
&actor\_lr & 3e-4 & Actor learning rate. \\

\hline
\multirow{7}{*}{TD3BC} & discount & 0.99 & Reward discount. \\
& expl\_noise & 0.1 & Std of Gaussian exploration noise. \\
& tau & 0.005 & Target network update rate. \\
& policy\_noise & 0.2 & Noise added to target actor during critic update. \\
& noise\_clip & 0.5 & Range to clip target actor noise. \\
& policy\_freq & 2 & Frequency of delayed actor updates. \\
& alpha & 2.5  & Coefficient for Q function in actor loss. \\
\hline
\multirow{5}{*}{IQL } & tau & 0.005 & Target network update rate. \\
& beta & 3.0 & Inverse temperature. \\
&iql\_tau & 0.7 & Coefficient for asymmetric loss. \\
&iql\_deterministic & True & Use deterministic actor. \\
&vf\_lr & 3e-4 & V function learning rate. \\
&qf\_lr & 3e-4 & Critic learning rate. \\
&actor\_lr & 3e-4 & Actor learning rate. \\

\hline
\multirow{2}{*}{OOD \& Teacher Settintgs} & data\_precentange & 0.01 & OOD data sparsity. \\
& ema & 0.9 & Knowledge transfer EMA value. \\
& buffer\_size\_principle & 20000 & Replay buffer size for Unlabeled dataset. \\
& pretrain\_num\_epochs & 1 & Pretrain epoch number. \\
& teacher\_update\_freq & 2 &Teacher update frequency. \\

\bottomrule

\end{tabular}
\end{table}

\newpage
\onecolumn
\section{Additional Evaluation Curves on General Performance of Our Proposed Method}
\label{app:general-random}
This section presents the additional evaluation curves that cannot show in Section \ref{general-performance-section} due to the space limitation.

\begin{figure*}[ht]
\begin{subfigure}{0.24\textwidth}
    \includegraphics[trim={0 0 0 12cm},clip,width=\textwidth]{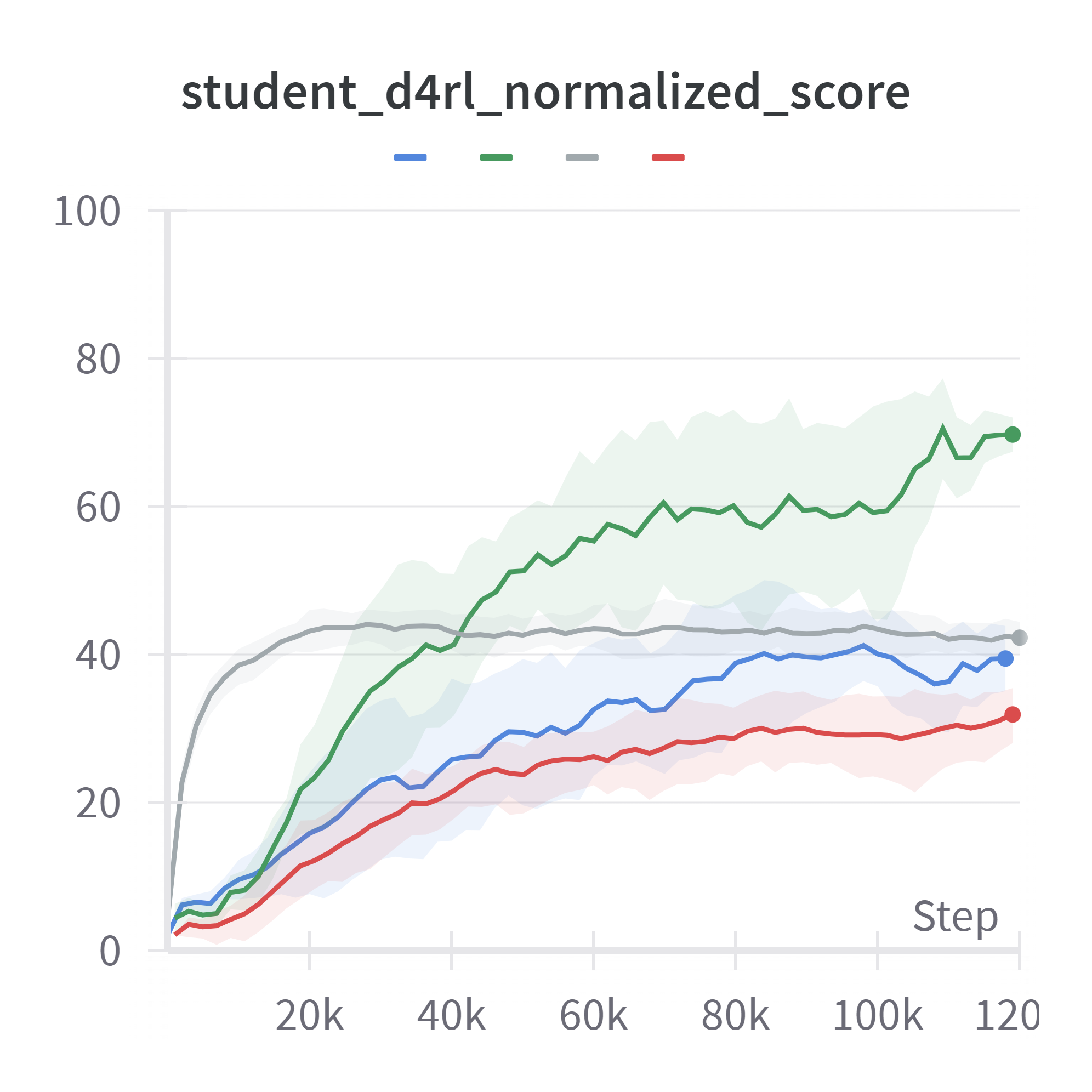}
\end{subfigure}
\hfill 
\begin{subfigure}{0.24\textwidth}
    \includegraphics[trim={0 0 0 14cm},clip,width=\textwidth]{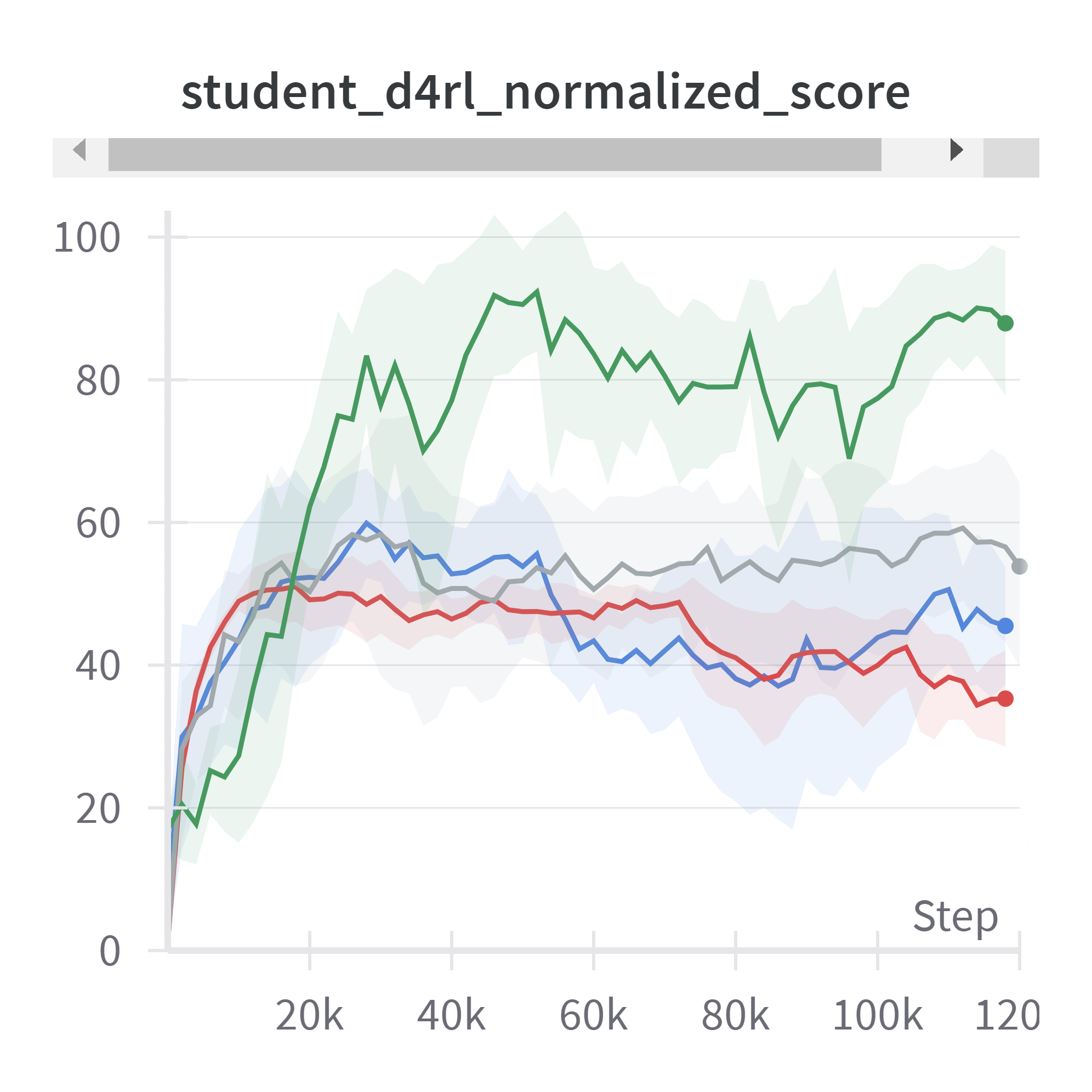}
\end{subfigure}
\hfill 
\begin{subfigure}{0.24\textwidth}
    \includegraphics[trim={0 0 0 12cm},clip,width=\textwidth]{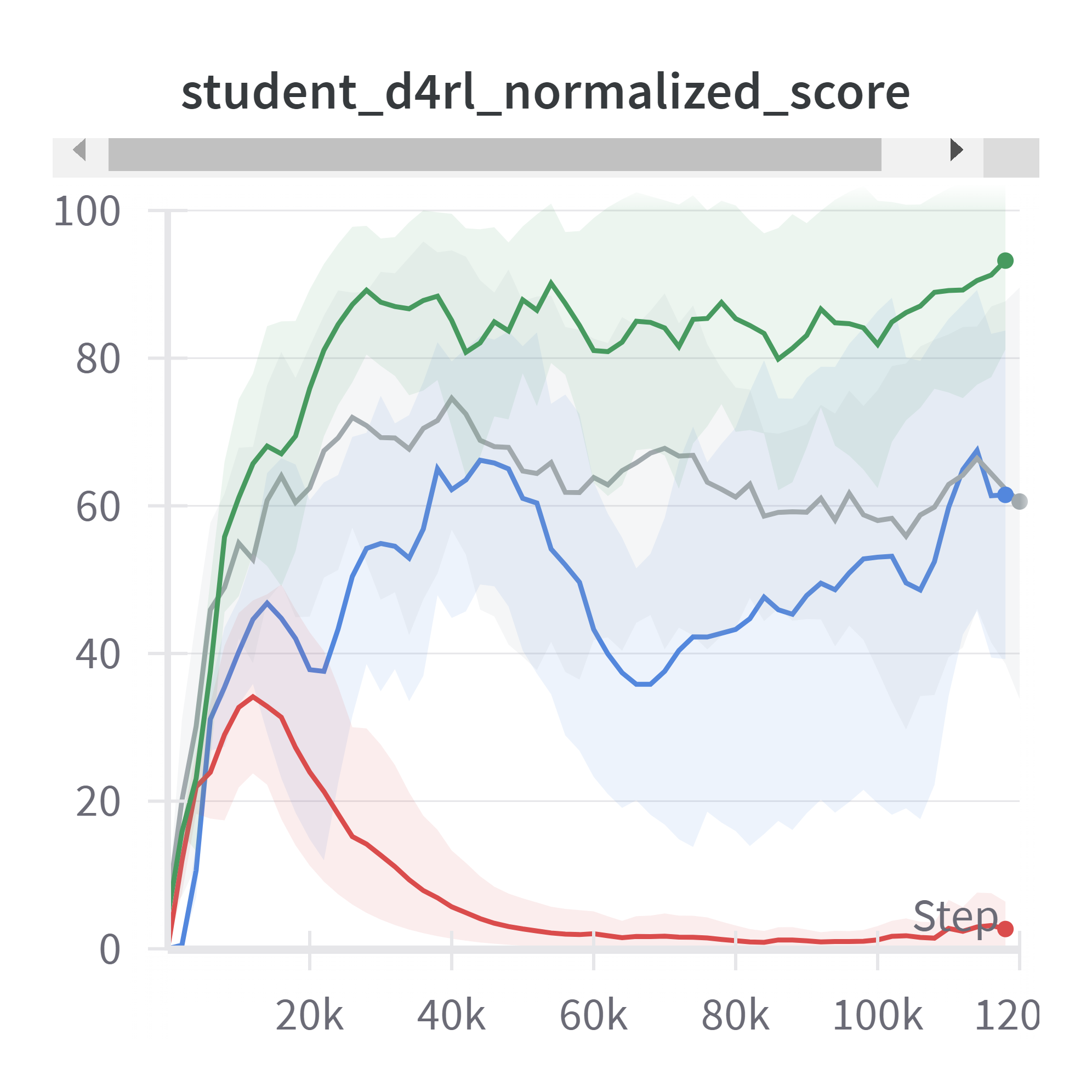}
\end{subfigure}
\hfill 
\begin{subfigure}{0.24\textwidth}
    \includegraphics[trim={0 0 0 12cm},clip,width=\textwidth]{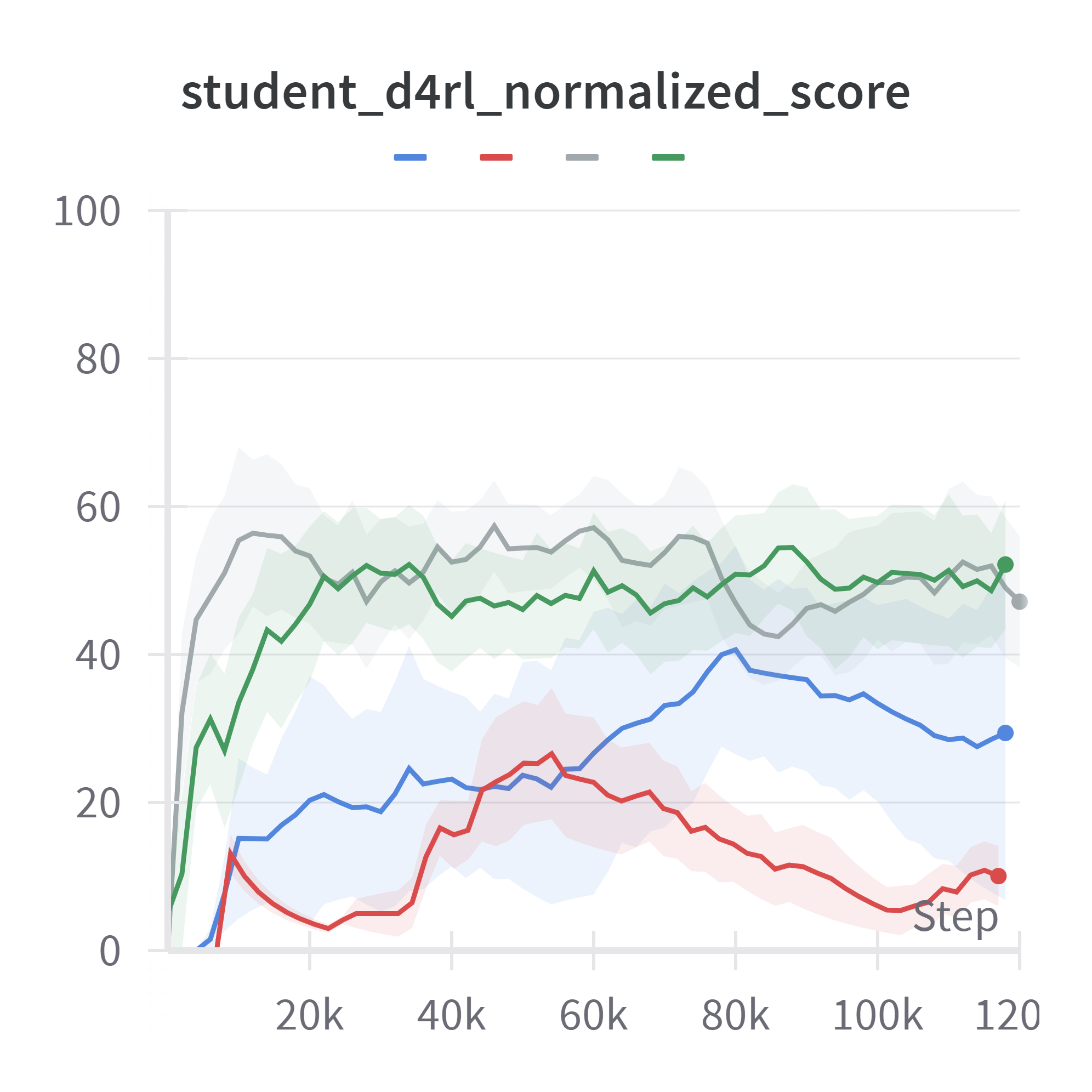}
\end{subfigure}
\vskip -0.1in
\centering
\includegraphics[width=0.6\textwidth,trim={0 11cm 0 0}, clip]{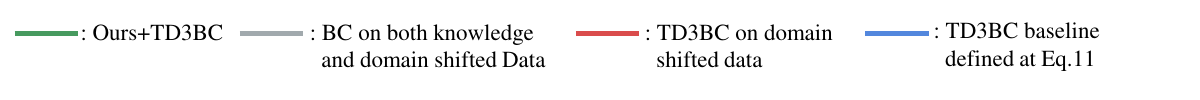}
\centering
\caption{Average normalized score on comparative analysis of Ludor built on TD3BC and baseline models across four environments: HalfCheetah, Hopper, Walker2D, and Antmaze, presented sequentially from left to right. The unlabeled data used is expert data.}

\label{fig:general-td3bc}
\end{figure*}

\begin{figure*}[ht]
\begin{subfigure}{0.24\textwidth}
    \includegraphics[trim={0 0 0 12cm},clip,width=\textwidth]{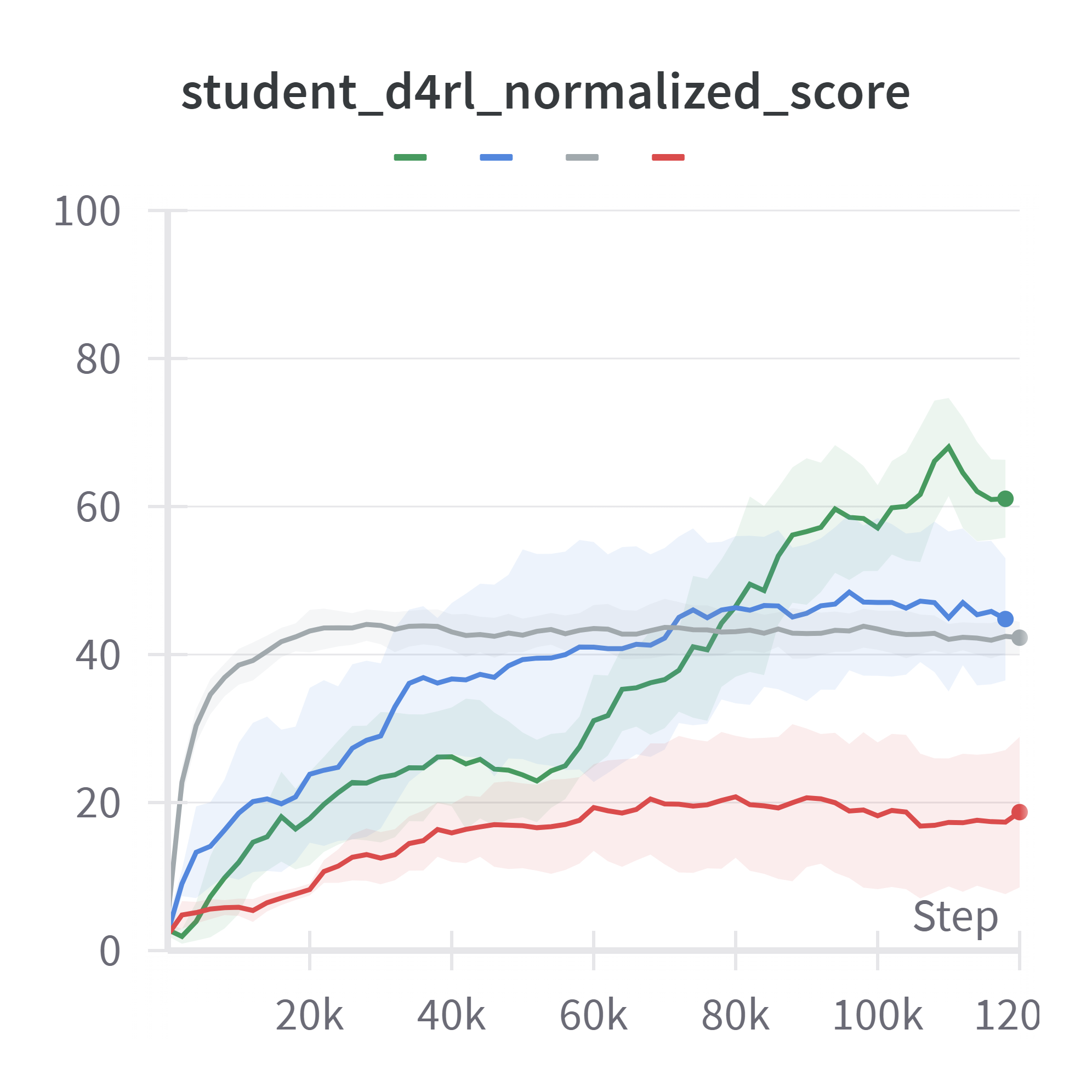}
\end{subfigure}
\hfill 
\begin{subfigure}{0.24\textwidth}
    \includegraphics[trim={0 0 0 12cm},clip,width=\textwidth]{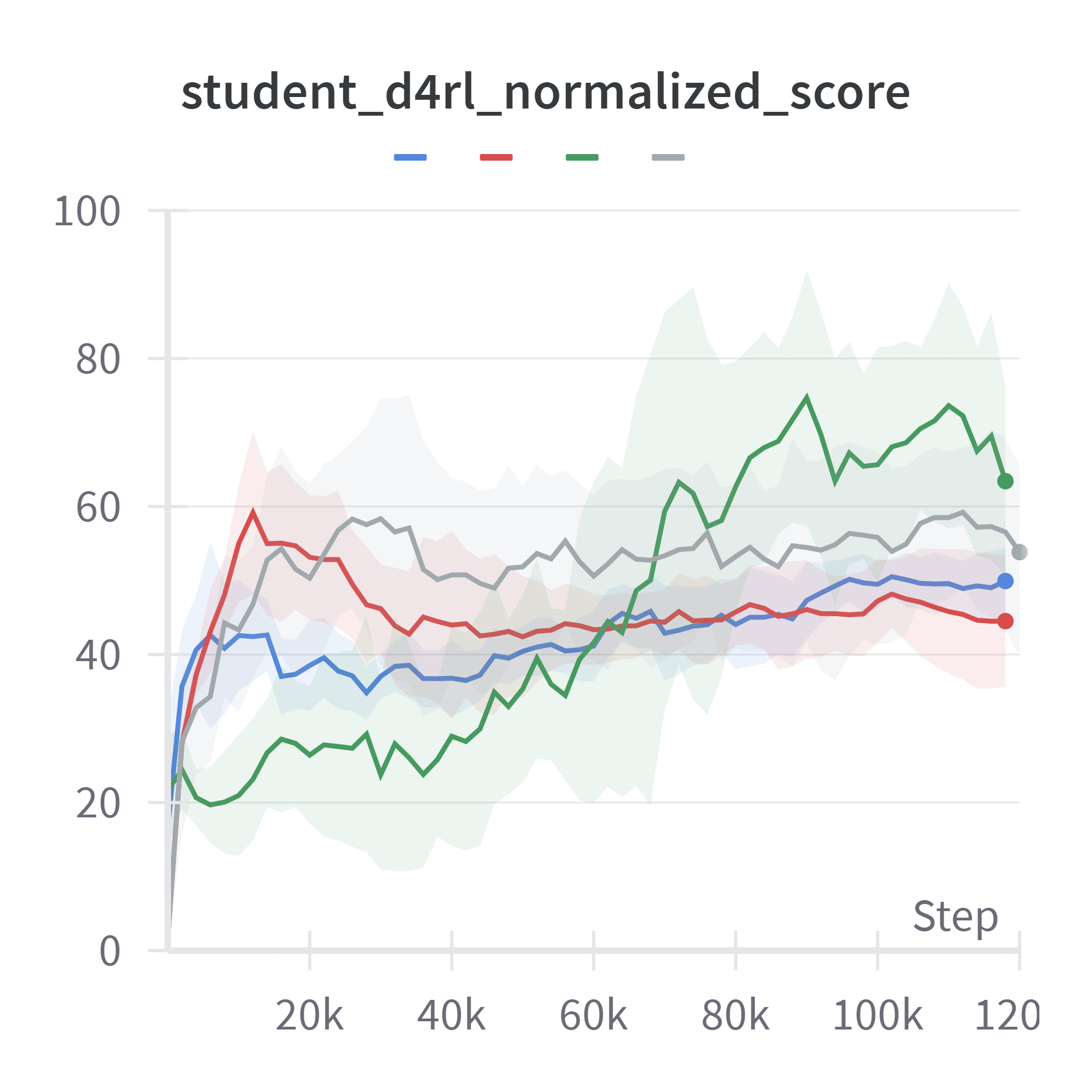}
\end{subfigure}
\hfill 
\begin{subfigure}{0.24\textwidth}
    \includegraphics[trim={0 0 0 12cm},clip,width=\textwidth]{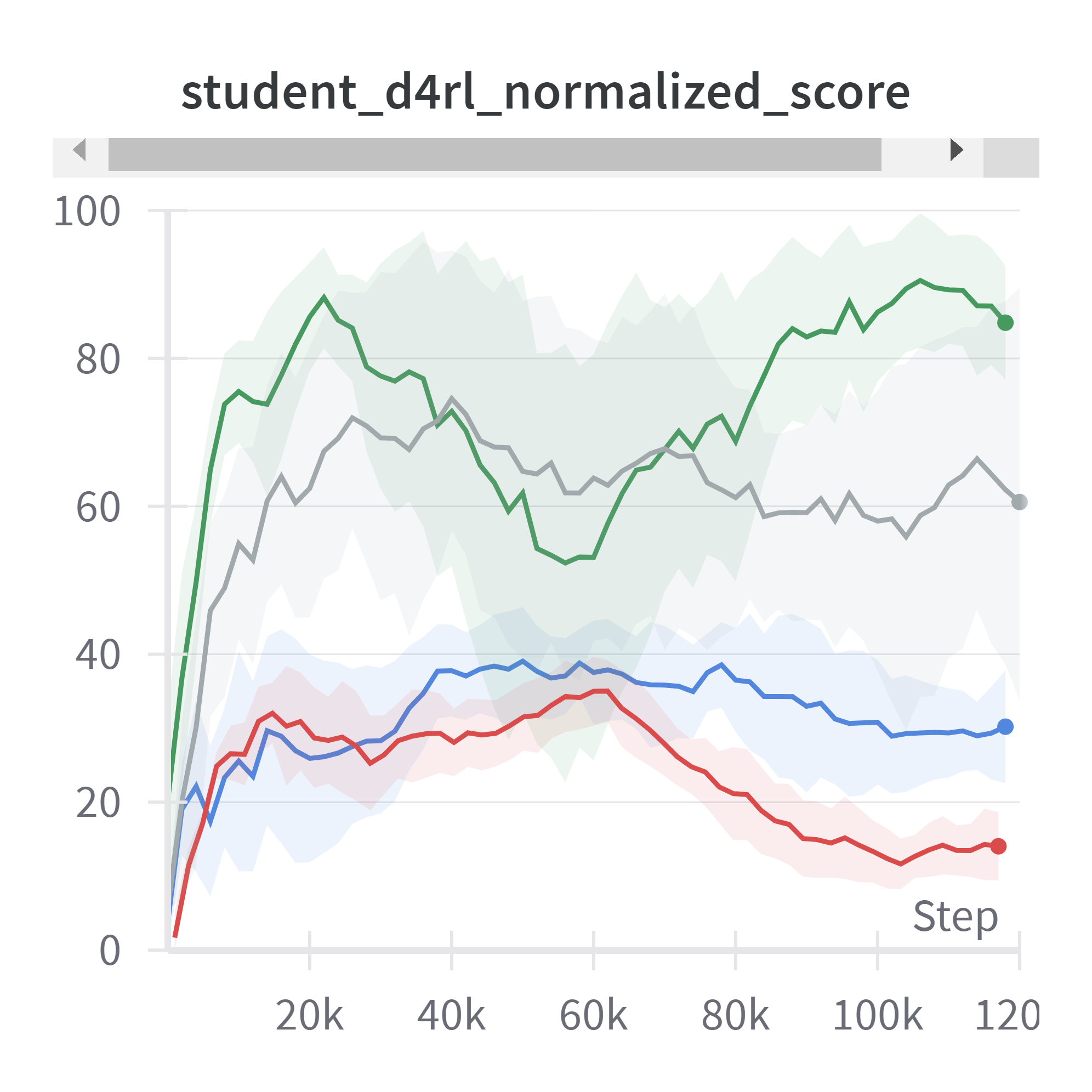}
\end{subfigure}
\hfill 
\begin{subfigure}{0.24\textwidth}
    \includegraphics[trim={0 0 0 12cm},clip,width=\textwidth]{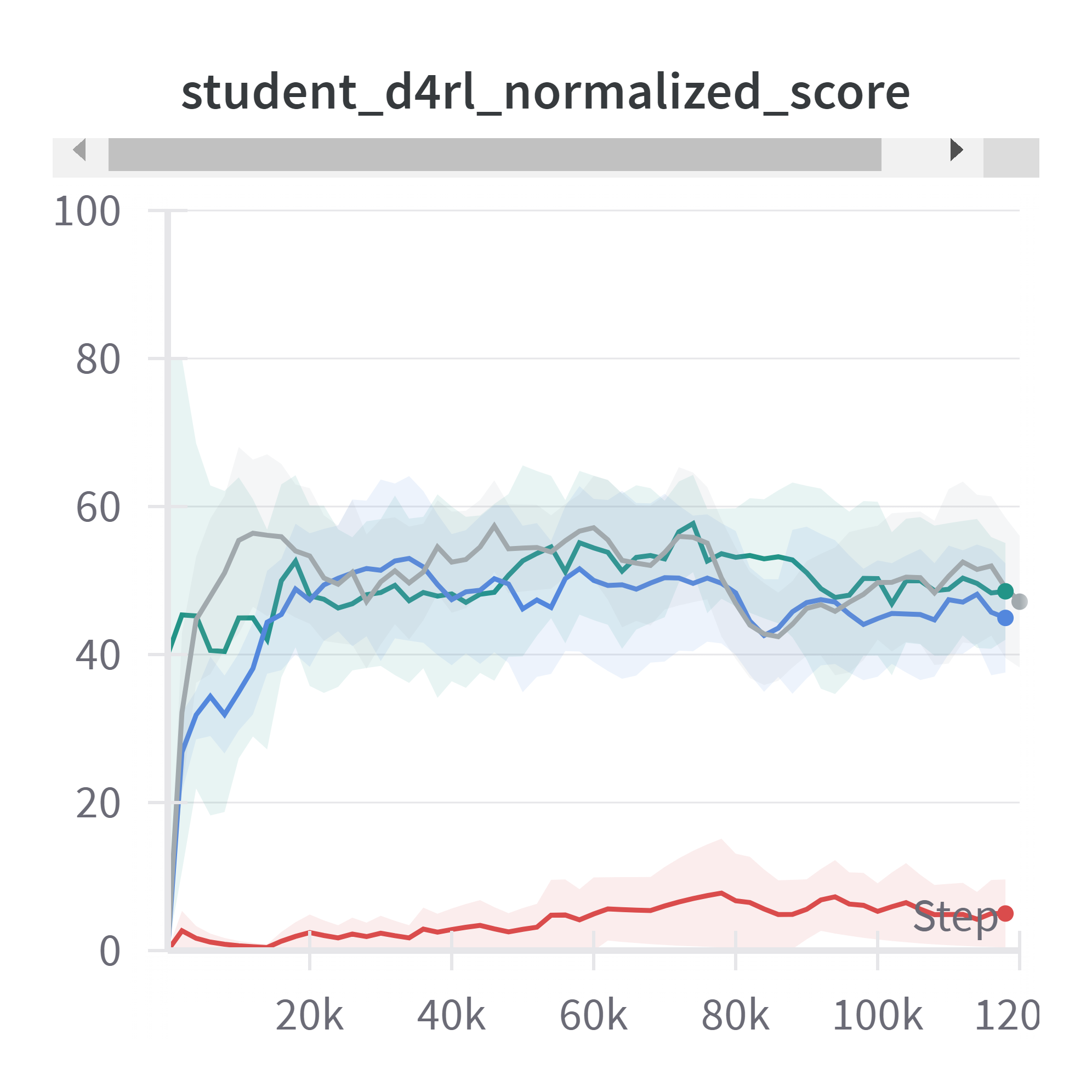}
\end{subfigure}
\vskip -0.1in
\centering
\includegraphics[width=0.6\textwidth,trim={0 11cm 0 0}, clip]{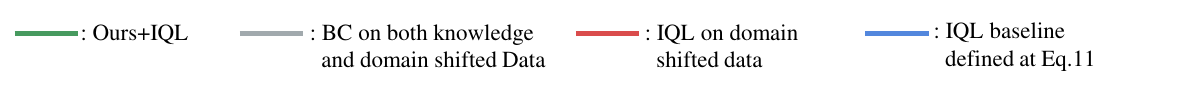}
\captionsetup{skip=1cm} 
\vskip -0.2in
\centering
\caption{Average normalized score on comparative analysis of Ludor built on IQL and baseline models across four environments: HalfCheetah, Hopper, Walker2D, and Antmaze, presented sequentially from left to right. The unlabeled data used is expert data.}
\label{fig:general-iql}
\end{figure*}

\begin{figure*}[ht]
\begin{subfigure}{0.24\textwidth}
    \includegraphics[trim={0 0 0 12cm},clip,width=\textwidth]
    {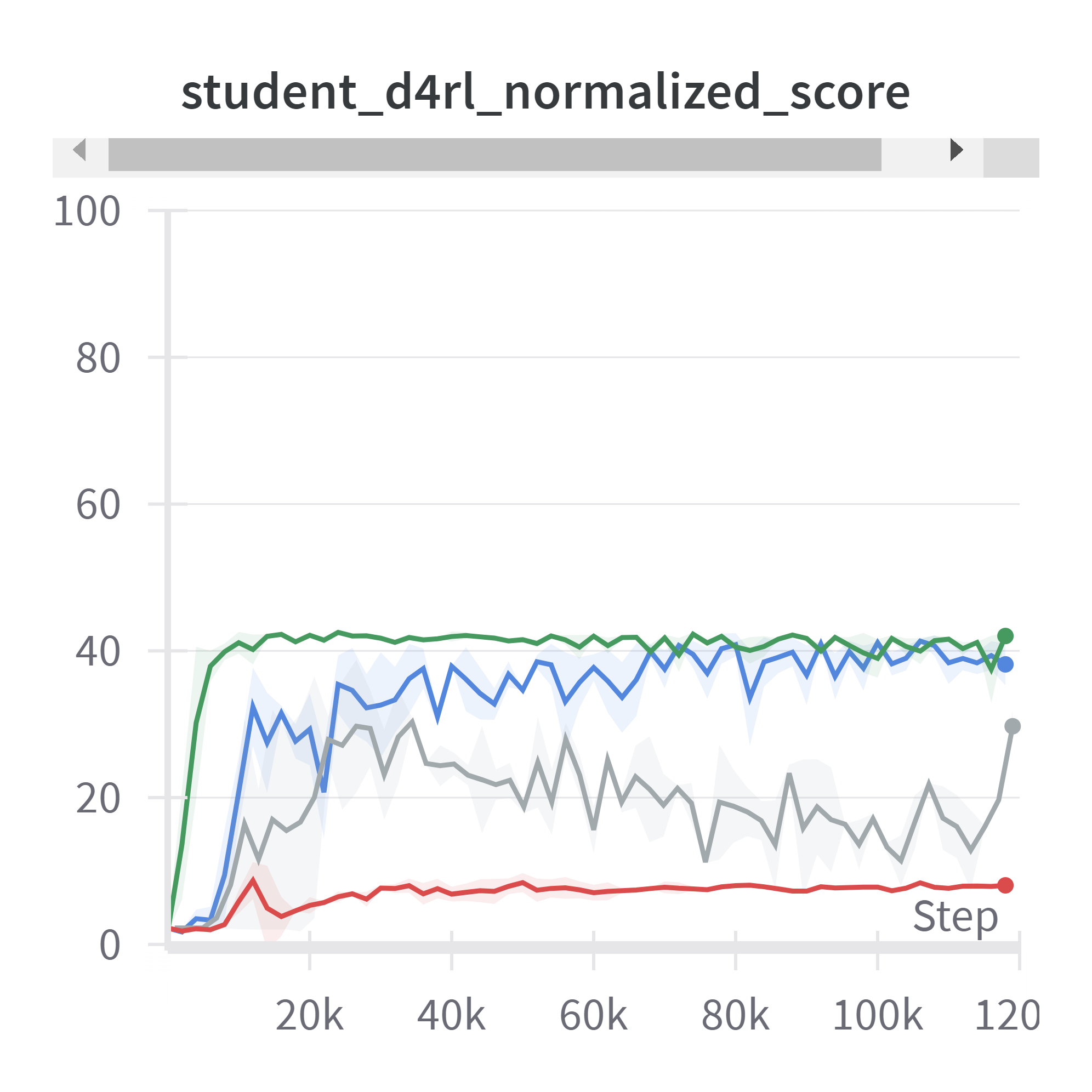}
\end{subfigure}
\hfill 
\begin{subfigure}{0.24\textwidth}
    \includegraphics[trim={0 0 0 14cm},clip,width=\textwidth]{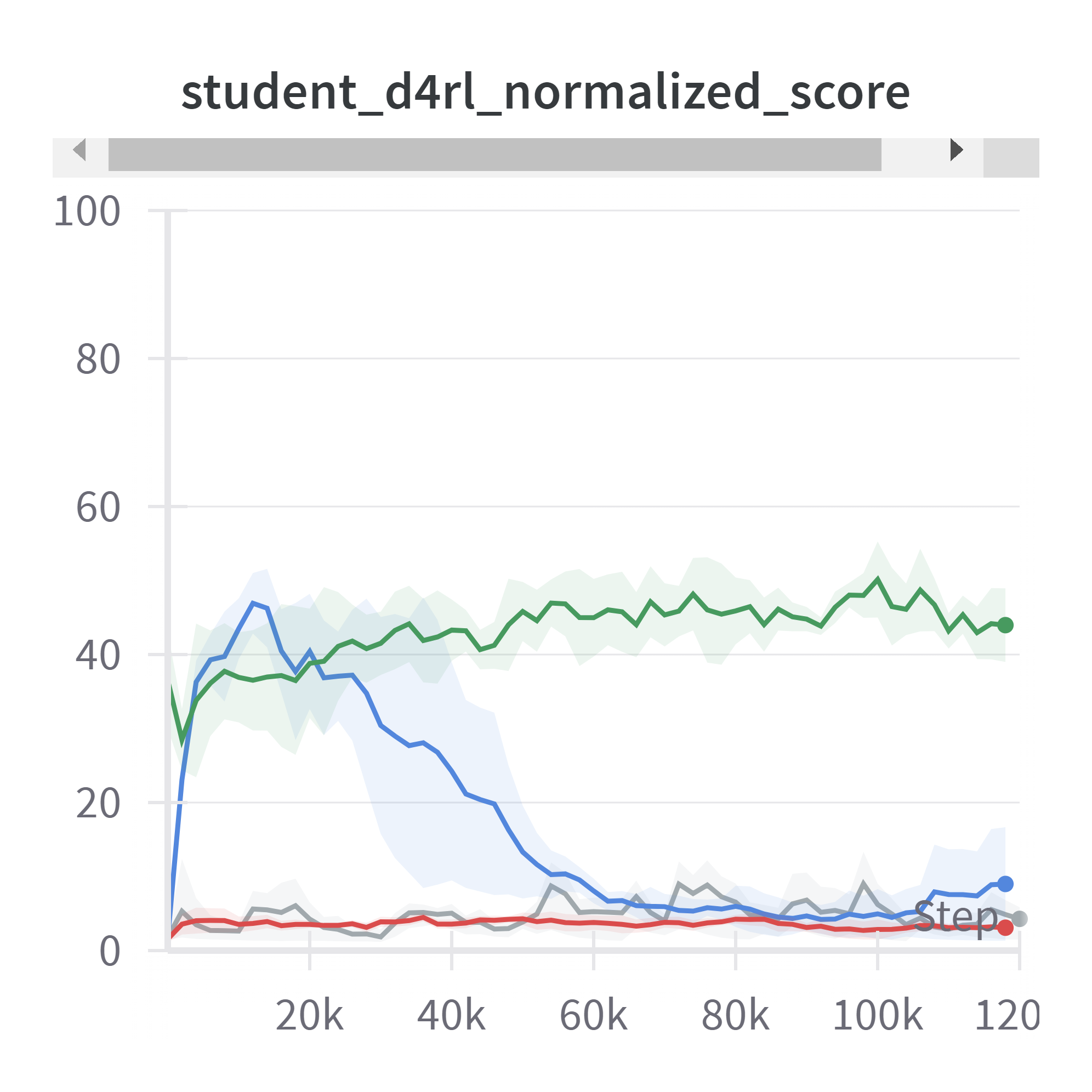}
\end{subfigure}
\hfill 
\begin{subfigure}{0.24\textwidth}
    \includegraphics[trim={0 0 0 12cm},clip,width=\textwidth]{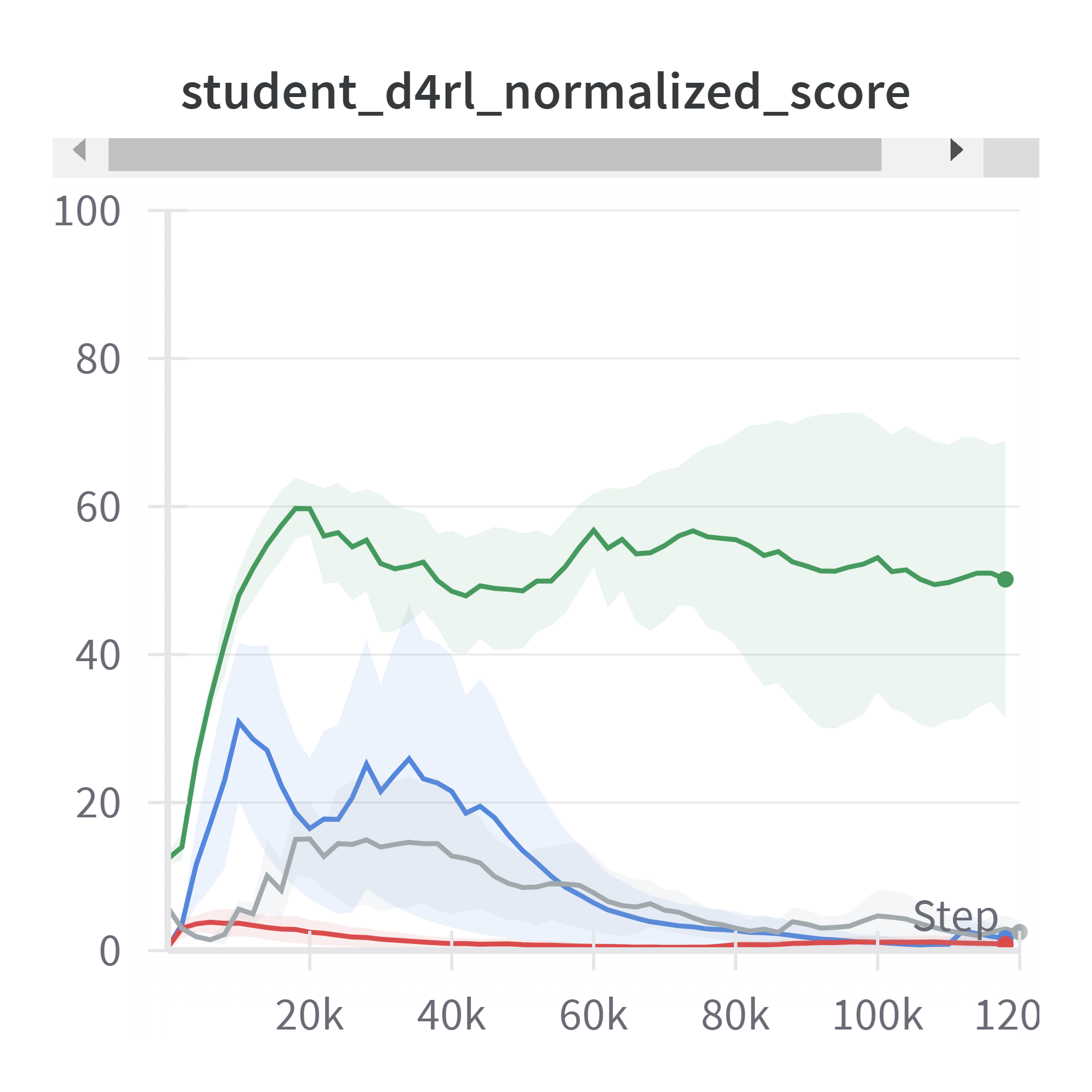}
\end{subfigure}
\hfill 
\begin{subfigure}{0.24\textwidth}
    \includegraphics[trim={0 0 0 12cm},clip,width=\textwidth]{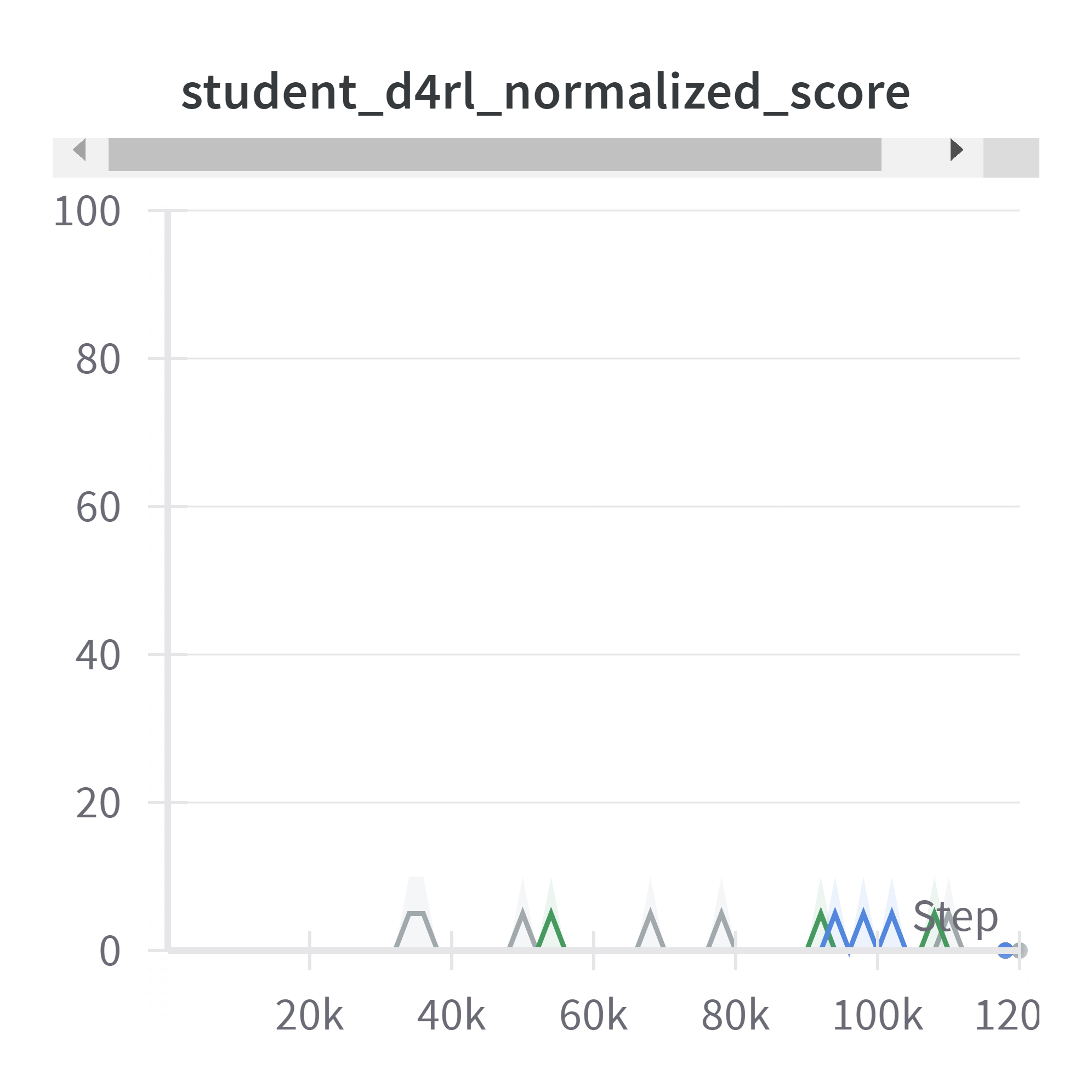}
\end{subfigure}
\centering
\includegraphics[width=0.6\textwidth,trim={0 11cm 0 0}, clip]{figure/TD3BC_caption.pdf}
\centering

\caption{Average normalized score on comparative analysis of Ludor built on TD3BC and baseline models across four environments: HalfCheetah, Hopper, Walker2D, and Antmaze, presented sequentially from left to right. The unlabeled data used is medium data.}

\label{fig:general-td3bc-random}
\end{figure*}

\begin{figure*}[ht]
\begin{subfigure}{0.24\textwidth}
    \includegraphics[trim={0 0 0 12cm},clip,width=\textwidth]{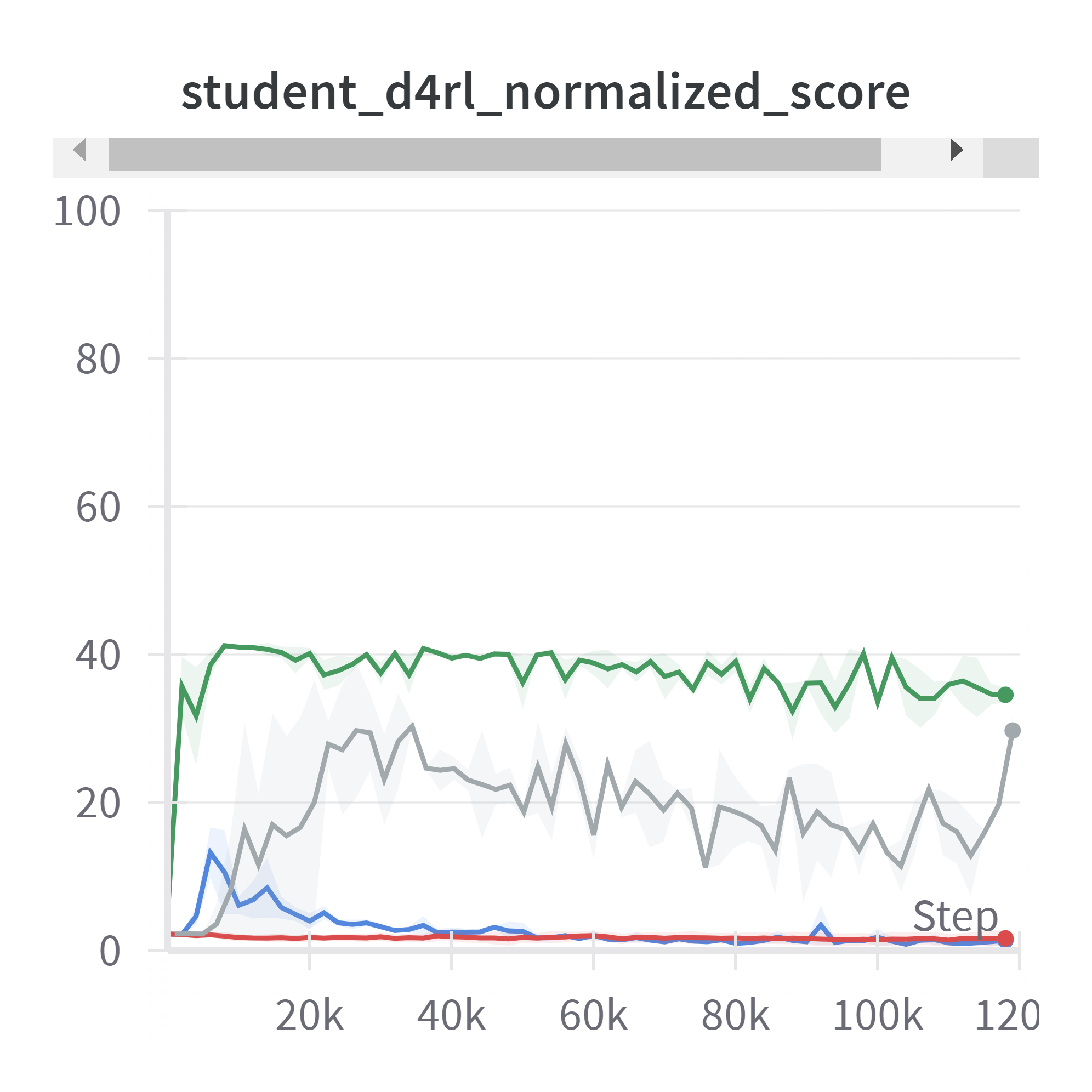}
\end{subfigure}
\hfill 
\begin{subfigure}{0.24\textwidth}
    \includegraphics[trim={0 0 0 12cm},clip,width=\textwidth]{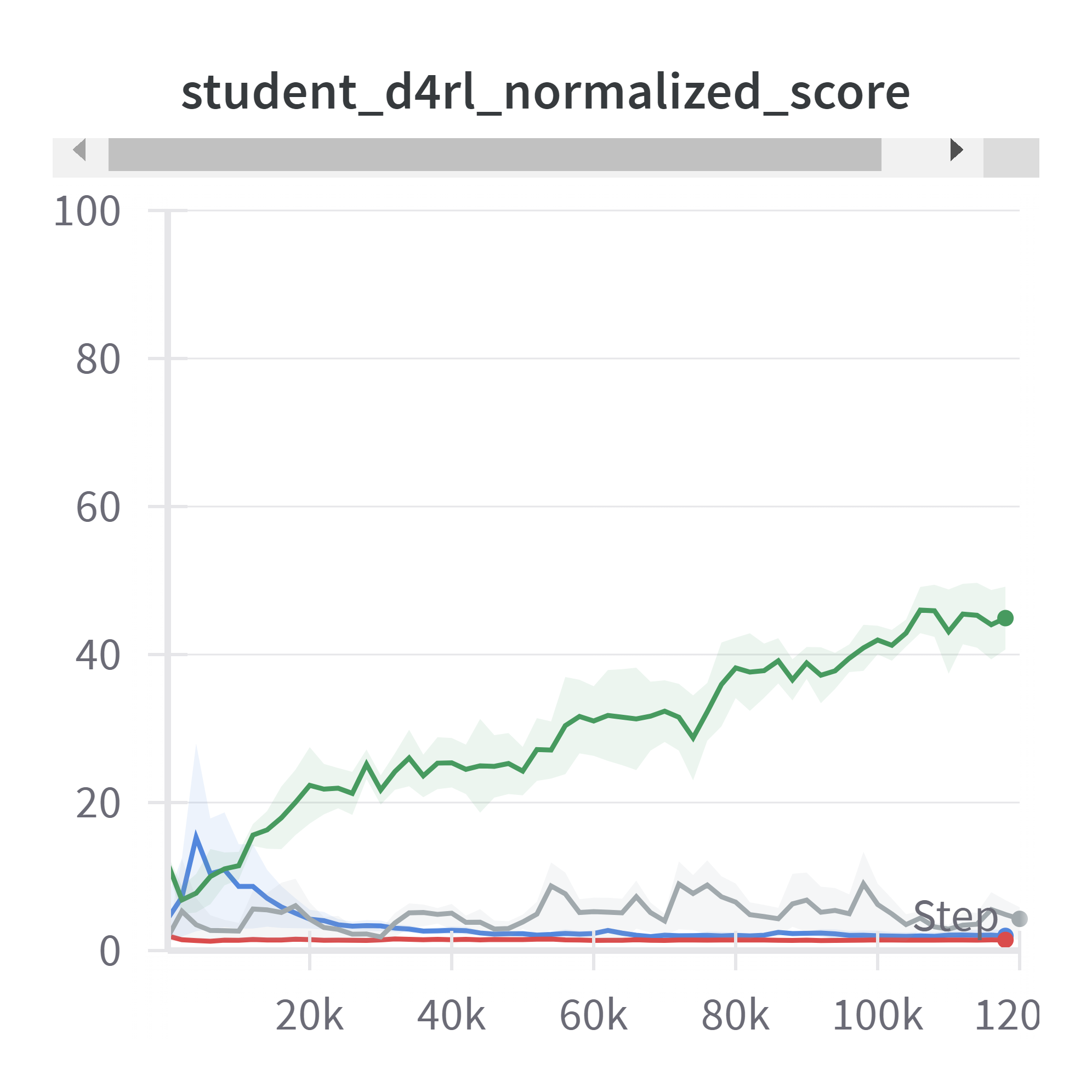}
\end{subfigure}
\hfill 
\begin{subfigure}{0.24\textwidth}
    \includegraphics[trim={0 0 0 12cm},clip,width=\textwidth]{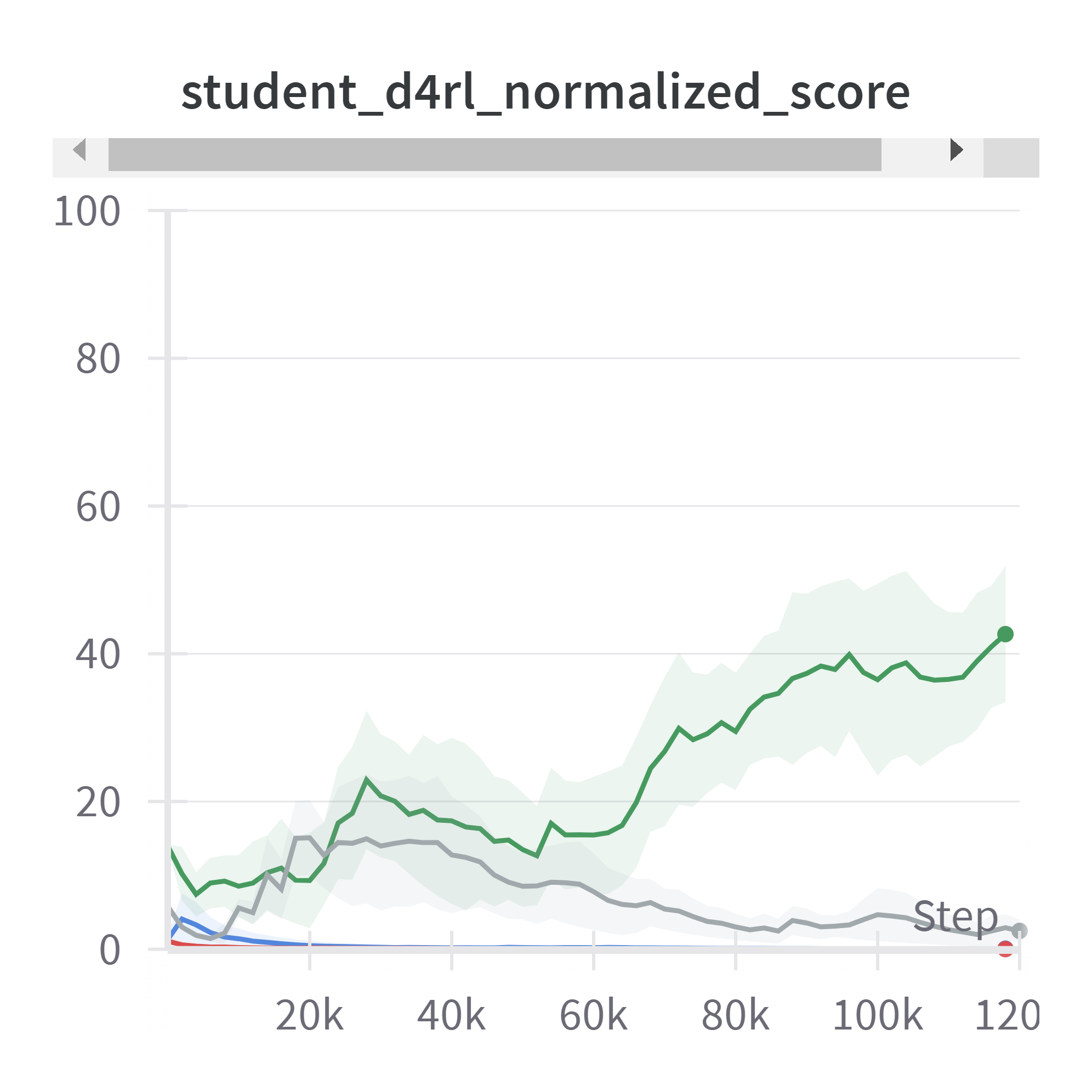}
\end{subfigure}
\hfill 
\begin{subfigure}{0.24\textwidth}
    \includegraphics[trim={0 0 0 12cm},clip,width=\textwidth]{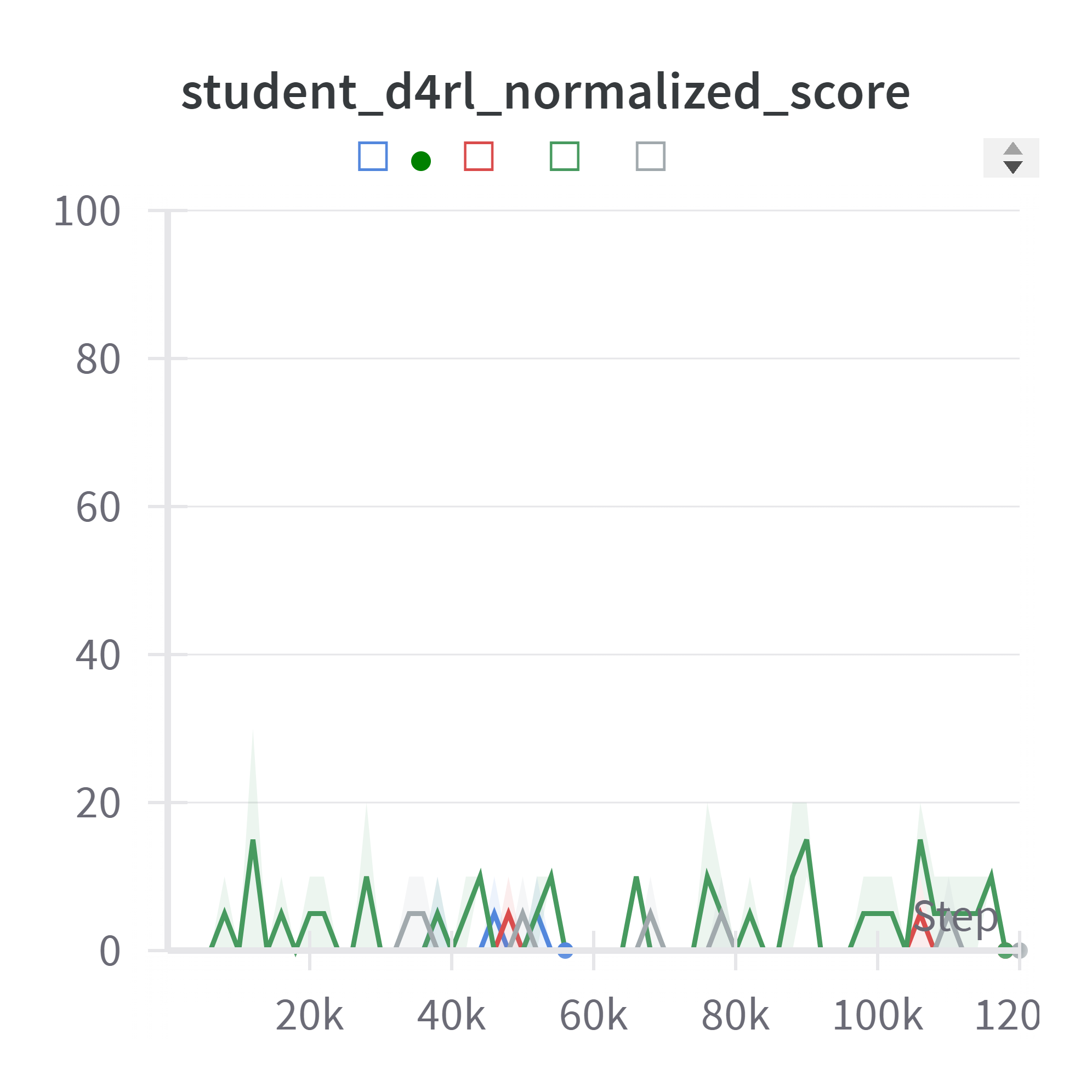}
\end{subfigure}
\vskip -0.1in
\centering
\includegraphics[width=0.6\textwidth,trim={0 11cm 0 0}, clip]{figure/IQL_caption.pdf}
\captionsetup{skip=1cm} 
\vskip -0.2in
\centering
\caption{Average normalized score on comparative analysis of Ludor built on IQL and baseline models across four environments: HalfCheetah, Hopper, Walker2D, and Antmaze, presented sequentially from left to right. The unlabeled data used is medium data.}

\label{fig:general-iql-random}
\end{figure*}

\begin{figure*}[ht]
\begin{subfigure}{0.24\textwidth}
    \includegraphics[trim={0 0 0 12cm},clip,width=\textwidth]{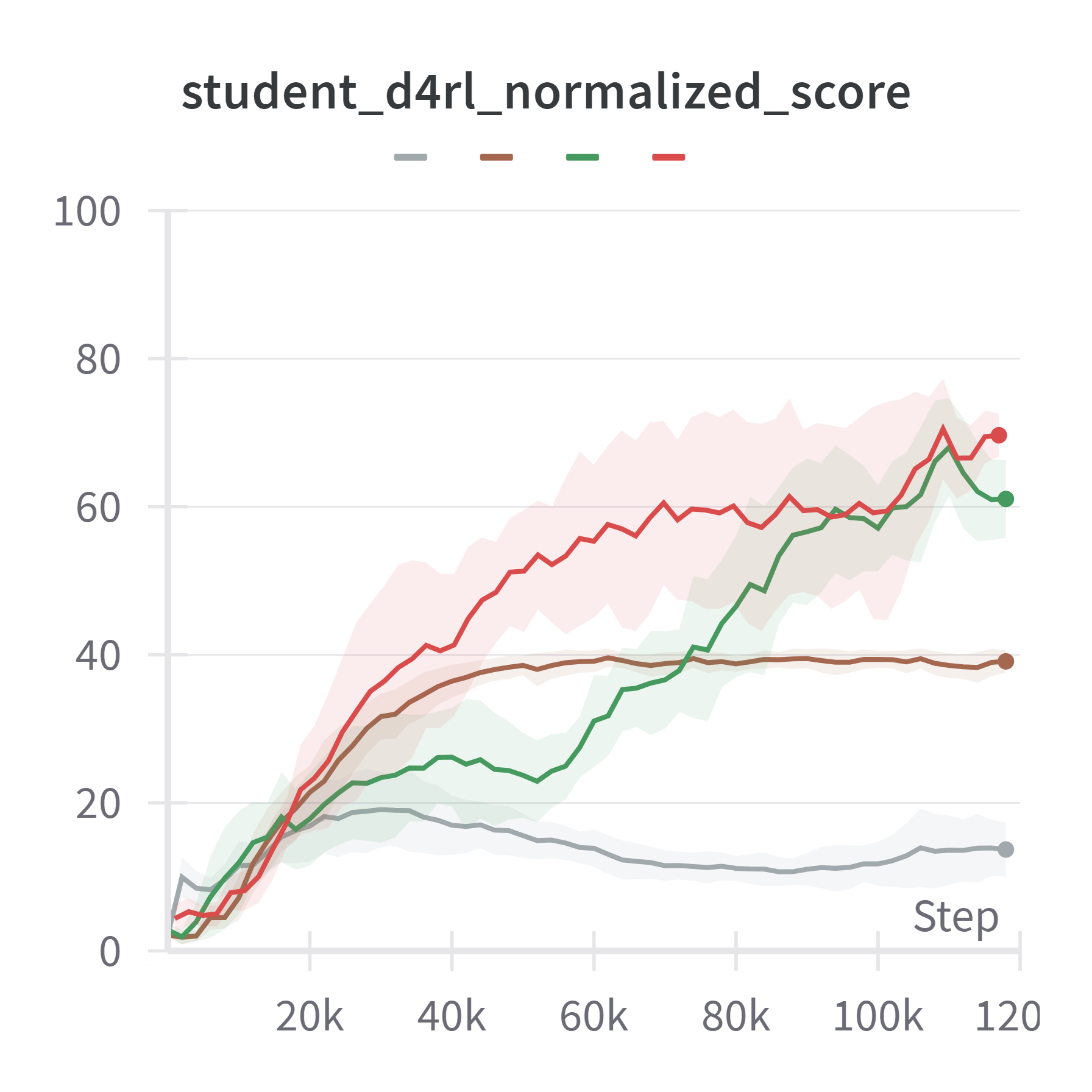}
\end{subfigure}
\hfill 
\begin{subfigure}{0.24\textwidth}
    \includegraphics[trim={0 0 0 12cm},clip,width=\textwidth]{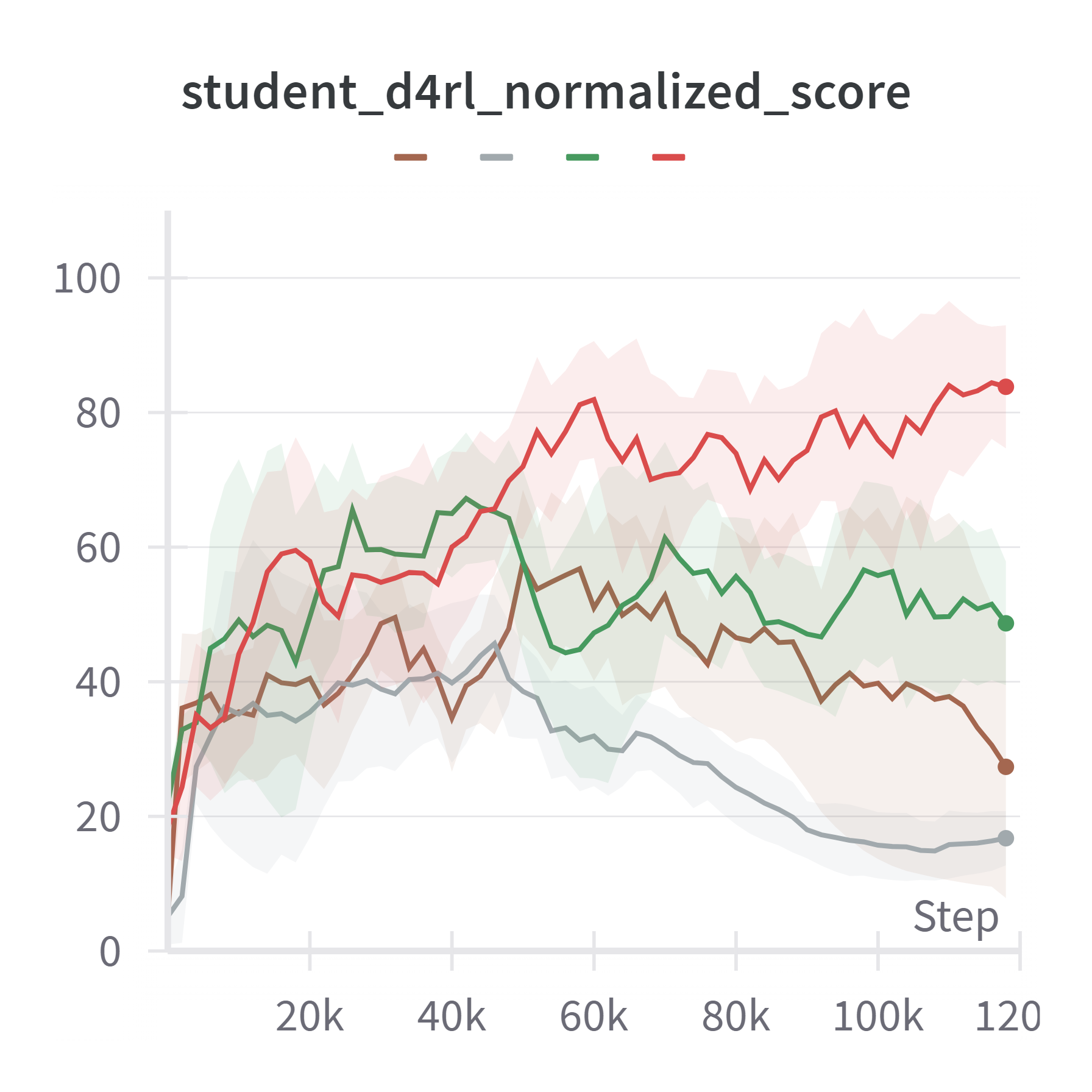}
\end{subfigure}
\hfill 
\begin{subfigure}{0.24\textwidth}
    \includegraphics[trim={0 0 0 12cm},clip,width=\textwidth]{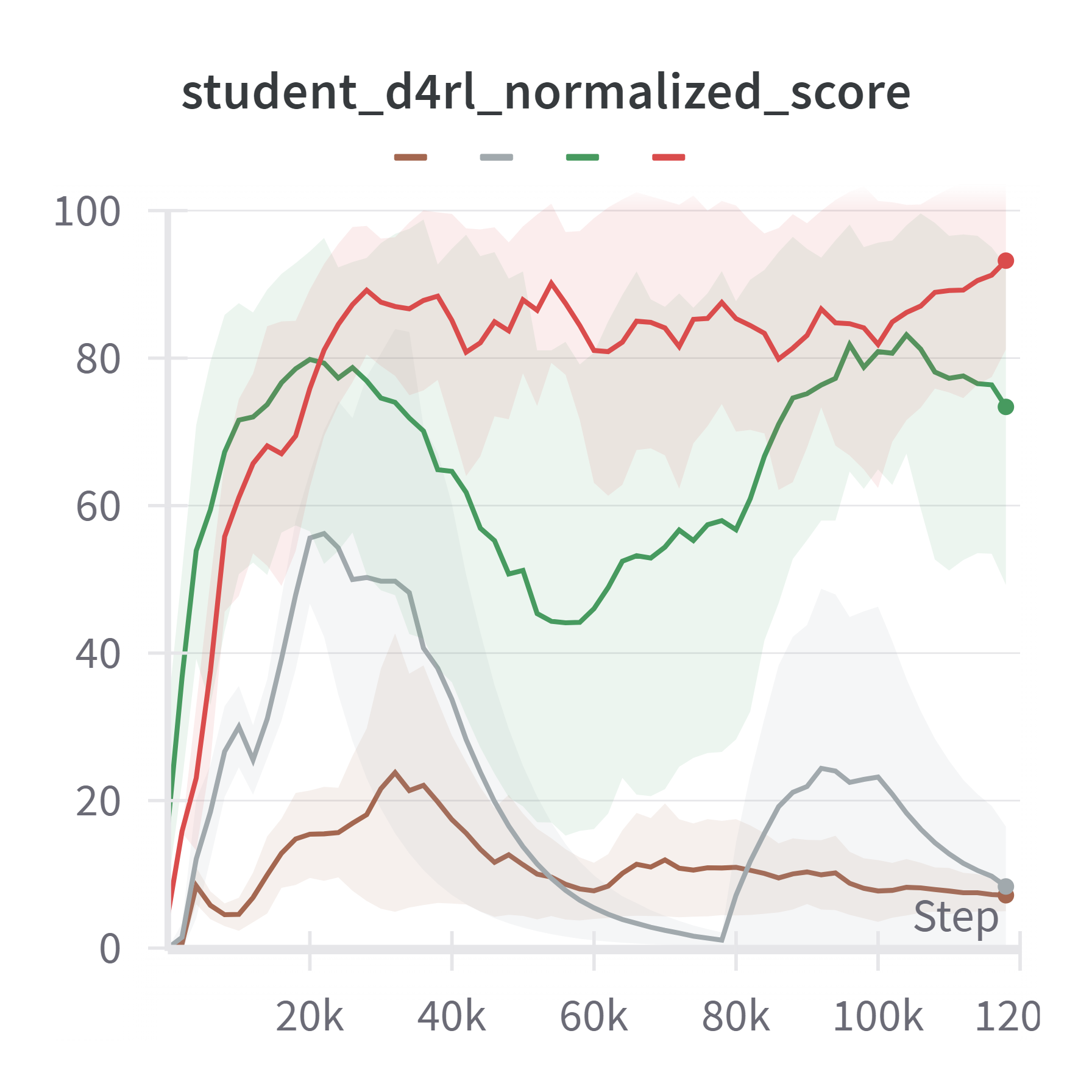}
\end{subfigure}
\hfill 
\begin{subfigure}{0.24\textwidth}
    \includegraphics[trim={0 0 0 12cm},clip,width=\textwidth]{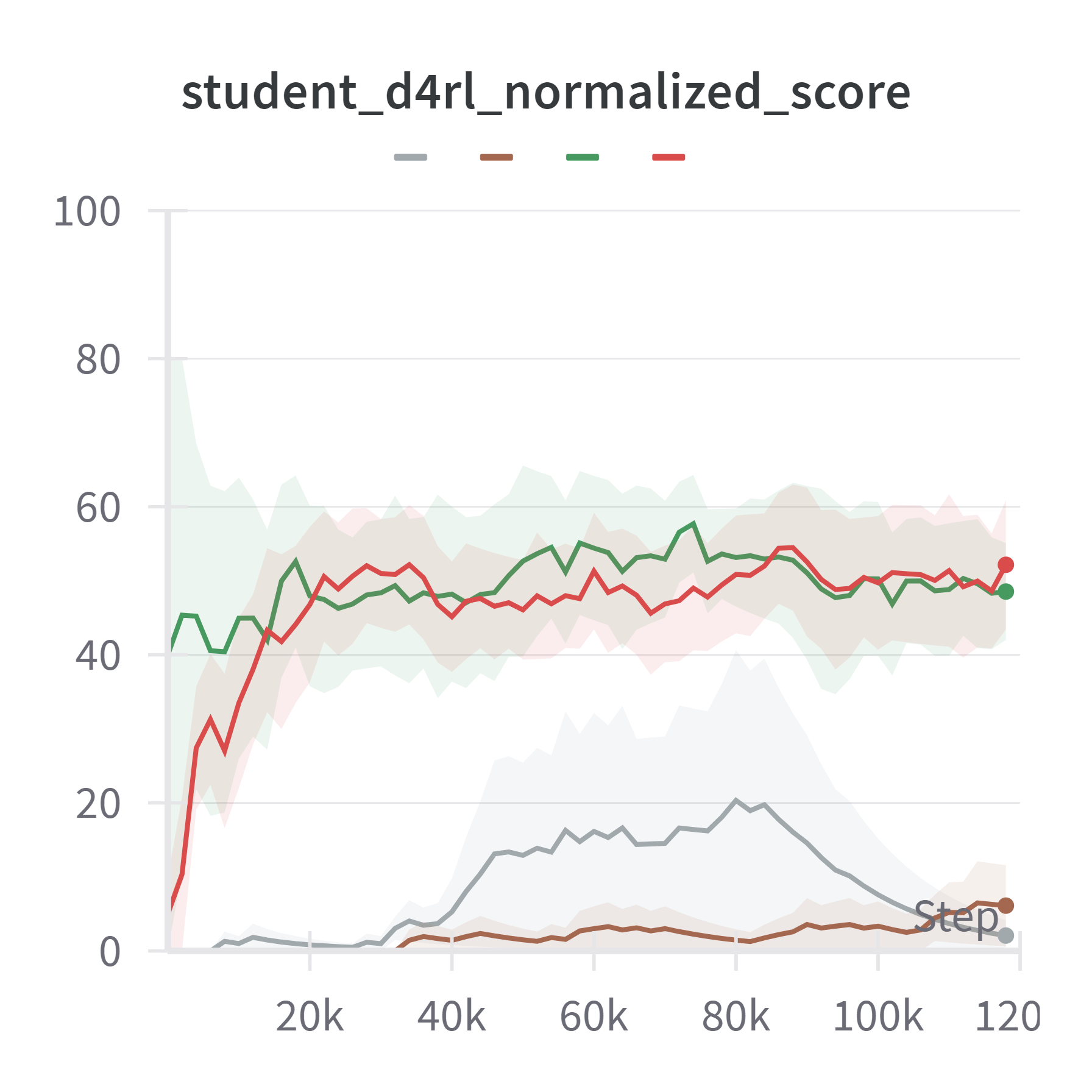}
\end{subfigure}
\vskip -0.1in
\centering
\includegraphics[width=0.6\textwidth,trim={0 11cm 0 0}, clip]{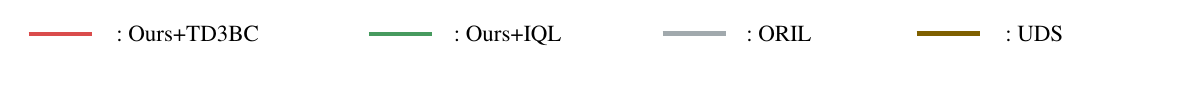}
\captionsetup{skip=1cm} 
\vskip -0.2in
\centering
\caption{Average normalized score on comparative analysis of Ludor and other SoTAs\citep{yu2022leverage, zolna2020offline} across four environments: HalfCheetah, Hopper, Walker2D, and Antmaze, presented sequentially from left to right. The unlabeled data used is expert data.}

\label{fig:sota-medium}
\end{figure*}

\begin{figure*}[ht]
\begin{subfigure}{0.24\textwidth}
    \includegraphics[trim={0 0 0 12cm},clip,width=\textwidth]{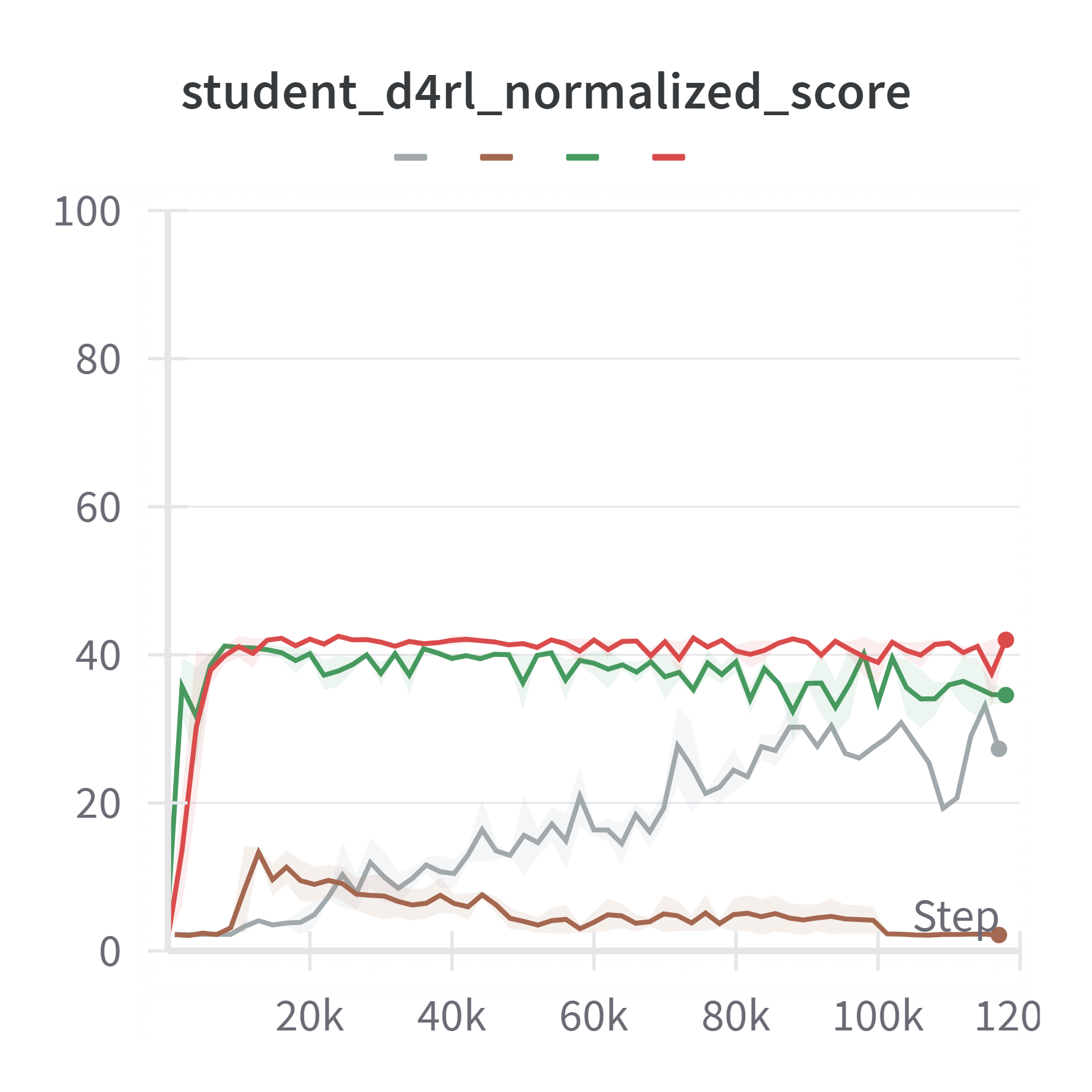}
\end{subfigure}
\hfill 
\begin{subfigure}{0.24\textwidth}
    \includegraphics[trim={0 0 0 12cm},clip,width=\textwidth]{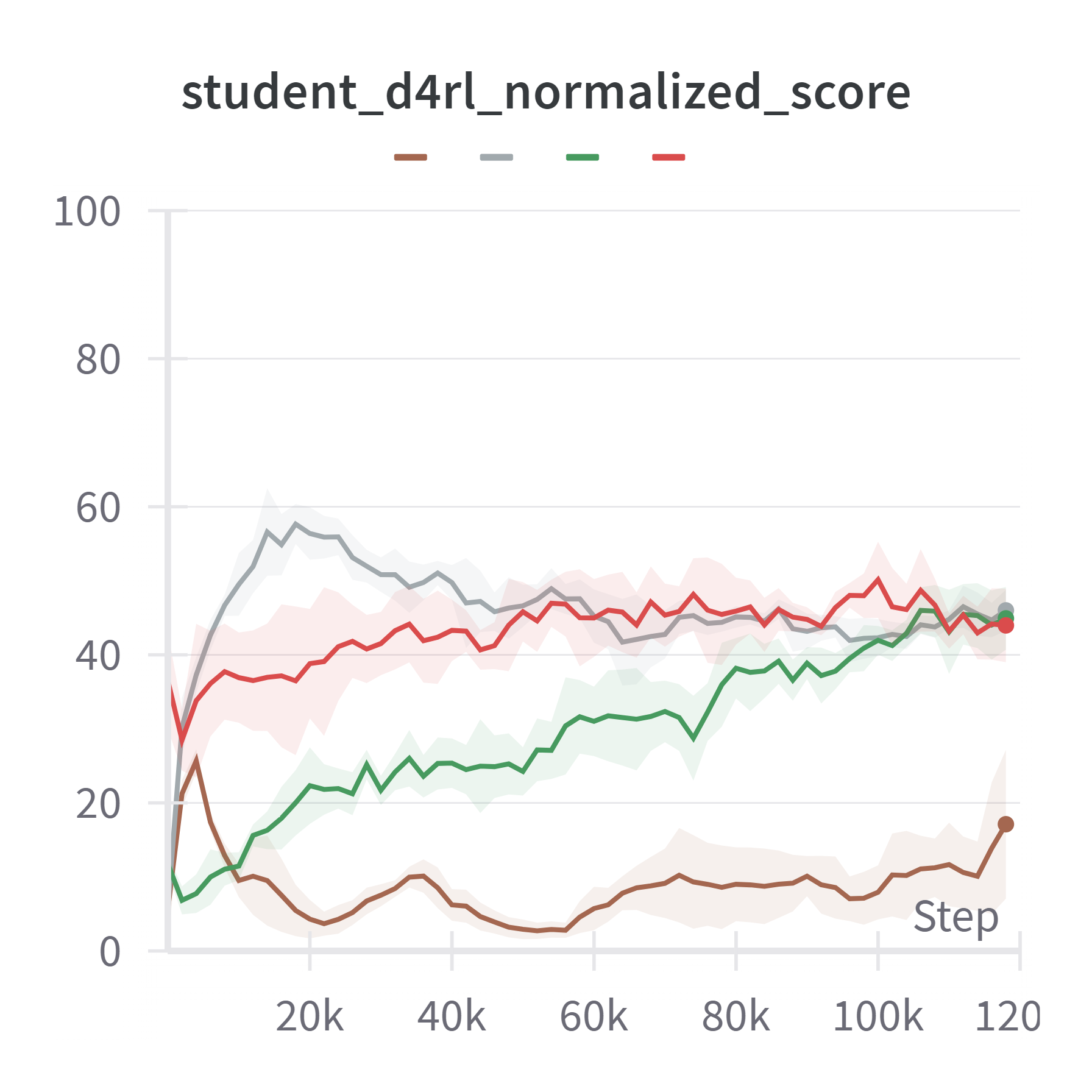}
\end{subfigure}
\hfill 
\begin{subfigure}{0.24\textwidth}
    \includegraphics[trim={0 0 0 12cm},clip,width=\textwidth]{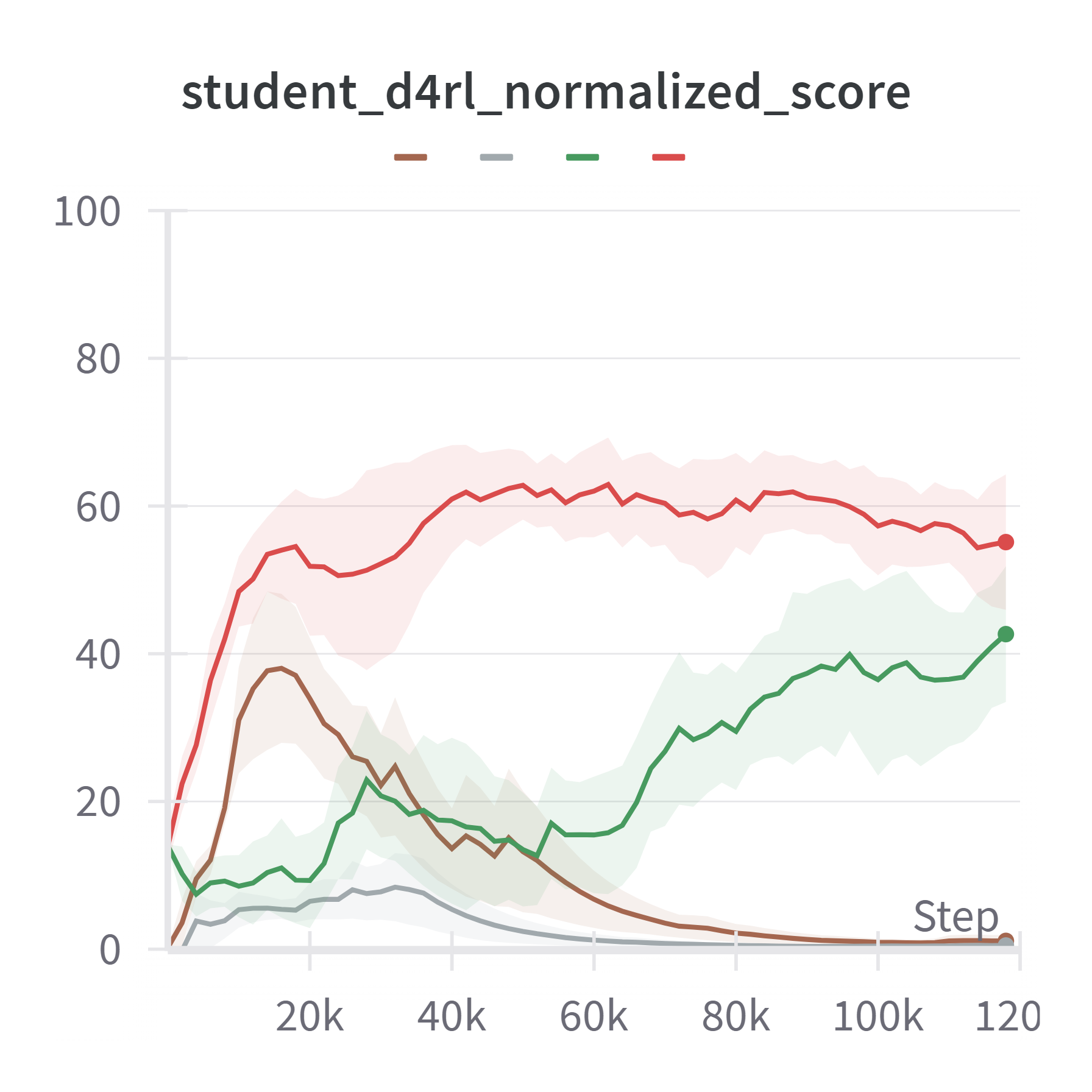}
\end{subfigure}
\hfill 
\begin{subfigure}{0.24\textwidth}
    \includegraphics[trim={0 0 0 12cm},clip,width=\textwidth]{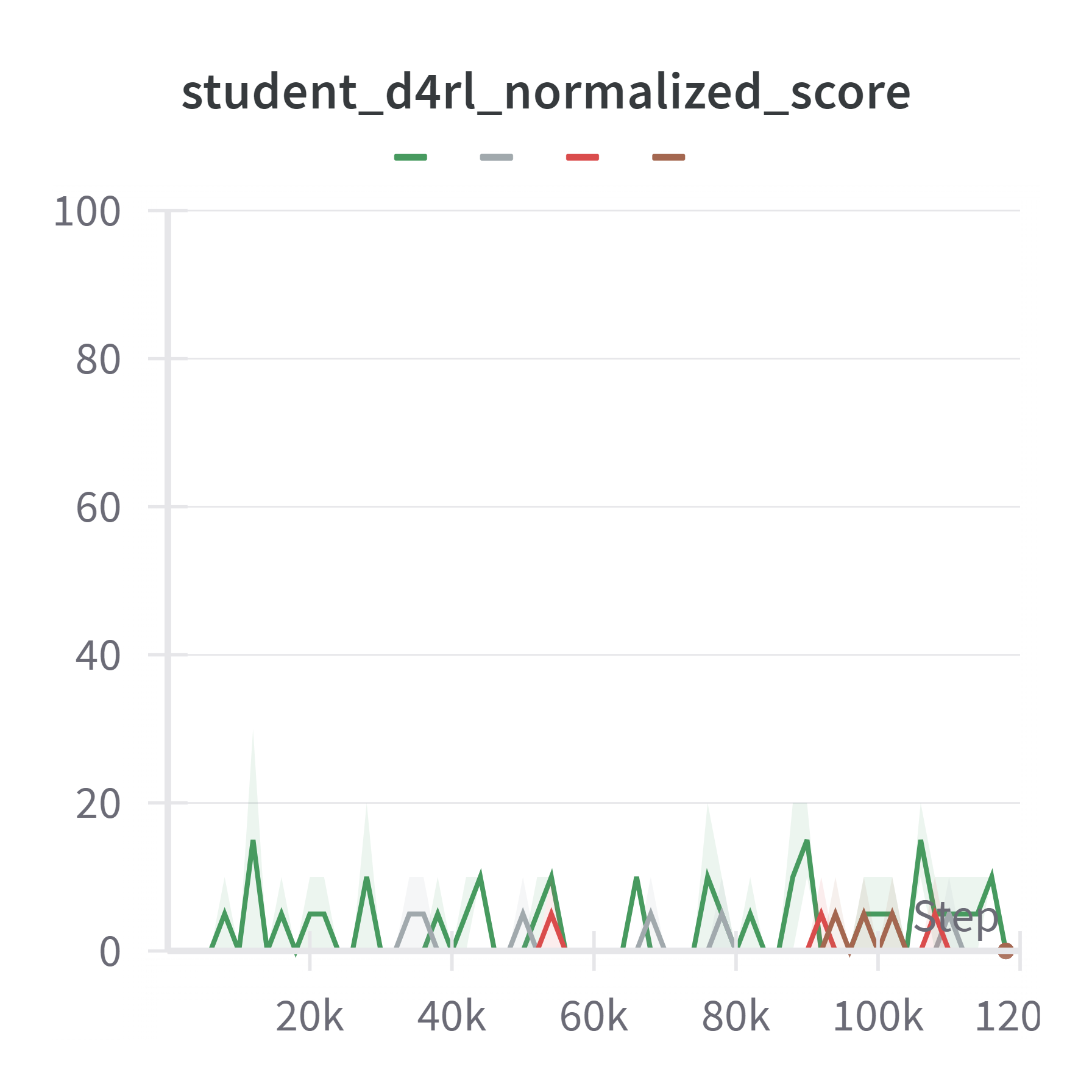}
\end{subfigure}
\vskip -0.1in
\centering
\includegraphics[width=0.6\textwidth,trim={0 11cm 0 0}, clip]{figure/medium-SoTA.pdf}
\captionsetup{skip=1cm} 
\vskip -0.2in
\centering
\caption{Average normalized score on comparative analysis of Ludor and other SoTAs\citep{yu2022leverage, zolna2020offline} across four environments: HalfCheetah, Hopper, Walker2D, and Antmaze, presented sequentially from left to right. The unlabeled data used is medium data.}

\label{fig:sota-}
\end{figure*}

\newpage
\onecolumn
\section{Teacher Unlabeled data Coverage Ratio and Removal Ratio of OOD Creation}
\label{app::robustness}

This section presents evaluation curves obtained under various teacher unlabeled data coverage settings, applied in different D4RL environments as referenced in \citep{fu2020d4rl}. We set the unlabeled data ratios at \(100\%\), \(80\%\), and \(60\%\) to demonstrate the extent of coverage of the unlabeled data on the state observation. To obtain different coverage ratios of unlabeled data, we filtered \(1\%\) of the expert dataset by setting thresholds on specific state dimensions in various environments. It was observed that in certain environments, such as HalfCheetah and Antmaze-umaze, the performance of the trained policy significantly deteriorates when the coverage ratio falls below \(60\%\). Generally, there is a direct correlation between the coverage ratio and the performance of the trained policy: a higher coverage ratio tends to yield better performance.

\begin{figure}[ht]
\begin{subfigure}{0.24\textwidth}
    \includegraphics[trim={0 0 0 12cm},clip,width=\textwidth]{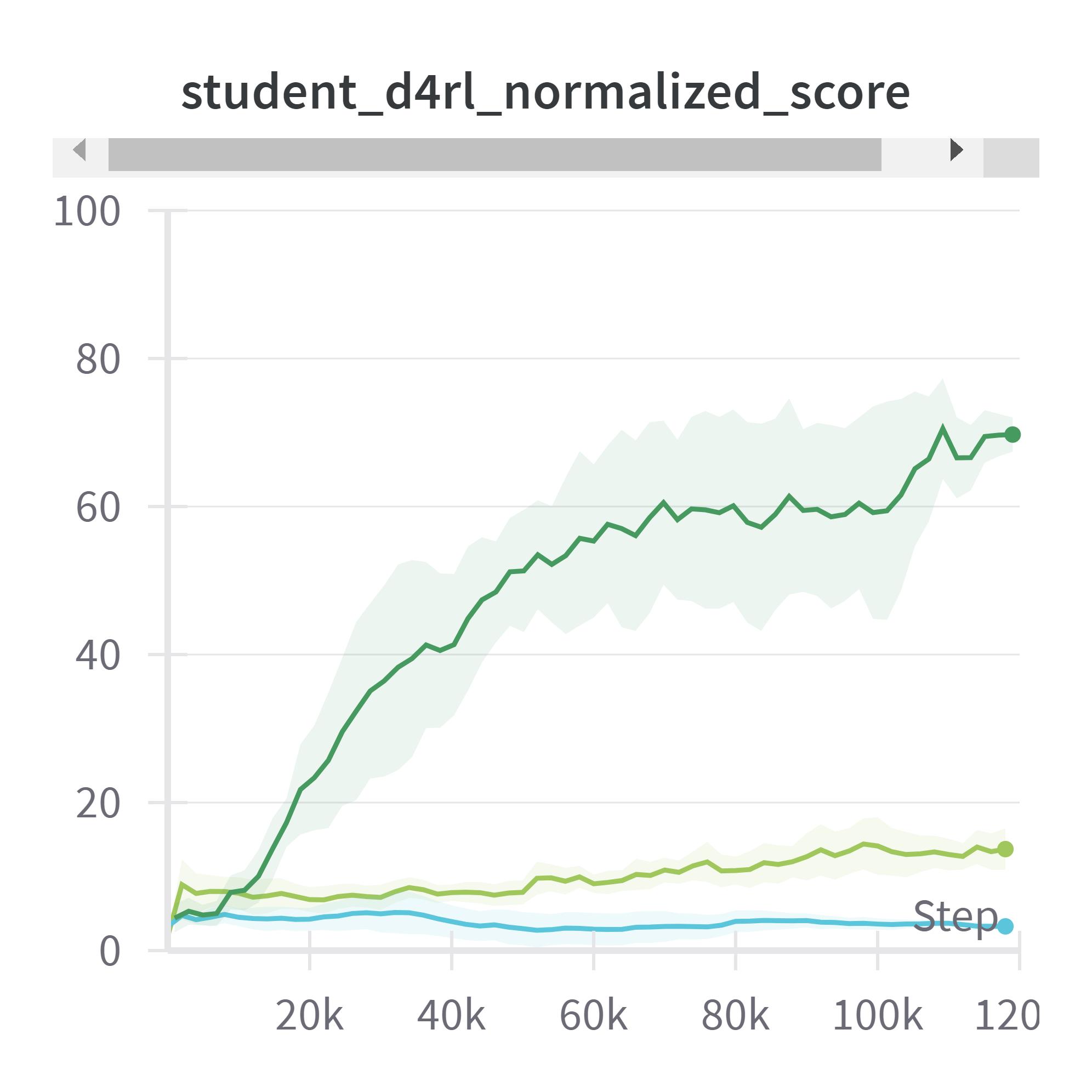}
\end{subfigure}
\hfill 
\begin{subfigure}{0.24\textwidth}
    \includegraphics[trim={0 0 0 13cm},clip,width=\textwidth]{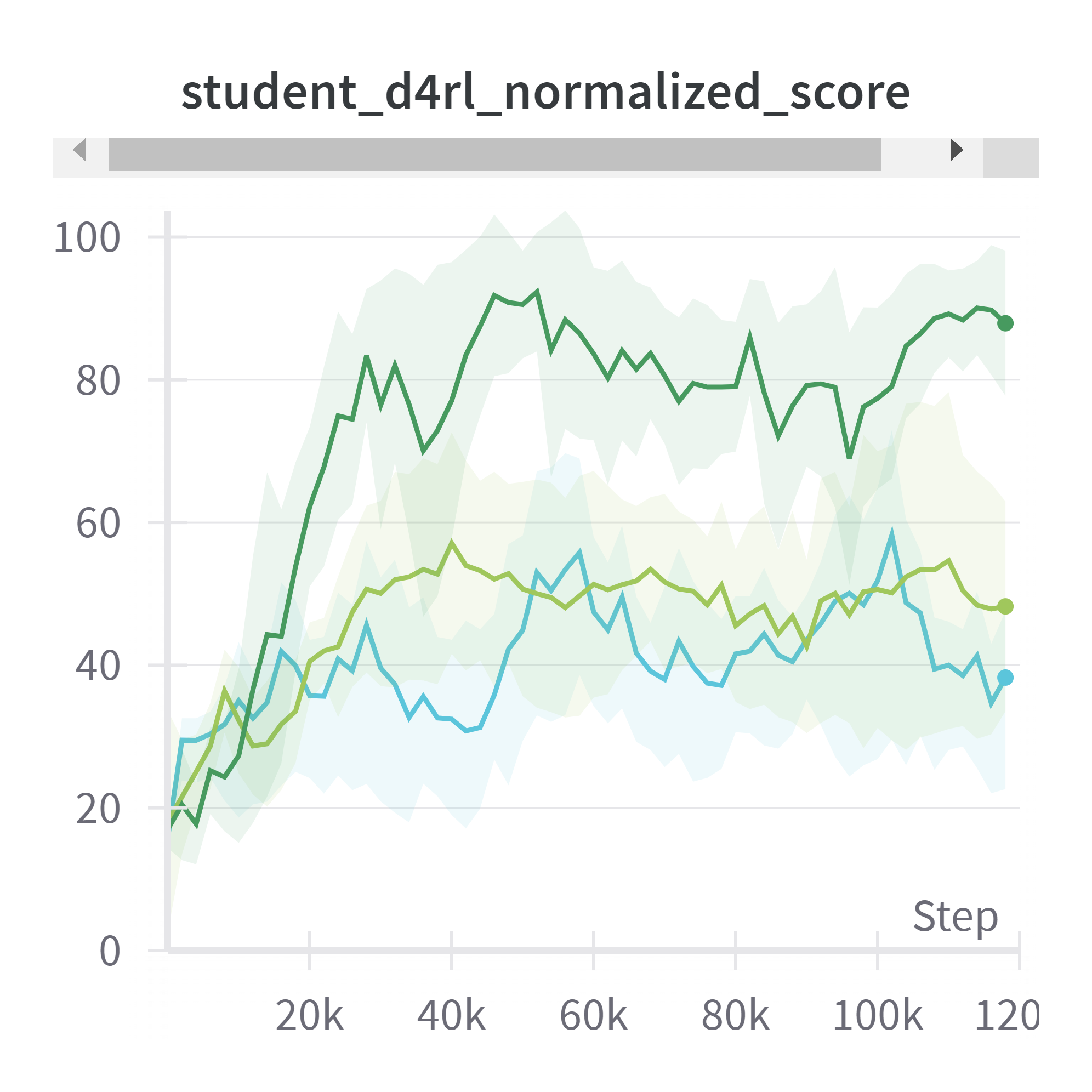}
\end{subfigure}
\hfill 
\begin{subfigure}{0.24\textwidth}
    \includegraphics[trim={0 0 0 12cm},clip,width=\textwidth]{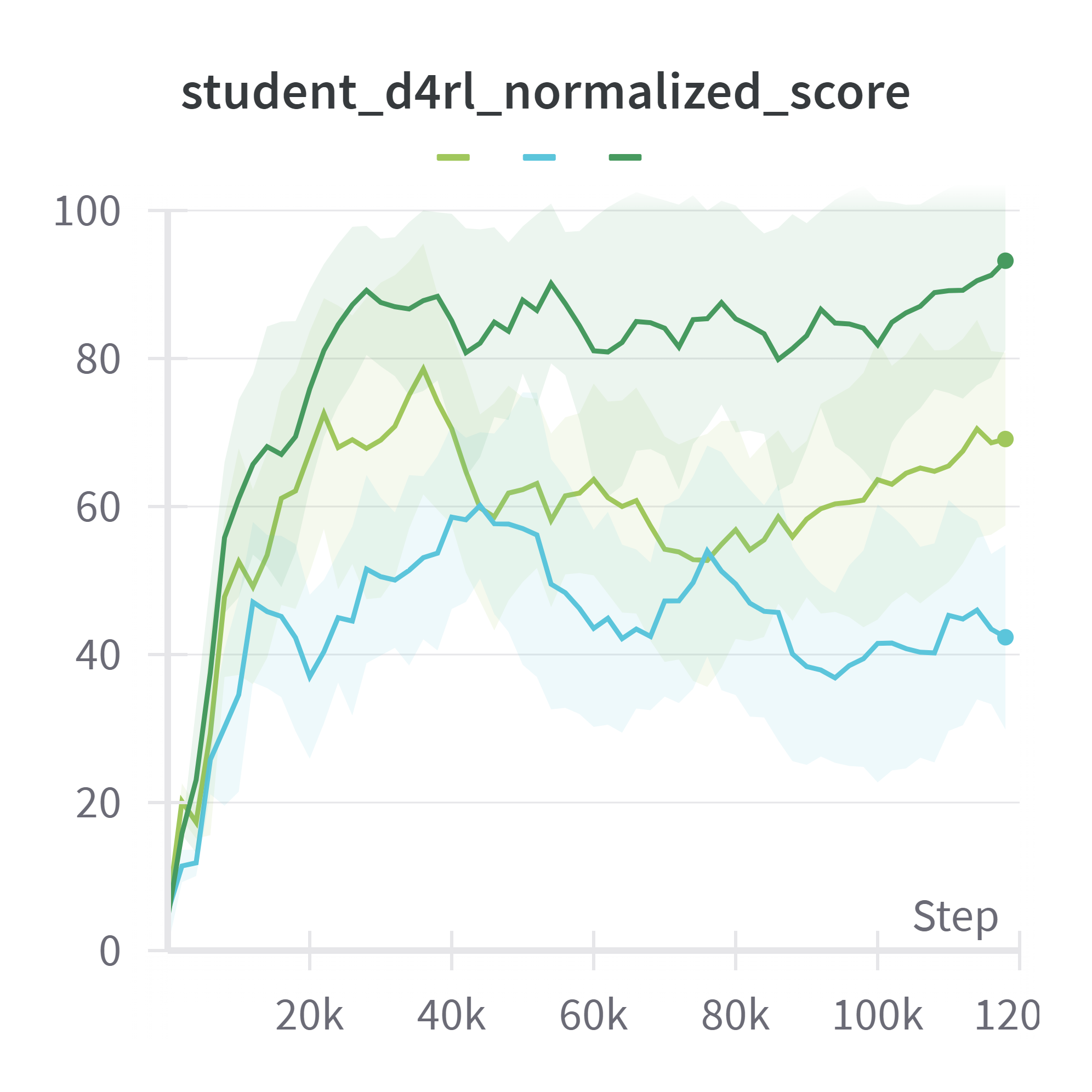}
\end{subfigure}
\hfill 
\begin{subfigure}{0.24\textwidth}
    \includegraphics[trim={0 0 0 12cm},clip,width=\textwidth]{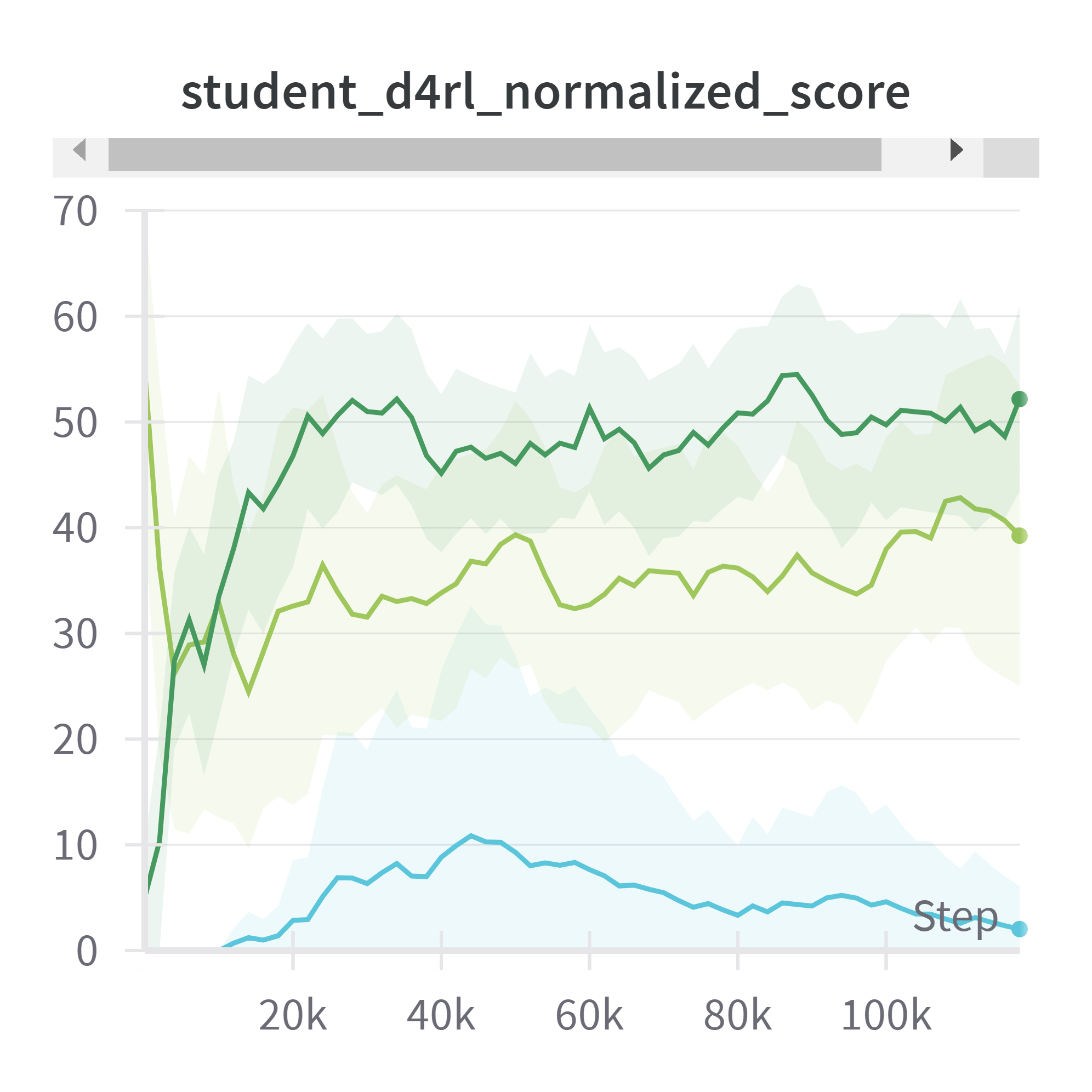}
\end{subfigure}
\vskip -0.1in
\centering
\includegraphics[width=0.6\textwidth,trim={0 11cm 0 0}, clip]{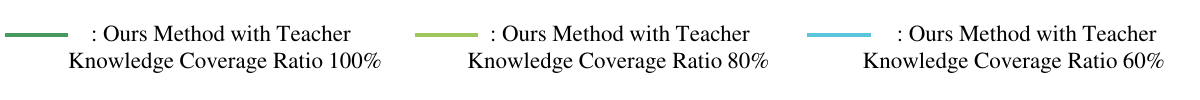}
\captionsetup{skip=1cm} 
\vskip -0.2in
\centering
\caption{Average normalized score depicting the impact of teacher unlabeled data coverage ratio in four environments: HalfCheetah, Hopper, Walker2D, and Antmaze, arranged sequentially from left to right.}

\label{fig:teacher-coverage}
\end{figure}

\begin{figure}[ht]
\begin{subfigure}{0.24\textwidth}
    \includegraphics[trim={0 0 0 12cm},clip,width=\textwidth]{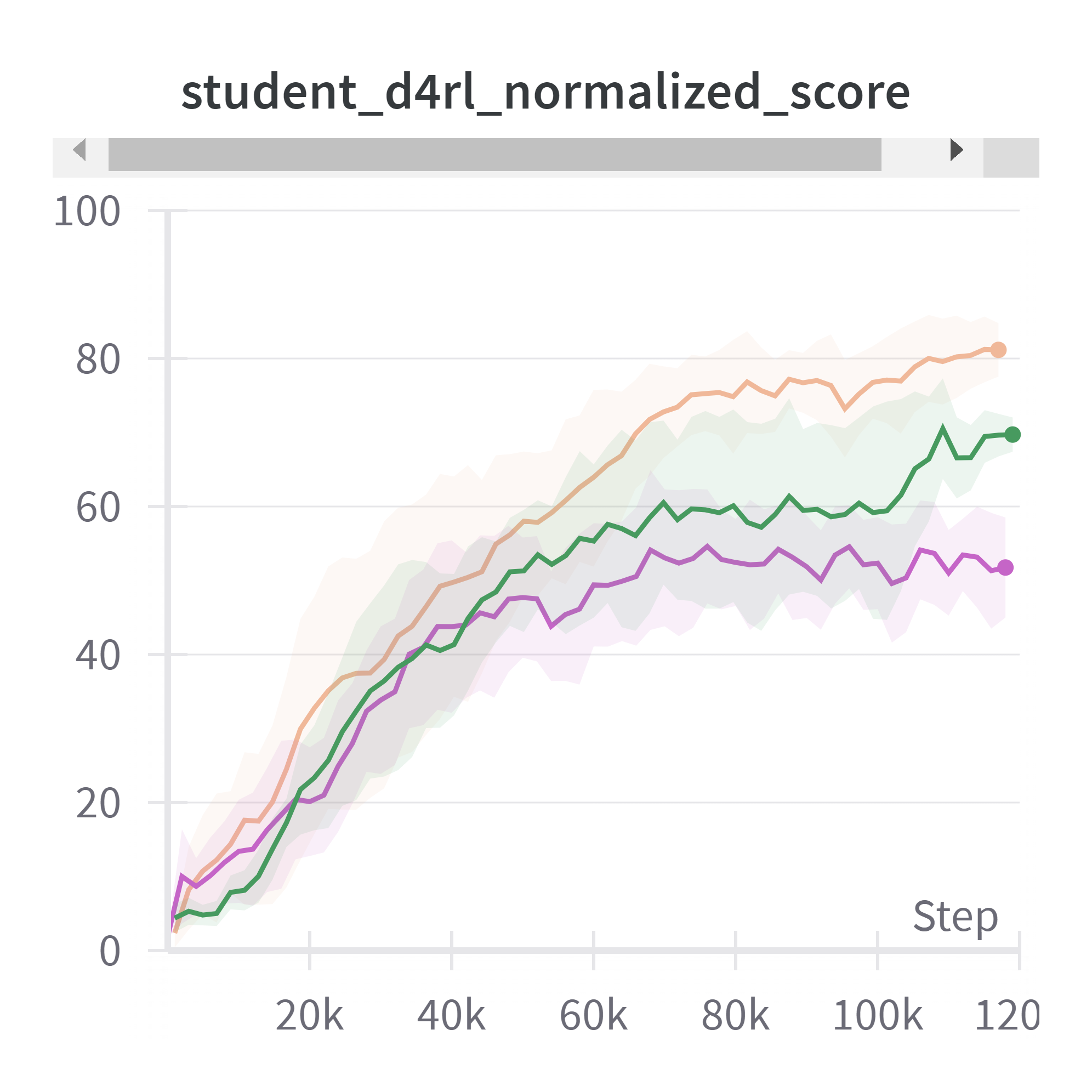}
\end{subfigure}
\hfill 
\begin{subfigure}{0.24\textwidth}
    \includegraphics[trim={0 0 0 13cm},clip,width=\textwidth]{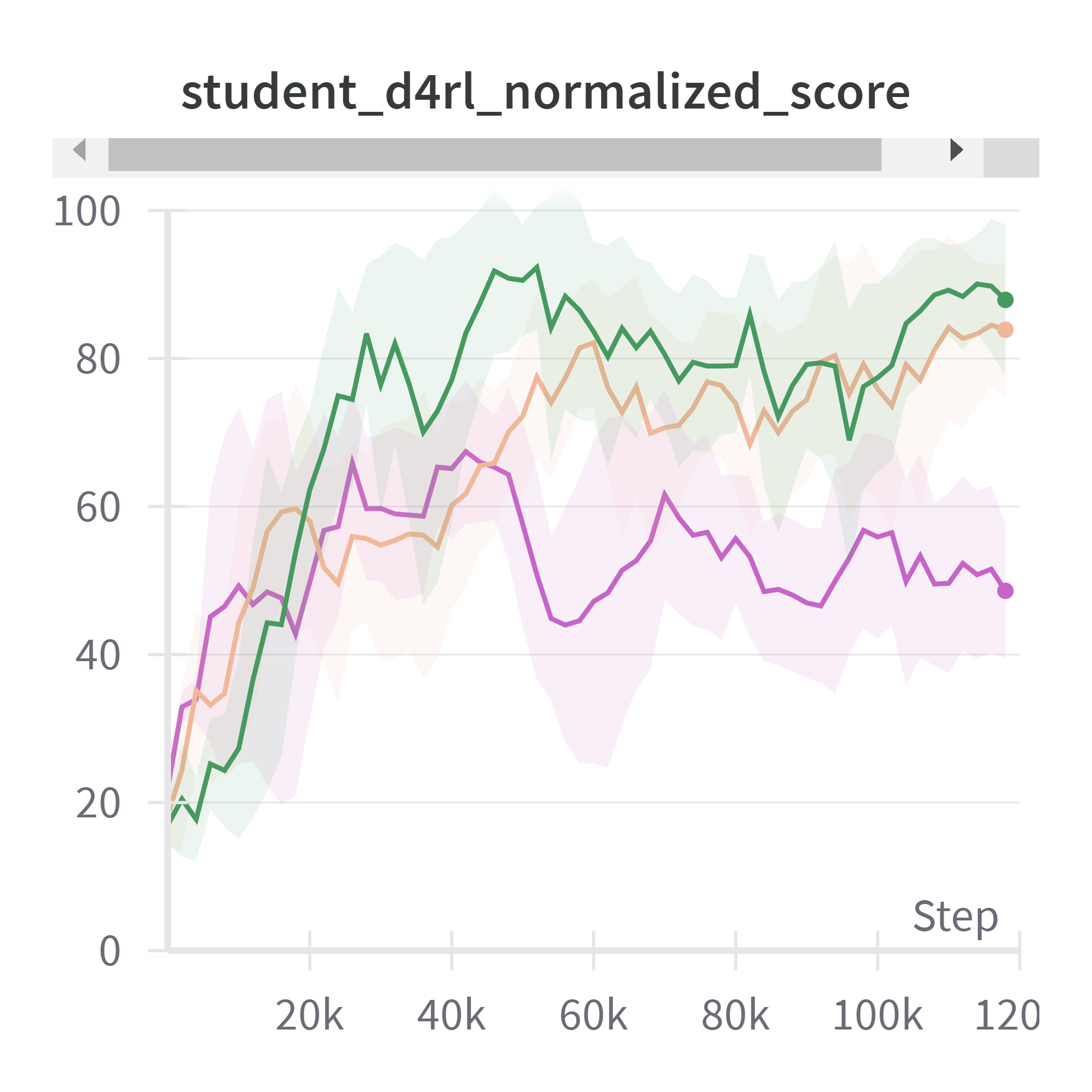}
\end{subfigure}
\hfill 
\begin{subfigure}{0.24\textwidth}
    \includegraphics[trim={0 0 0 12cm},clip,width=\textwidth]{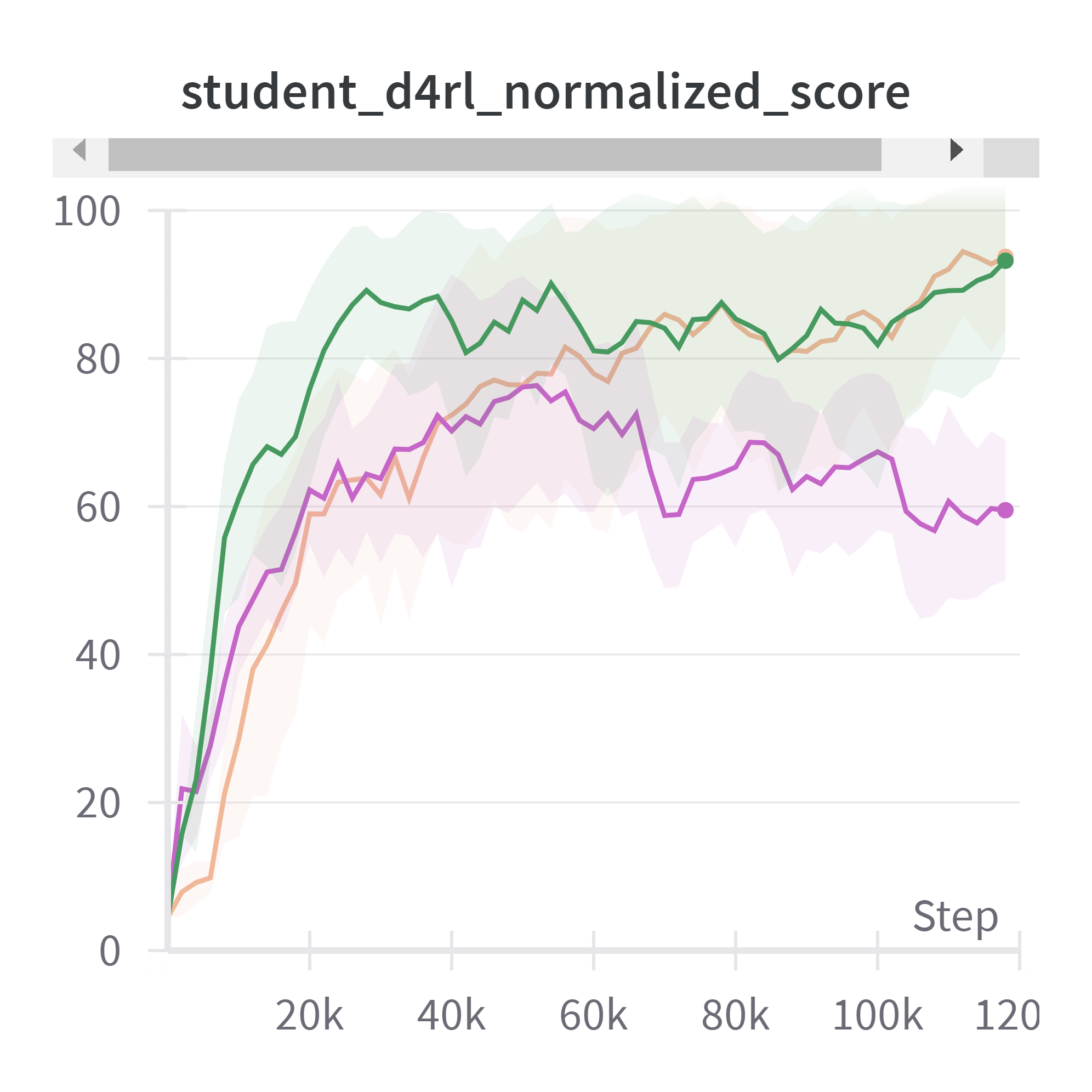}
\end{subfigure}
\hfill 
\begin{subfigure}{0.24\textwidth}
    \includegraphics[trim={0 0 0 12cm},clip,width=\textwidth]{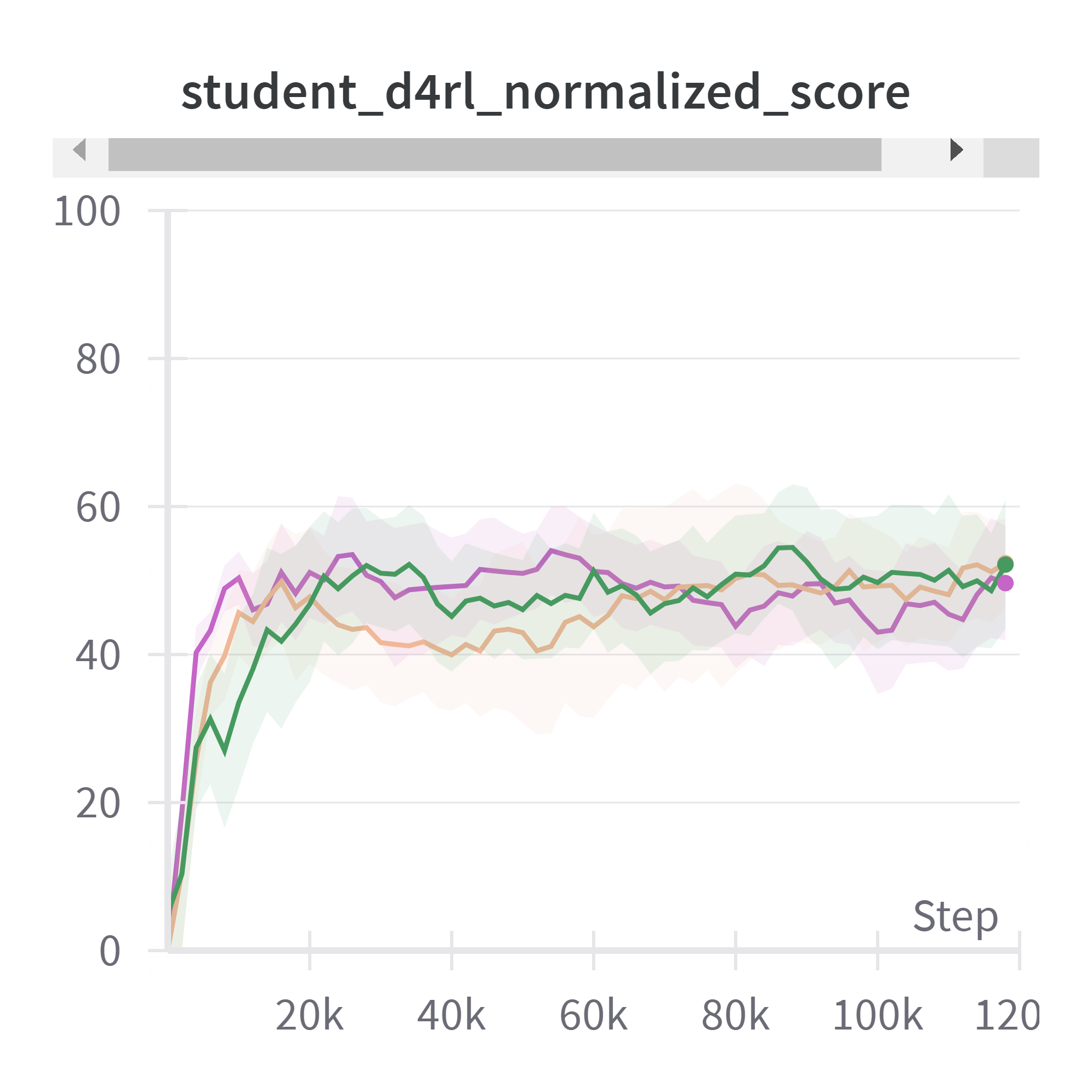}
\end{subfigure}
\vskip -0.1in
\centering
\includegraphics[width=0.6\textwidth,trim={0 11cm 0 0}, clip]{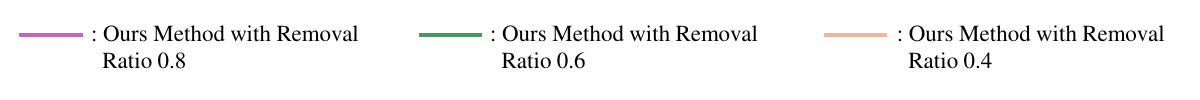}
\captionsetup{skip=1cm} 
\vskip -0.2in
\centering
\caption{Average normalized score depicting the impact of removal ratio for creating OOD offline RL dataset in four environments: HalfCheetah, Hopper, Walker2D, and Antmaze, arranged sequentially from left to right.}

\label{fig:OOD-removal}
\end{figure}

Then, we also show the performance curve of removal ratios in the domain shifted RL dataset. As described in Section \ref{knowledge-ood}, we intentionally create a OOD offline RL dataset by removing portions of the data — specifically 0.4, 0.6, and 0.8 (using fraction numbers to distinguish from unlabeled data coverage percentages) — from a specific range of the state space. Figure \ref{fig:OOD-removal} shows the performance curve of different settings of removal ratios for creating OOD offline RL dataset.

\newpage
\onecolumn
\section{Empirical Demonstration of the Student Policy Surpassing the Teacher's One (1\% BC baseline)}
\label{app:teacher-student}
As we mentioned before, one of the major concern is whether the student is merely replicating the teacher's policy (as same as whether the student outperforms the 1\% BC baseline). Here, we demonstrate the comparison between the teacher's and the student's performances during the training process. 
\begin{figure}[ht]
\begin{subfigure}{0.24\textwidth}
    \includegraphics[trim={0 0 0 12cm},clip,width=\textwidth]{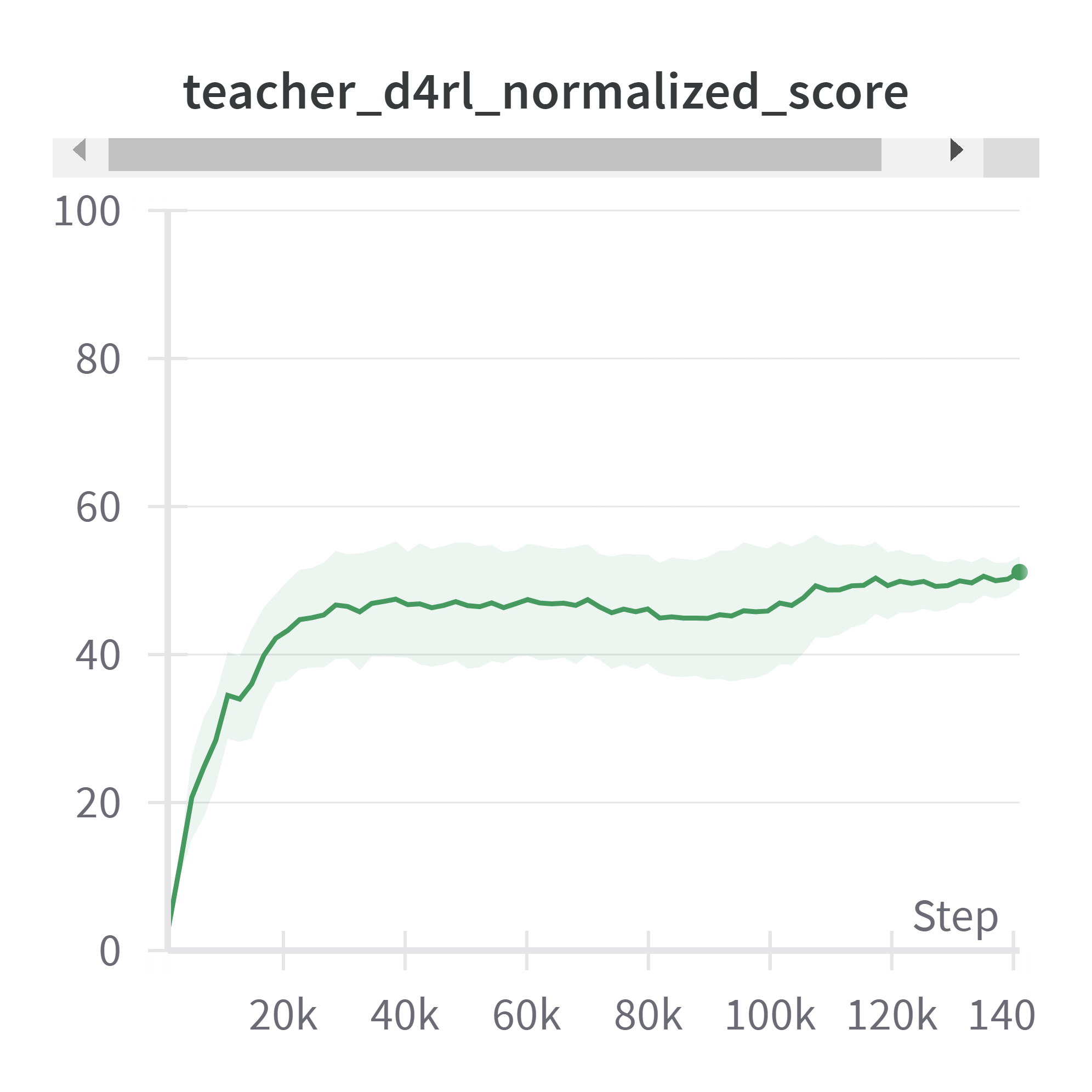}
\end{subfigure}
\hfill 
\begin{subfigure}{0.24\textwidth}
    \includegraphics[trim={0 0 0 12cm},clip,width=\textwidth]{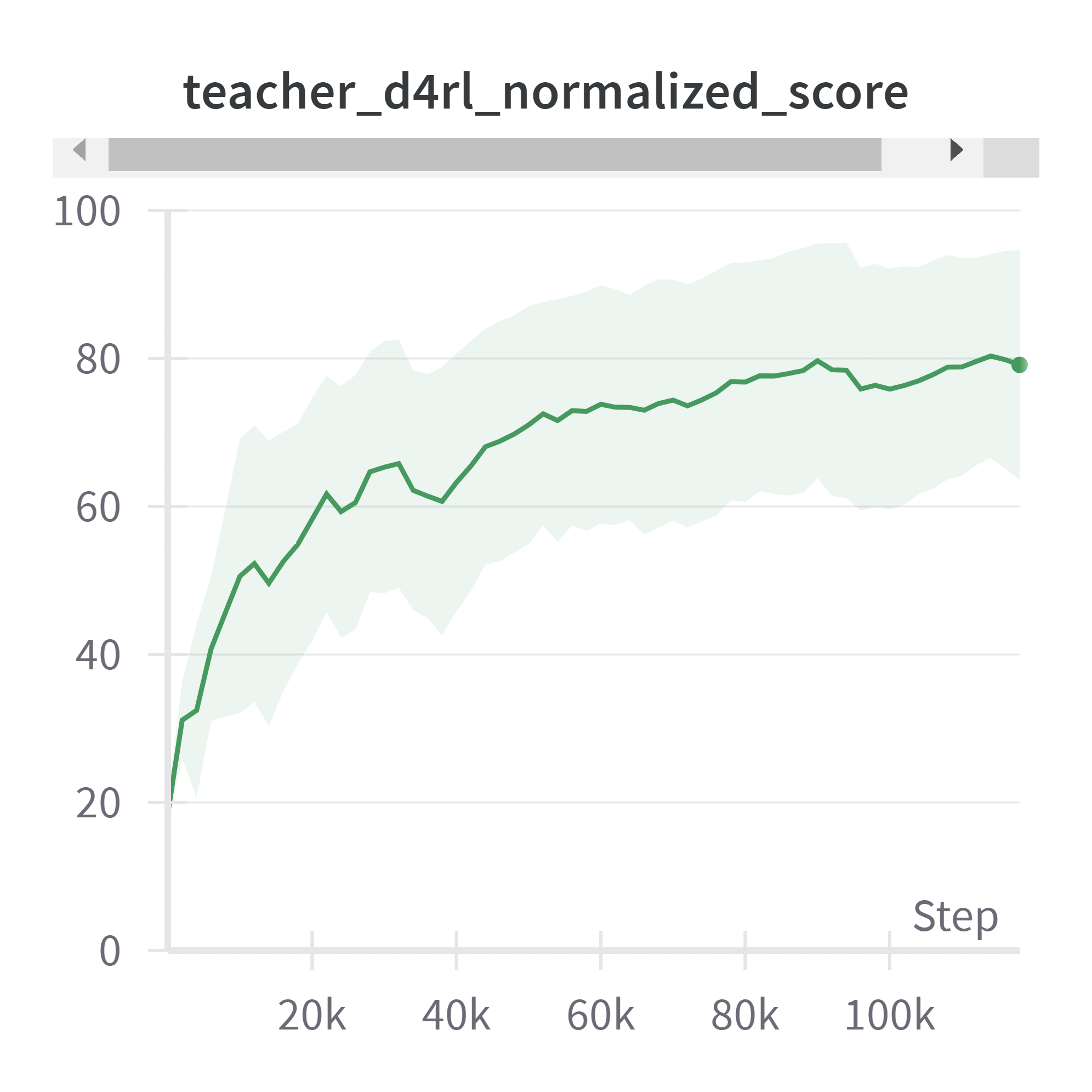}
\end{subfigure}
\hfill 
\begin{subfigure}{0.24\textwidth}
    \includegraphics[trim={0 0 0 12cm},clip,width=\textwidth]{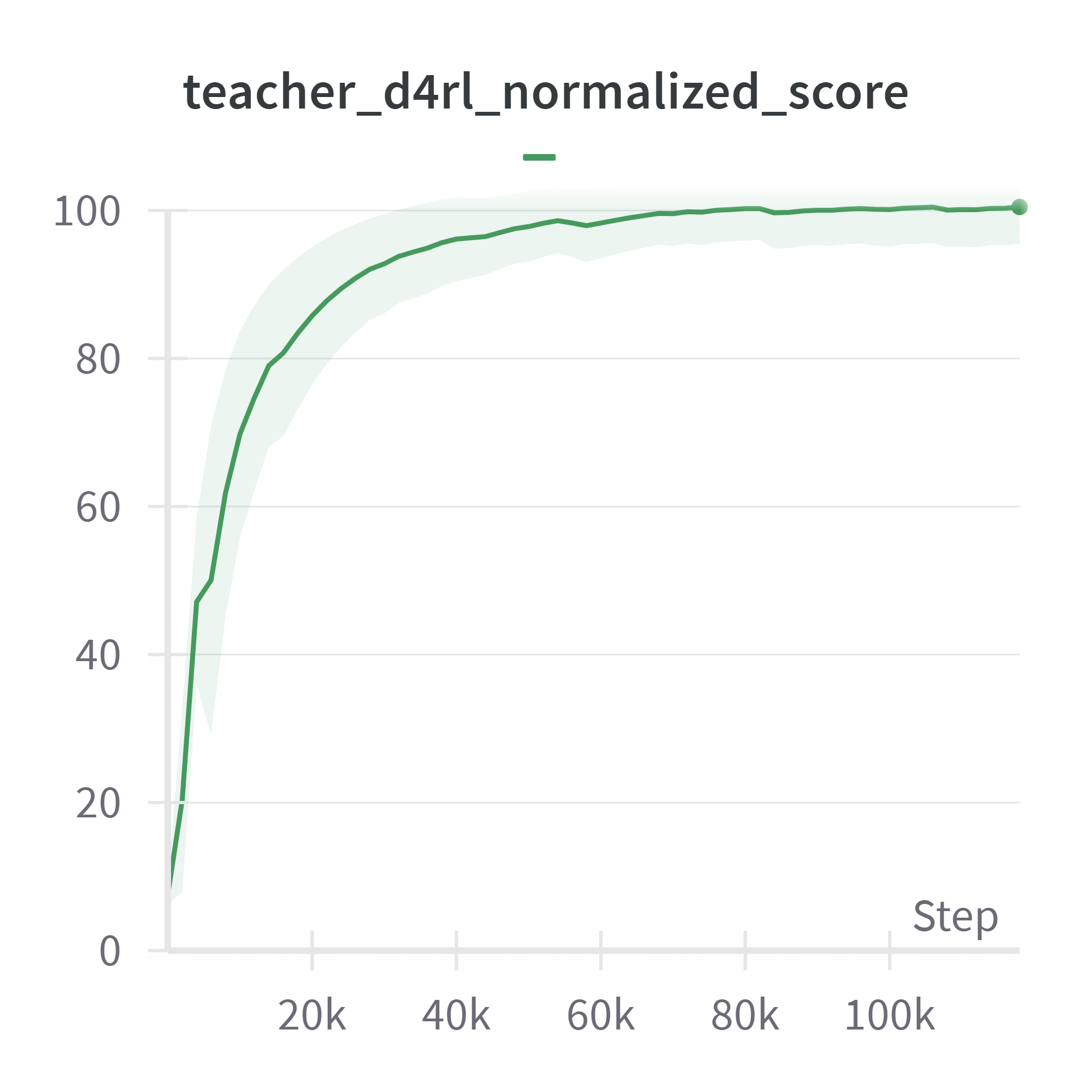}
\end{subfigure}
\hfill 
\begin{subfigure}{0.24\textwidth}
    \includegraphics[trim={0 0 0 12cm},clip,width=\textwidth]{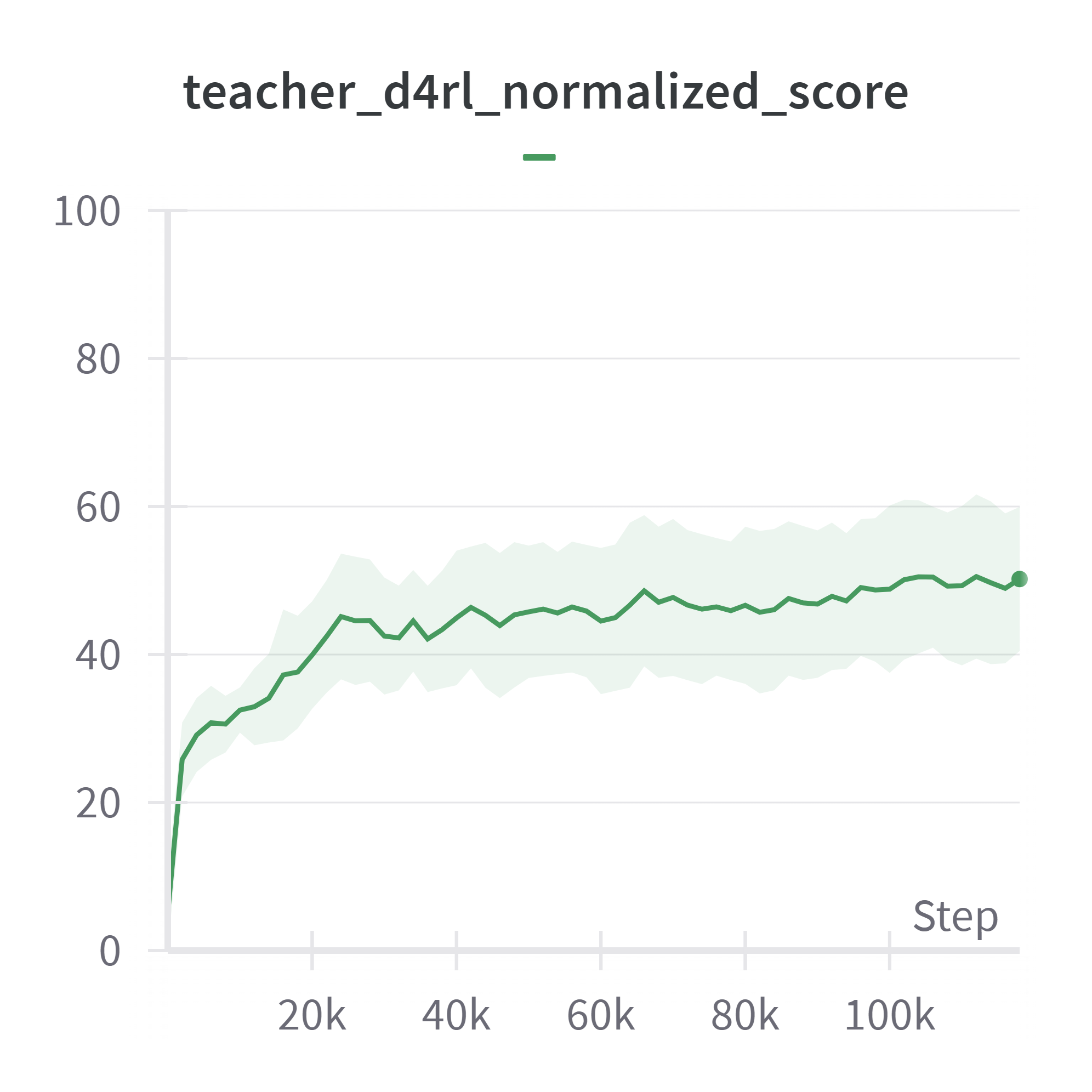}
\end{subfigure}
\vskip -0.1in
\centering
\includegraphics[width=0.6\textwidth,trim={0 11cm 0 0}, clip]{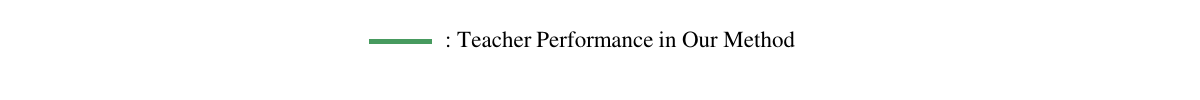}
\captionsetup{skip=1cm} 
\vskip -0.2in
\centering
\caption{Teacher performance in four environments: HalfCheetah, Hopper, Walker2D, and Antmaze, arranged sequentially from left to right.}

\label{fig:teacher}
\end{figure}

For the convenience of comparison, we show the correspondent student performance once again. These figures are the same figures of general performance in our experiment before. 
\begin{figure}[ht]
\begin{subfigure}{0.24\textwidth}
    \includegraphics[trim={0 0 0 12cm},clip,width=\textwidth]{figure/TD3BC-HalfCheetah-General.png}
\end{subfigure}
\hfill 
\begin{subfigure}{0.24\textwidth}
    \includegraphics[trim={0 0 0 12cm},clip,width=\textwidth]{figure/TD3BC-Hopper-General.png}
\end{subfigure}
\hfill 
\begin{subfigure}{0.24\textwidth}
    \includegraphics[trim={0 0 0 12cm},clip,width=\textwidth]{figure/TD3BC-Walker2D-General.png}
\end{subfigure}
\hfill 
\begin{subfigure}{0.24\textwidth}
    \includegraphics[trim={0 0 0 12cm},clip,width=\textwidth]{figure/TD3BC-Antmaze-General.png}
\end{subfigure}
\vskip -0.1in
\centering
\includegraphics[width=0.6\textwidth,trim={0 11cm 0 0}, clip]{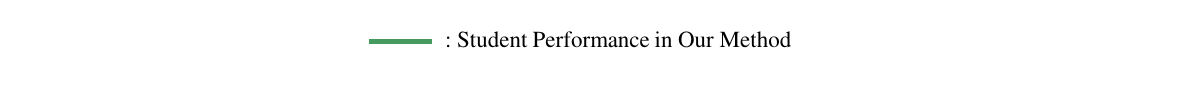}
\captionsetup{skip=1cm} 
\vskip -0.2in
\centering
\caption{Student performance in four environments: HalfCheetah, Hopper, Walker2D, and Antmaze, arranged sequentially from left to right.}

\label{fig:student}
\end{figure}

The performance curves depicted in Figures \ref{fig:teacher} and \ref{fig:student} demonstrate that our student model surpasses the teacher model in most tasks. This suggests that the student is not merely copying actions from the teacher; it is also learning from the offline RL dataset. We believe that our proposed teacher-student network, which can learn from both OOD offline RL dataset and unlabeled data data, is the key reason why the student model is able to outperform the teacher.

\newpage
\section{Ablation Studies Curves}
\label{app:ablation}

This section shows the performance curves of the ablation studies shown at \ref{AblationStudy}. 

\textbf{Component Ablation}. The performance curves for component ablation \ref{tab:Ablation-com} are illustrated below. 

\begin{figure}[ht]
\begin{subfigure}{0.24\textwidth}
    \includegraphics[trim={0 0 0 12cm},clip,width=\textwidth]{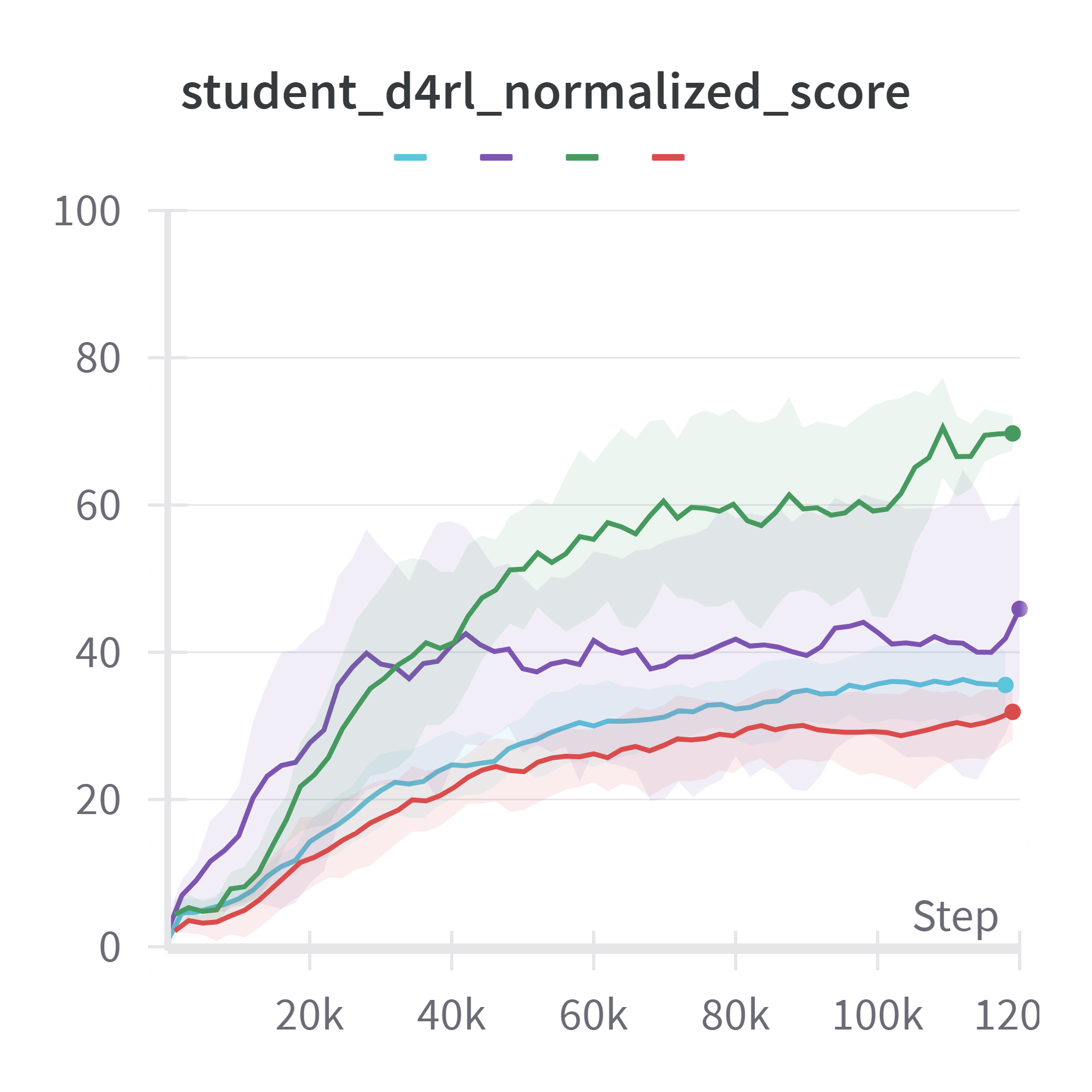}
\end{subfigure}
\hfill 
\begin{subfigure}{0.24\textwidth}
    \includegraphics[trim={0 0 0 13cm},clip,width=\textwidth]{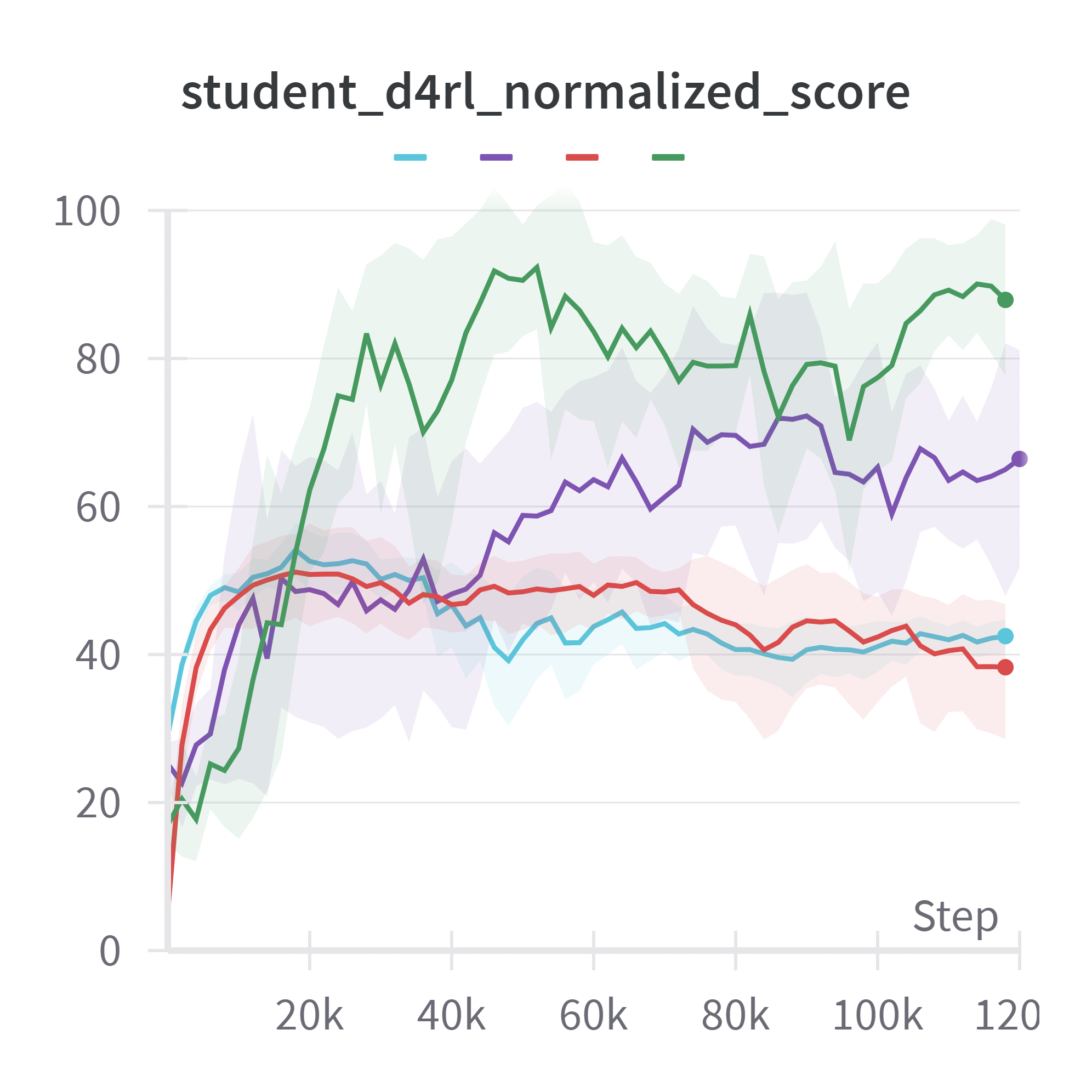}
\end{subfigure}
\hfill 
\begin{subfigure}{0.24\textwidth}
    \includegraphics[trim={0 0 0 12cm},clip,width=\textwidth]{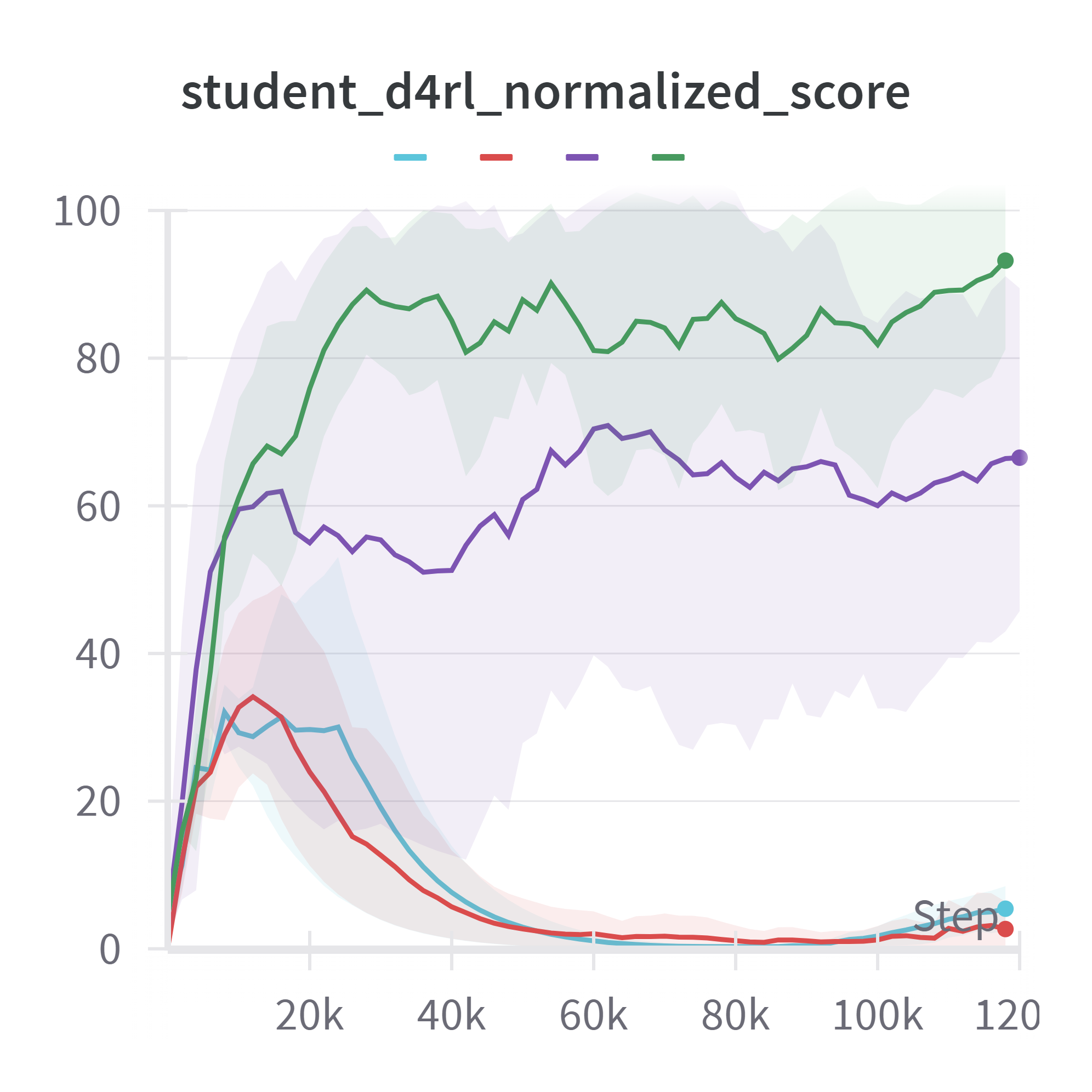}
\end{subfigure}
\hfill 
\begin{subfigure}{0.24\textwidth}
    \includegraphics[trim={0 0 0 12cm},clip,width=\textwidth]{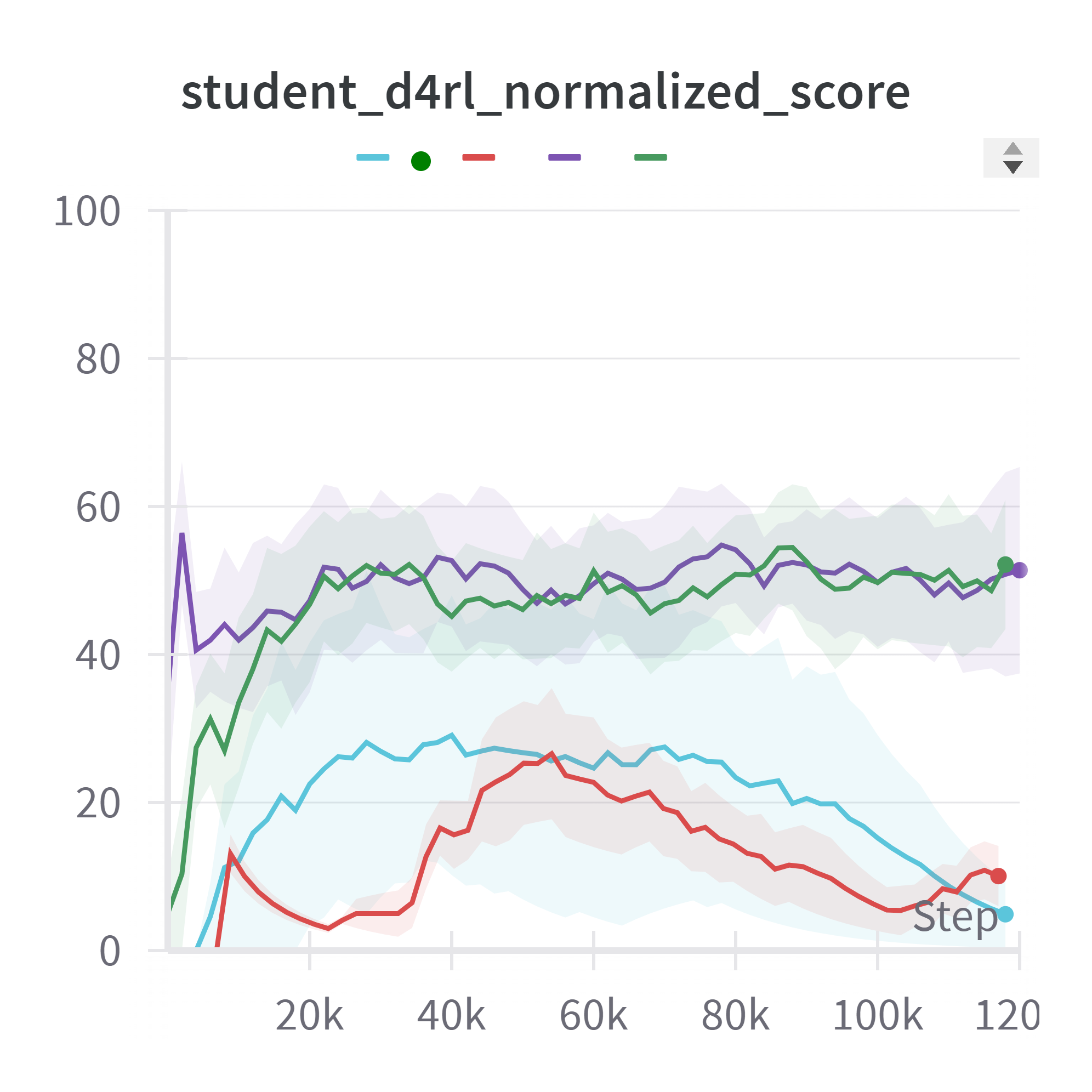}
\end{subfigure}
\vskip -0.1in
\centering
\includegraphics[width=0.6\textwidth,trim={0 11cm 0 0}, clip]{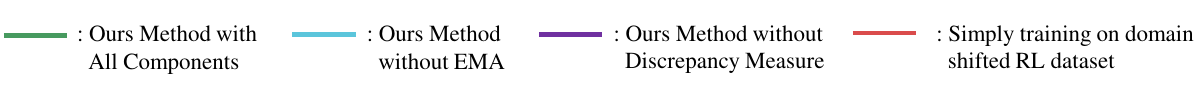}
\captionsetup{skip=1cm} 
\vskip -0.2in
\centering
\caption{Average normalized score on component ablation study}

\label{fig:ablation}
\end{figure}

\textbf{Data Utilization Ablation Curve}. The performance curves for Table \ref{tab:data_utilization}  are illustrated below. 

\begin{figure}[ht]
\begin{subfigure}{0.24\textwidth}
    \includegraphics[trim={0 0 0 12cm},clip,width=\textwidth]{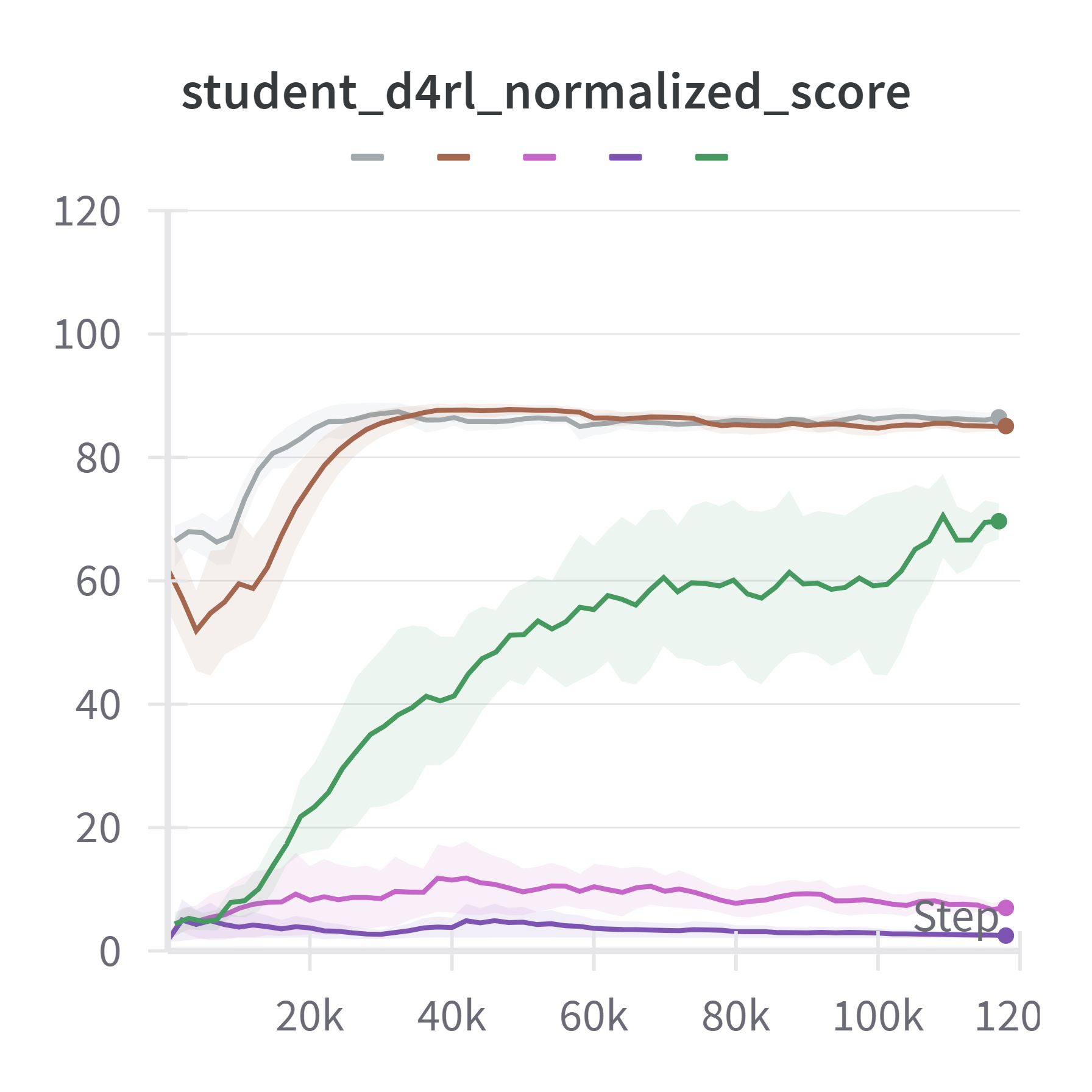}
\end{subfigure}
\hfill 
\begin{subfigure}{0.24\textwidth}
    \includegraphics[trim={0 0 0 13cm},clip,width=\textwidth]{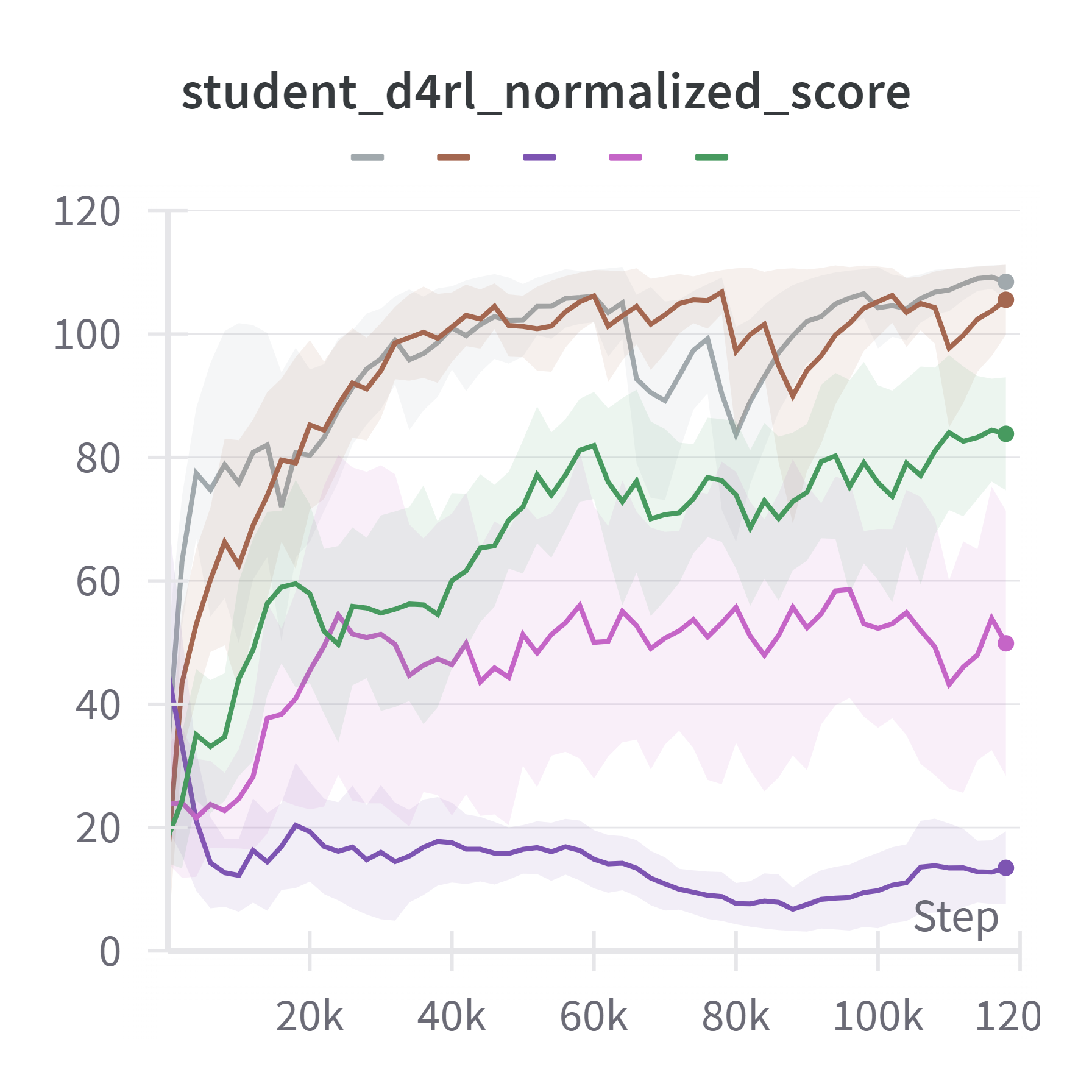}
\end{subfigure}
\hfill 
\begin{subfigure}{0.24\textwidth}
    \includegraphics[trim={0 0 0 12cm},clip,width=\textwidth]{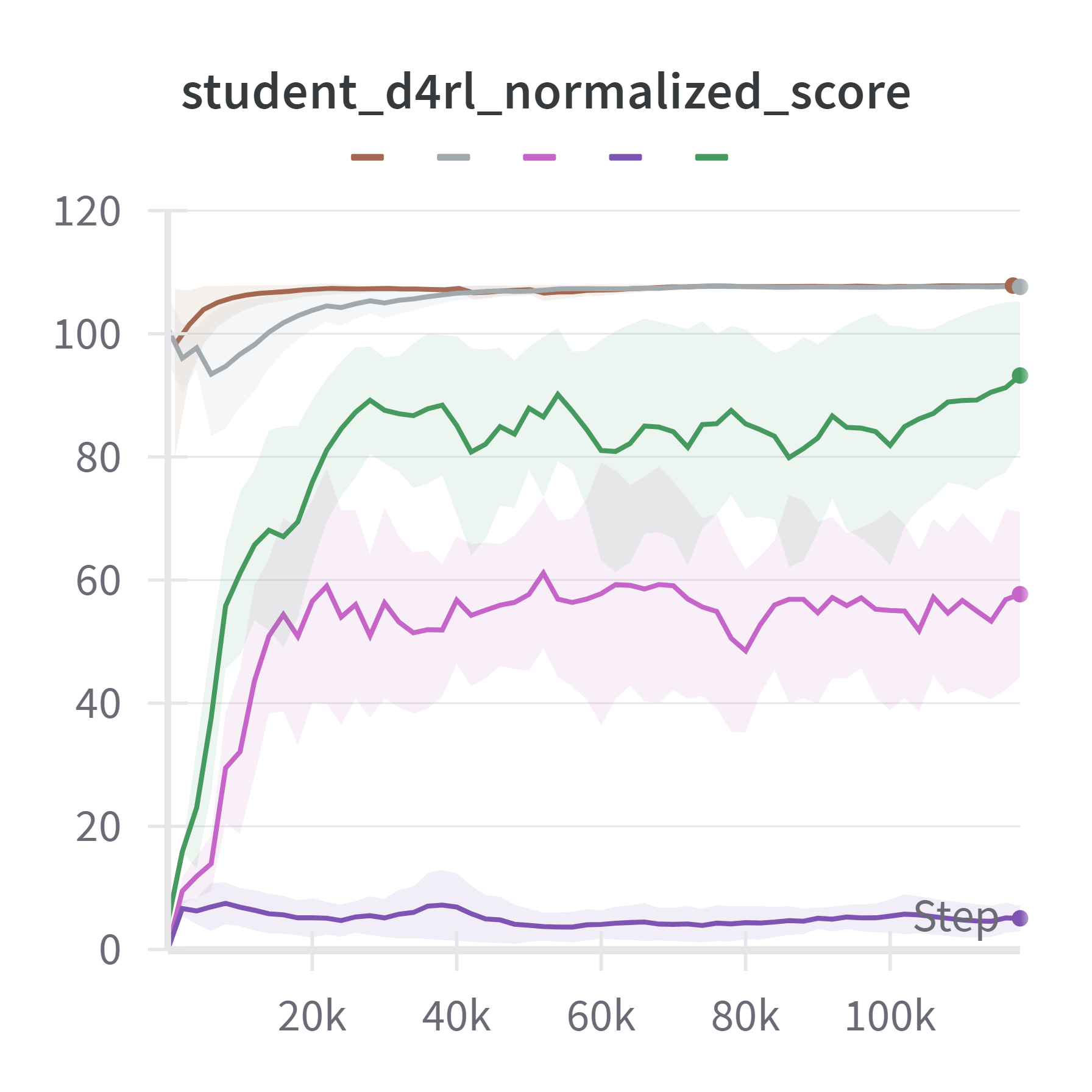}
\end{subfigure}
\hfill 
\begin{subfigure}{0.24\textwidth}
    \includegraphics[trim={0 0 0 12cm},clip,width=\textwidth]{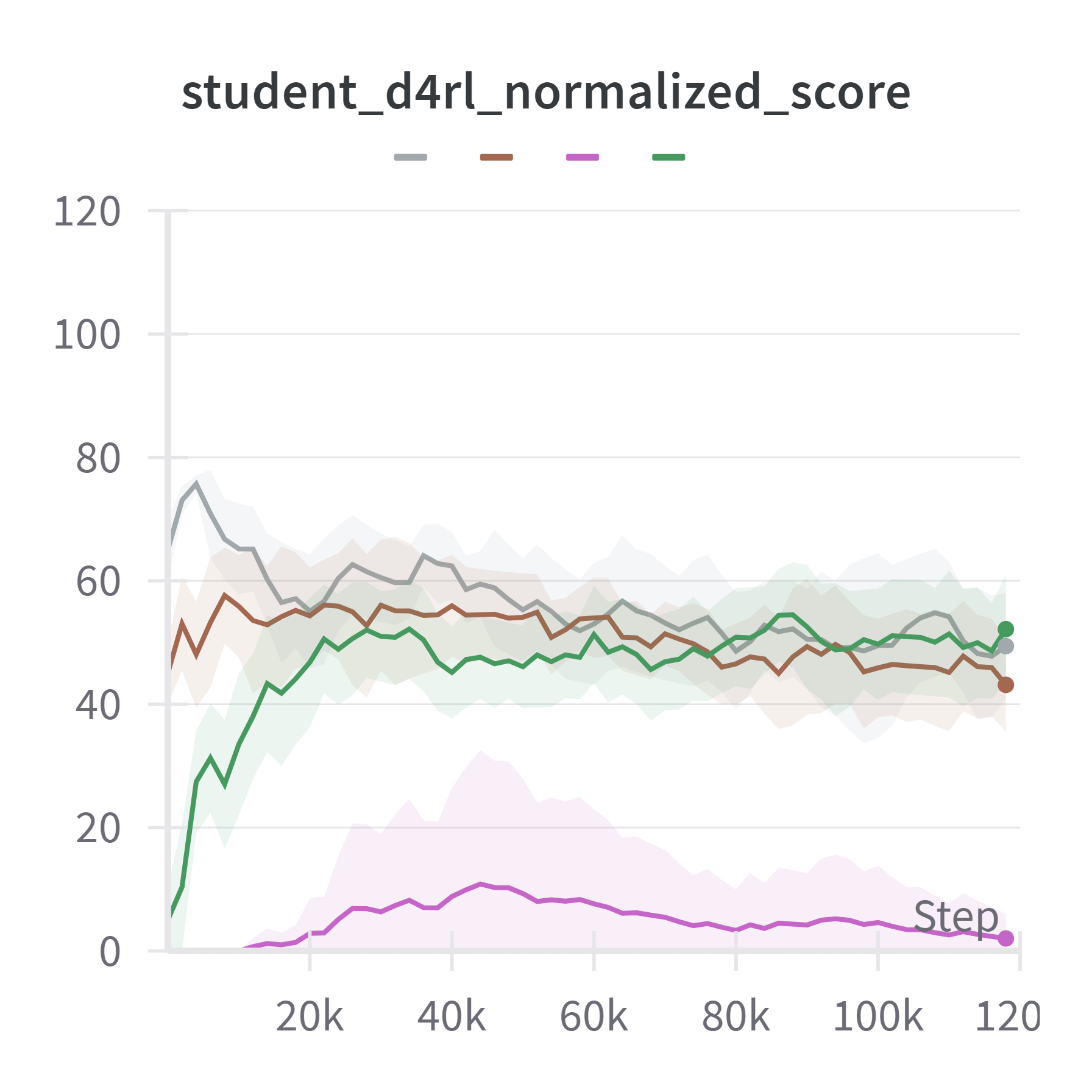}
\end{subfigure}
\vskip -0.1in
\centering
\includegraphics[width=0.6\textwidth,trim={0 11cm 0 0}, clip]{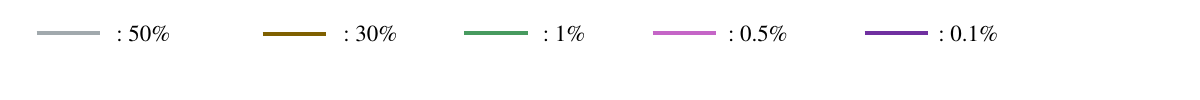}
\captionsetup{skip=1cm} 
\vskip -0.2in
\centering
\caption{Average normalized score on data utilization percentage (how much percentage of data used from original RL dataset) ablation study}

\label{fig:dataUsage}
\end{figure}

\newpage
\onecolumn
\section{Related Works on Off-policy Correction and Ablation on Discrepancy Measure}
\label{app::cos}

As described in Section 1, when OOD occurs in offline RL , the approximation of the value function or the trained policy comes with potential pitfalls \citep{offpolicy}. To address the OOD problem, we have decided to utilize unlabeled data, which typically originates from domain expertise or expert instruction. However, given the inherent discrepancy between the underlying distributions of the offline RL data and the unlabeled data in the OOD problem setting, it is necessary to develop discrepancy measures. These measures will assist our teacher-student framework in overcoming the OOD challenge.
We believe this topic is highly related to the traditional off-policy correction method, which is the correction method towards the discrepancy between offline data collection policy and agent's being trained policy. the studies of off-policy correction started by using a weighted importance sampling with linear function approximators\citep{NIPS2014_be53ee61} to sample the trajectories of offline RL data. For the continuous action control problem, IS has seen limited use in actor-critic continuous action domains\citep{saglam2022safe}. Techniques like DISC\citep{han2019dimension} address bias in clipped IS weights by clipping each action dimension separately. KL Divergence-based Batch Prioritized Experience Replay\citep{cicek2021off}(KLPER) optimizes on-policy sample prioritization in actor-critic algorithms. Recently, \citep{ying2022reuse} identified 'Reuse Bias' in off-policy evaluation due to repeated use of off-policy samples and proposed Bias-Regularized Importance Sampling (BIRIS) to reduce the off-policy error.

Similar to \citep{offpolicy}, we conducted a brief experiment to demonstrate the performance of three distinct kinds of action constraints: two different KL divergences and a JS divergence for all three D4RL tasks. For KL divergence, we implemented two types as mentioned in the previous rebuttal box: distribution formulated by a. (KL-1)automatically formulation across batch and b. (KL-2) manually specifying std. As the result shown in the Table \ref{tab:constraint-results}, there is a huge performance gap between the cosine similarity based one and the others.

\begin{table}[h]
\centering
\caption{Results from the ablation study on different action constraints}
\label{tab:constraint-results}
\begin{tabular}{ccccccc}
\toprule
K.D & RL.D & Task & KL-1 & KL-2 & JS & Cos \\ \midrule
Exp.1\% & Med.1\% & HalfCheetah & 10.92 & 30.23 & 37.79 & 69.73 \\ 
Exp.1\% & Med.1\% & Hopper & 14.45 & 38.26 & 50.53 & 82.92 \\ 
Exp.1\% & Med.1\% & Walker2d & 28.74 & 40.46 & 60.69 & 100.21 \\ \bottomrule
\end{tabular}
\end{table}

\begin{figure}[ht]
\begin{subfigure}{0.30\textwidth}
    \includegraphics[trim={0 0 0 12cm},clip,width=\textwidth]{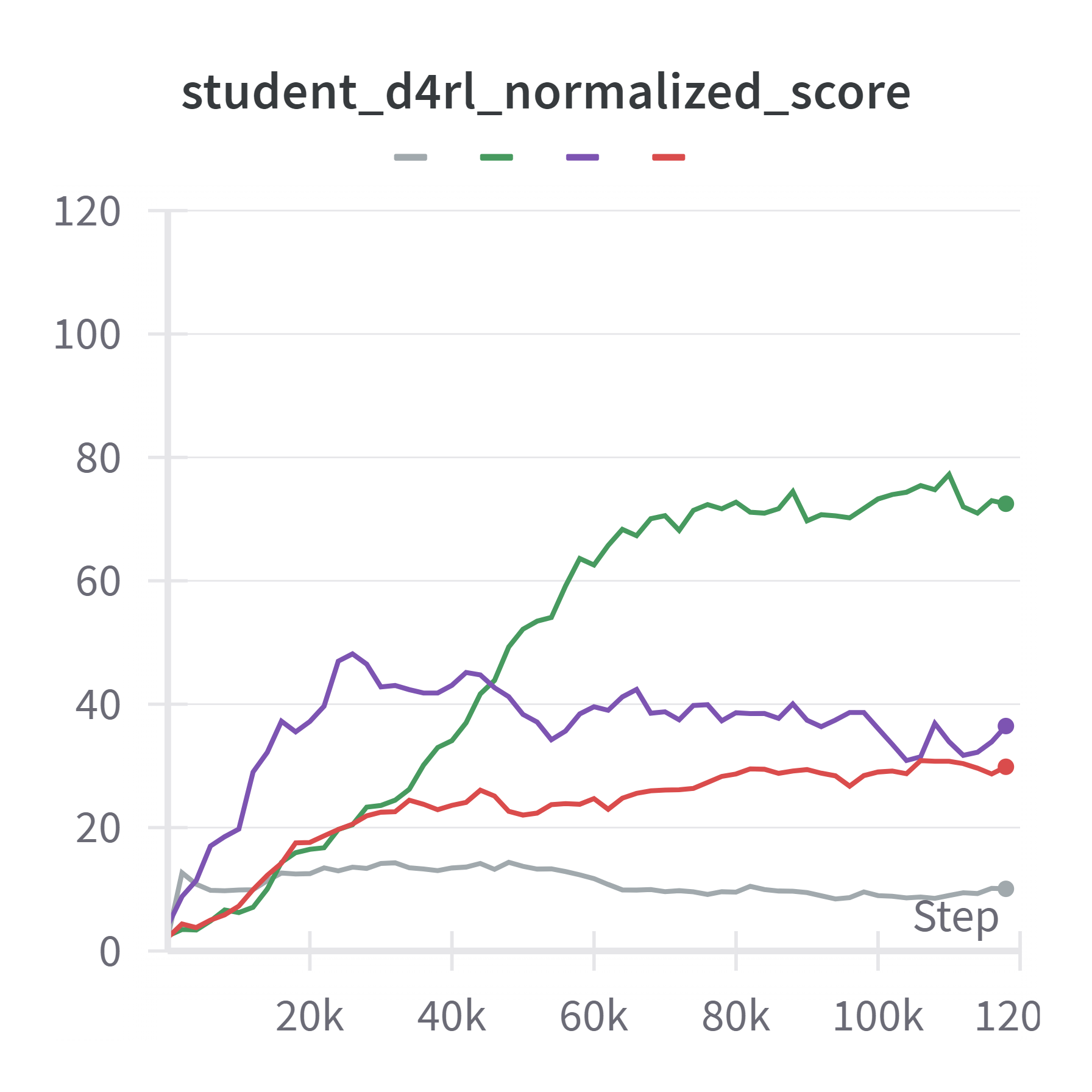}
\end{subfigure}
\hfill 
\begin{subfigure}{0.30\textwidth}
    \includegraphics[trim={0 0 0 12cm},clip,width=\textwidth]{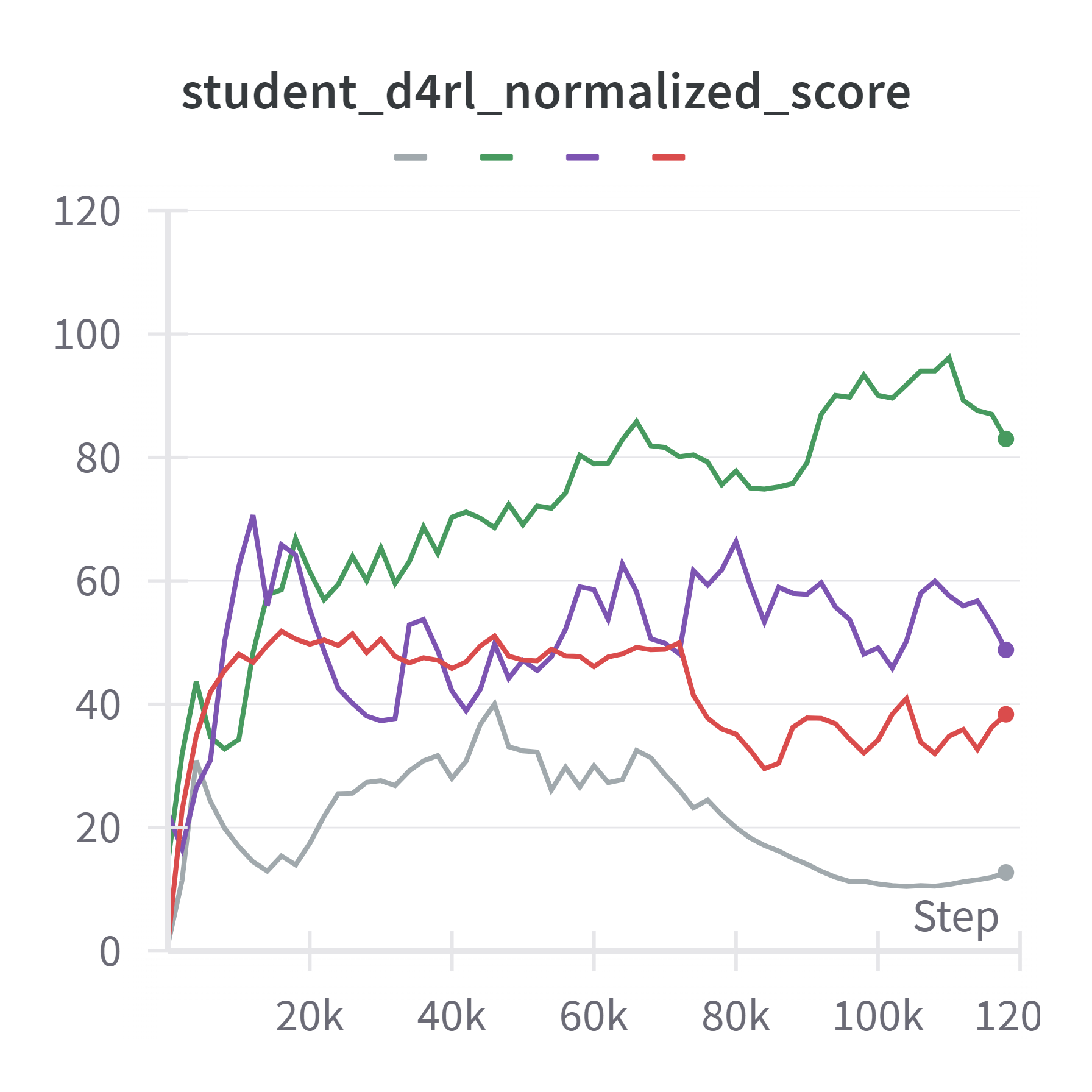}
\end{subfigure}
\hfill 
\begin{subfigure}{0.30\textwidth}
    \includegraphics[trim={0 0 0 12cm},clip,width=\textwidth]{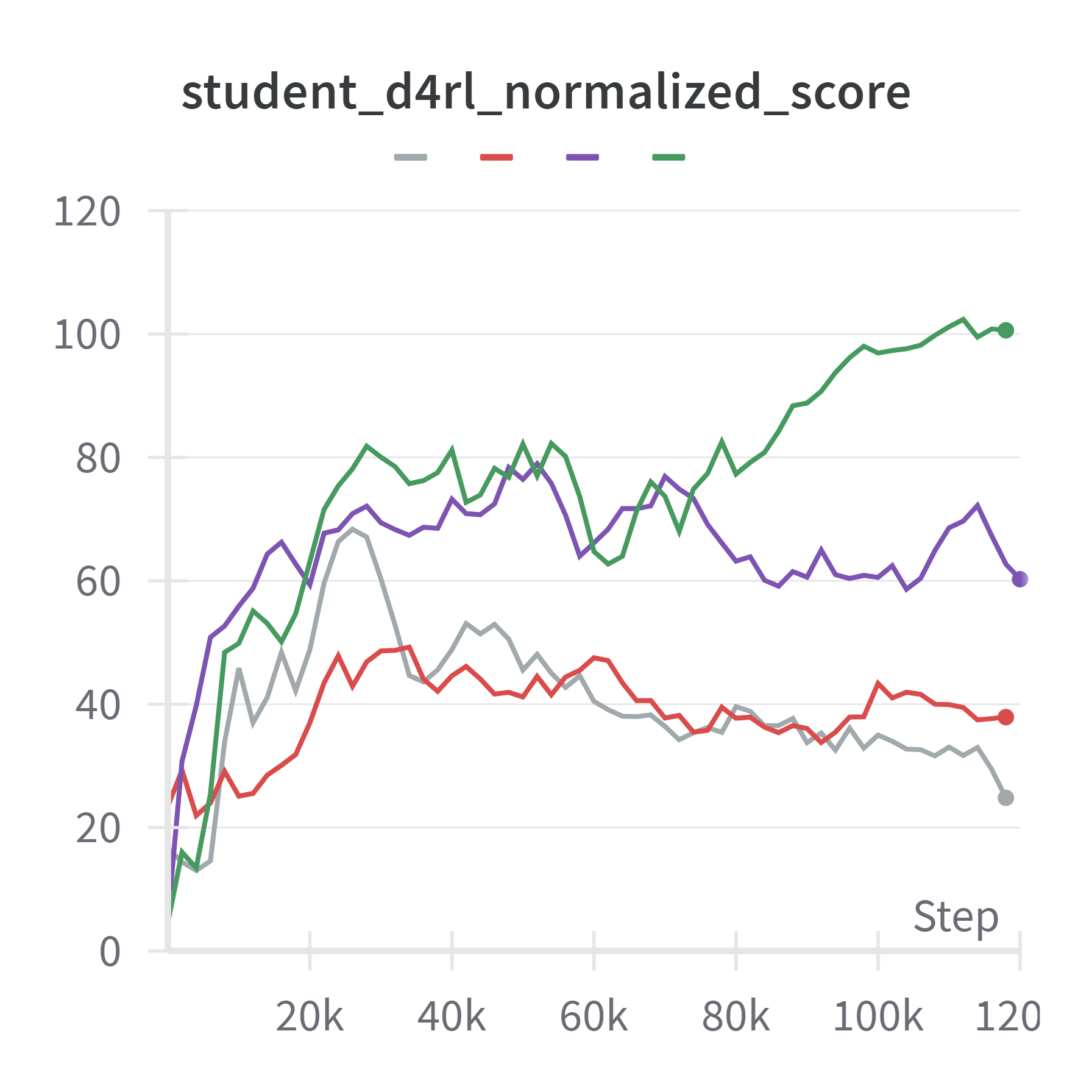}
\end{subfigure}
\hfill 
\vskip -0.1in
\centering
\includegraphics[width=0.6\textwidth,trim={0 11cm 0 0}, clip]{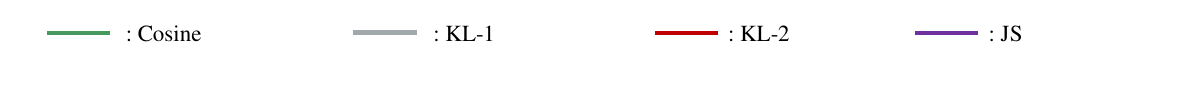}
\captionsetup{skip=1cm} 
\vskip -0.2in
\centering
\caption{Average normalized score on ablation study of different action constraints}

\label{fig:dataUsage}
\end{figure}

\newpage
\onecolumn
\section{Analyzing Data Removal in Different Observation Space to Create OOD Offline RL Datasets}
\label{app:features}

As previously introduced in our general experimental setting, to create a OOD offline RL dataset, we intentionally remove a certain percentage of specified data from the observation space. This raises a major concern: removing data from different dimensions of the observation space might lead to varied consequences, as the importance of each dimension can differ. In this section, we utilize the Hopper-Medium-v2 dataset from D4RL\citep{fu2020d4rl} to test how the performance of our proposed method varies when different dimensional data are removed from the observation spaces. The feature spaces of observations in the Hopper task are depicted in Table \ref{tab:hopper_state_dimensions} below.

\begin{table}[ht]
\centering
\caption{State Dimensions for Hopper-v2 in OpenAI Gym}
\begin{tabular}{clccp{5cm}}
\toprule
Index & Observation & Min & Max & Description \\
\midrule
0 & Z-Coordinate of Top & -Inf & Inf & Height of the hopper \\
1 & Angle of Top & -Inf & Inf & Angle of the hopper's body relative to the ground \\
2 & Angle of Thigh Joint & -Inf & Inf & Angle of the thigh joint \\
3 & Angle of Leg Joint & -Inf & Inf & Angle of the leg joint \\
4 & Angle of Foot Joint & -Inf & Inf & Angle of the foot joint \\
5 & Velocity of X-Coordinate of Top & -Inf & Inf & Velocity of the hopper along the ground \\
6 & Velocity of Z-Coordinate of Top & -Inf & Inf & Vertical velocity of the hopper \\
7 & Angular Velocity of Top & -Inf & Inf & Angular velocity of the hopper's body \\
8 & Angular Velocity of Thigh Joint & -Inf & Inf & Angular velocity of the thigh joint \\
9 & Angular Velocity of Leg Joint & -Inf & Inf & Angular velocity of the leg joint \\
10 & Angular Velocity of Foot Joint & -Inf & Inf & Angular velocity of the foot joint \\
\bottomrule
\end{tabular}

\label{tab:hopper_state_dimensions}
\end{table}

\begin{figure}[ht]
\centering

\foreach \i in {0,...,10} {
  \begin{subfigure}{0.2\textwidth}
    \includegraphics[width=\linewidth]{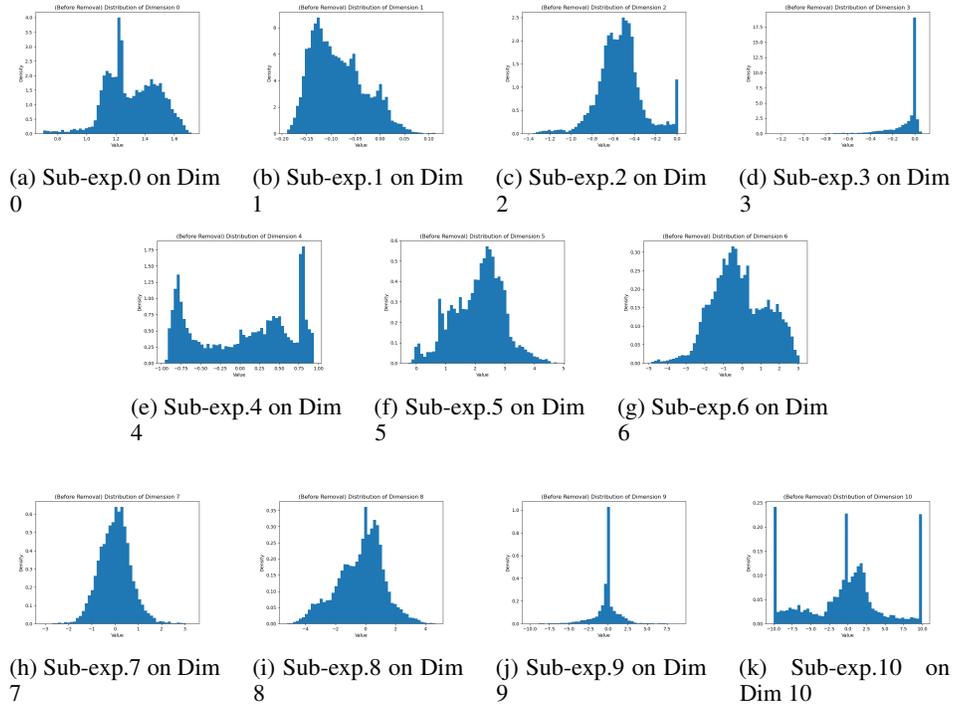}  
    \caption{Sub-exp.{\i} on Dim {\i}}
  \end{subfigure}\quad
  \ifnum\i=6 \par\bigskip\fi
  \ifnum\i=10 \par\bigskip\fi
}
\caption{Distribution of Each Dimension in Hopper-Medium-v2 Observations Before Removal of Densest 60\% in Each Dimension}
\label{fig:11}
\end{figure}

\begin{figure}[ht]
\centering

\foreach \i in {0,...,10} {
  \begin{subfigure}{0.2\textwidth}
    \includegraphics[width=\linewidth]{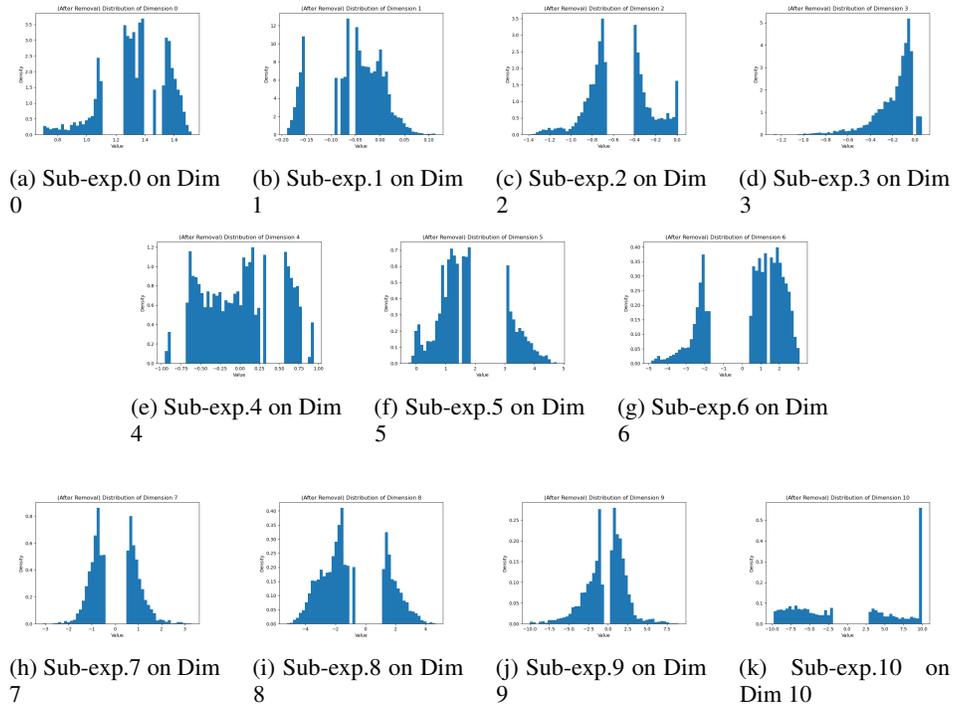}  
    \caption{Sub-exp.{\i} on Dim {\i}}
  \end{subfigure}\quad
  \ifnum\i=6 \par\bigskip\fi
  \ifnum\i=10 \par\bigskip\fi
}
\caption{Distribution of Each Dimension in Hopper-Medium-v2 Observations After Removal of Densest 60\% in Each Dimension}
\label{fig:22}
\end{figure}

In our analysis of the Hopper-Medium-v2 environment, we removed the most densely populated segments within each observational dimension for each experiment. Specifically, we identified and excluded the densest 60\% of data points in each dimension, effectively filtering out sections with the highest concentration of observations. This strategic removal was based on the assumption that such dense regions could potentially skew the overall analysis or obscure subtler, yet significant, patterns within the dataset. Figure \ref{fig:11} and \ref{fig:22} illustrate one experiment in our removal strategy and the resulting distribution shift in dimension 0 when the most dense part of this dimension is removed. We also noted that, except for dimension 0, other dimensions (from dimension 1 to dimension 10) showed only slight changes in value distribution even when some data were removed.

\begin{figure}[ht]
    \centering
    \includegraphics[trim={0 0 0 12cm},clip,width=0.9\textwidth]{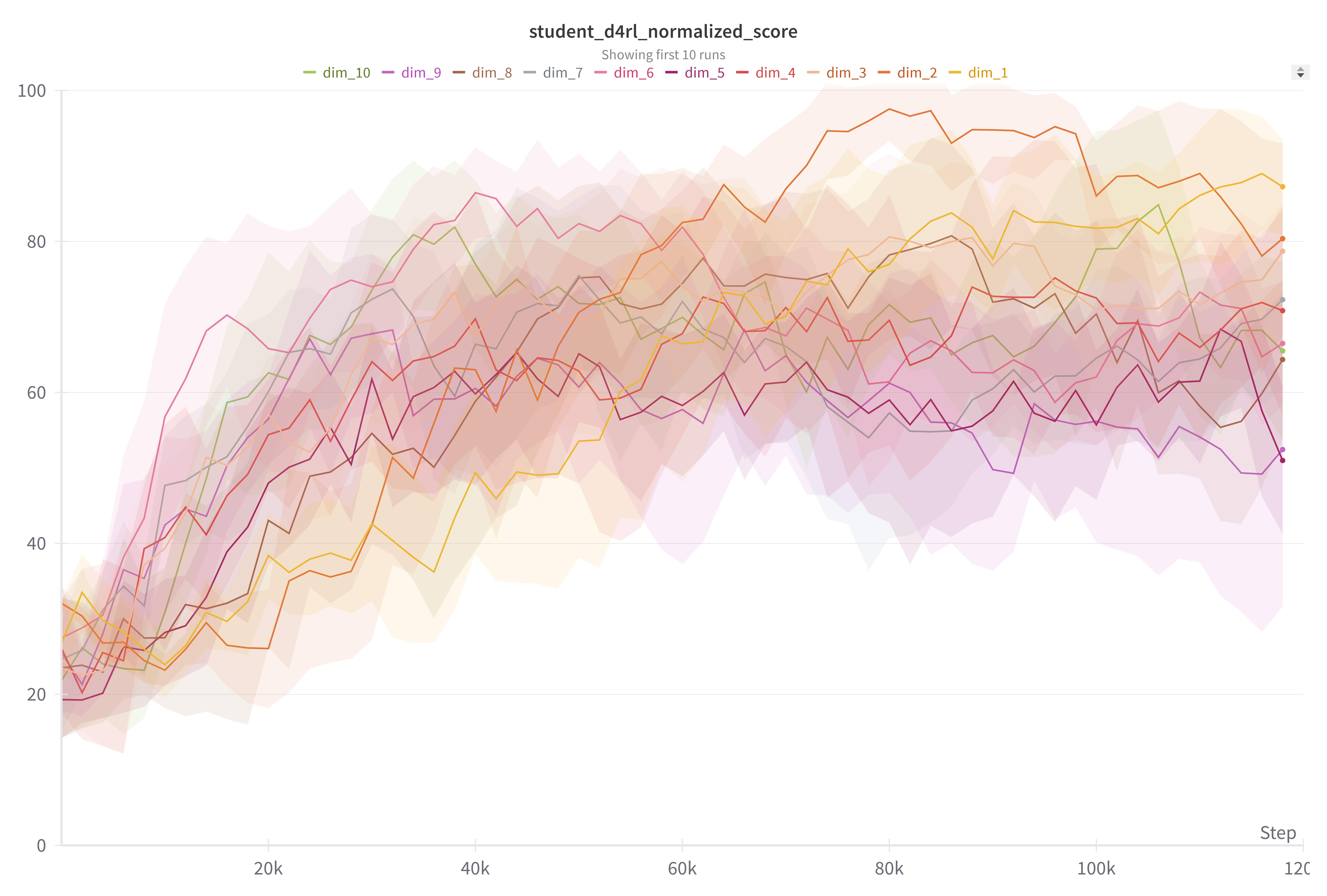}
    \caption{Performance Scores after Data Removal Across Various Observation Dimensions in Hopper-Medium-v2}
    \label{fig:obs_scores}
\end{figure}

Following this data removal process, we observed slight shifts in the normalized performance scores across all dimensions. As depicted in Table \ref{tab:performance_scores}, which provides a comparative view of the performance scores post-removal, we can discern that: 1) removing the same percentage of data from different dimensions of observation spaces leads to varying performance scores; 2) overall, there are no extremely significant features that cause a huge performance gap when removing 30\% of data from the most densely populated part of those dimensions. Figure \ref{fig:obs_scores} shows the performance curves for each experiment when data is removed from different dimensions of observation in the Hopper task.

\begin{table}[ht]
\centering
\caption{Normalized Performance Scores After Dense Part Removal in Each Hopper-Medium-V2 Observational Dimension}
\label{tab:performance_scores}
\scriptsize 
\begin{tabular}{@{}ccccccc@{}}
\toprule
 & Dim 0 & Dim 1 & Dim 2 & Dim 3 & Dim 4 & Dim 5 \\ \midrule
Normalized Score & 79.95$\pm$14.77 & 87.24$\pm$22.98 & 80.37$\pm$15.71& 78.70 $\pm$18.89 & 70.83 $\pm$21.15 & 50.98 $\pm$17.32\\ 
\end{tabular}

\begin{tabular}{@{}ccccccc@{}}
\toprule
 & Dim 6 & Dim 7 & Dim 8 & Dim 9 & Dim 10 \\ \midrule
Normalized Score & 66.48 $\pm$12.13 & 72.28 $\pm$12.32  & 64.35 $\pm$18.96 & 52.43 $\pm$24.26 & 65.50 $\pm$19.21 \\
\bottomrule
\end{tabular}
\end{table}

\end{document}